\documentclass[sn-mathphys-ay]{sn-jnl}% Math and Physical Sciences Author Year Reference Style
\setcitestyle{aysep={}}%Removes comma between author and year in citations as in the published paper
\usepackage{graphicx}%
\usepackage{multirow}%
\usepackage{amsmath,amssymb,amsfonts}%
\usepackage{amsthm}%
\usepackage{mathrsfs}%
\usepackage[title]{appendix}%
\usepackage{xcolor}%
\usepackage{textcomp}%
\usepackage{manyfoot}%
\usepackage{booktabs}%
\usepackage{algorithm}%
\usepackage{algorithmicx}%
\usepackage{algpseudocode}%
\usepackage{listings}%
\usepackage{array}
\usepackage{ulem}
\usepackage{utfsym}
\usepackage{booktabs}
\usepackage{soul}
\usepackage{url}

\begin{document}

\title[Article Title]{\bf Edge Deep Learning in Computer Vision and Medical Diagnostics: A Comprehensive Survey}


\author{Yiwen Xu$^{*\dagger}$, Tariq M. Khan$^{\dagger}$, Yang Song, Erik Meijering}

\affil{\normalsize
School of Computer Science and Engineering\break
University of New South Wales\break
Sydney, NSW 2052, Australia\break
\break
$^{*}$Corresponding author: \href{mailto:yiwen.xu1@unsw.edu.au}{yiwen.xu1@unsw.edu.au}\break
$^{\dagger}$These authors contributed equally.\break
\break
Published in \textit{Artificial Intelligence Review} 58(3):93 March 2025.\break
\break
Updated version with corrections in the text and references.
}

%\author[1]{\fnm{Yiwen}\sur{Xu}}\email{yiwen.xu1@unsw.edu.au}
%\equalcont{These authors contributed equally.}
%\author[1]{\fnm{Tariq} \sur{M.\ Khan}}
%\equalcont{These authors contributed equally.}
%\author[1]{\fnm{Yang} \sur{Song}}
%\author[1]{\fnm{Erik} \sur{Meijering}}

%\affil[1]{\orgdiv{School of Computer Science and Engineering}, \orgname{University of New South Wales}, \orgaddress{\city{Sydney}, \postcode{NSW 2052}, \country{Australia}}}


\abstract{Edge deep learning, a paradigm change reconciling edge computing and deep learning, facilitates real-time decision making attuned to environmental factors through the close integration of computational resources and data sources. Here we provide a comprehensive review of the current state of the art in edge deep learning, focusing on computer vision applications, in particular medical diagnostics. An overview of the foundational principles and technical advantages of edge deep learning is presented, emphasising the capacity of this technology to revolutionise a wide range of domains. Furthermore, we present a novel categorisation of edge hardware platforms based on performance and usage scenarios, facilitating platform selection and operational effectiveness. Following this, we dive into approaches to effectively implement deep neural networks on edge devices, encompassing methods such as lightweight design and model compression. Reviewing practical applications in the fields of computer vision in general and medical diagnostics in particular, we demonstrate the profound impact edge-deployed deep learning models can have in real-life situations. Finally, we provide an analysis of potential future directions and obstacles to the adoption of edge deep learning, with the intention to stimulate further investigations and advancements of intelligent edge deep learning solutions. This survey provides researchers and practitioners with a comprehensive reference shedding light on the critical role deep learning plays in the advancement of edge computing applications.}

\keywords{Edge Computing, Deep Learning, Computer Vision, Medical Diagnostics, Lightweight Neural Networks}



\maketitle

\section{Introduction}
\begin{figure}[!t]
    \centering
    \includegraphics[width=\textwidth]{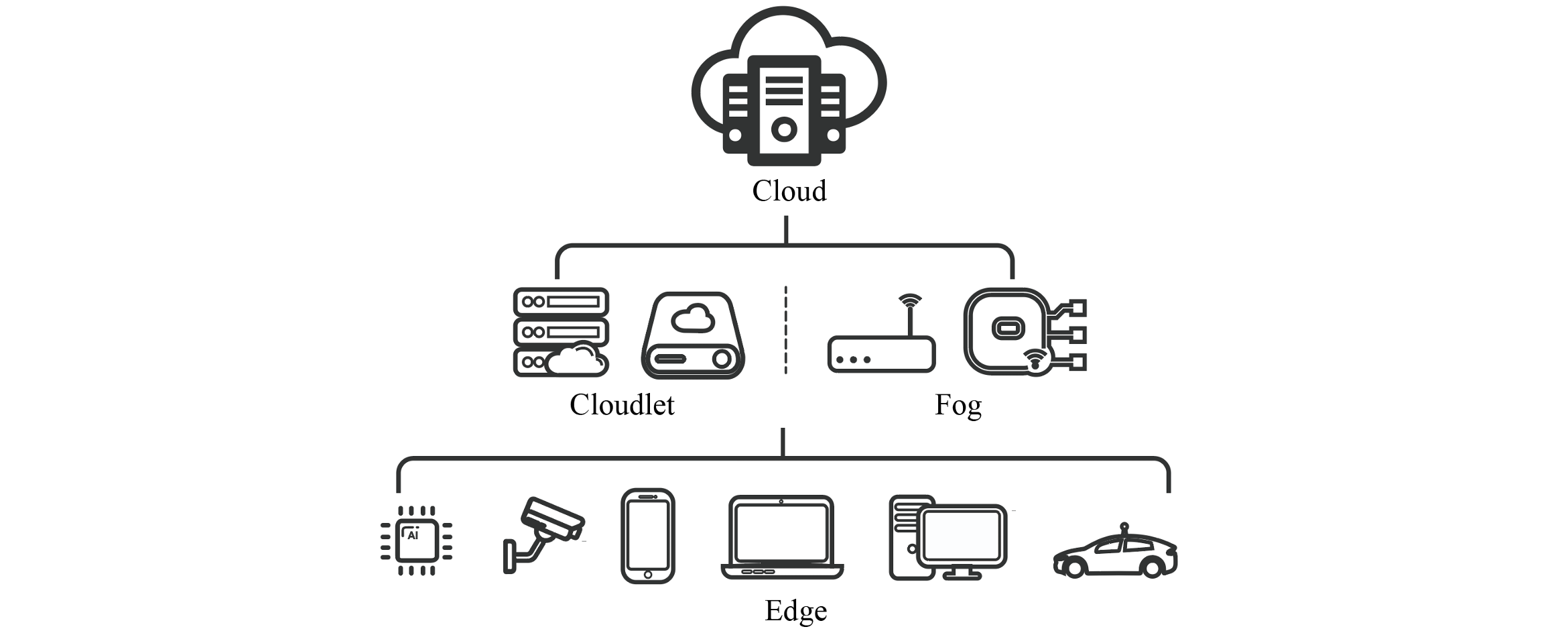}
    \vspace{-1.5\baselineskip}
    \caption{Illustrative overview of edge computing in relation to cloud computing.}
    \label{fig:edge_computing_framework}
\end{figure}

Edge computing (Fig.~\ref{fig:edge_computing_framework}), an emerging computing paradigm, is increasingly impacting the modern technological landscape by executing computational tasks near the data source. This not only increases response speed but also brings improvements in security, privacy protection, scalability, and distributed processing. Edge computing \citep{cao2020overview} aims to process data on devices proximate to the data-generating source (edge devices) rather than transmitting it to distant cloud servers. In fact, edge computing and cloud computing are not mutually exclusive \citep{shi2016edge}. Instead, the edge serves as a complement and extension to the cloud, reducing communication latency, protecting data privacy, and offering faster responses. Variations of edge computing, such as fog computing \citep{bonomi2012fog} and cloudlets \citep{babar2021cloudlet}, converge on a central philosophy: decentralising computational capabilities closer to the data origins. This not only alleviates the computational burden on cloud centres, but also provides the necessary support for real-time or near-real-time applications, catering to the increasing demands of the modern digital society \citep{chen2019deep}.

Deep learning (DL), particularly the development of advanced convolutional neural networks (CNNs), has made significant strides in many areas of computer vision, offering powerful tools for image and video data processing \citep{voulodimos2018deep,dong2021survey,esteva2021deep,zhang2023applying,jiang2023deep}. Typically, DL models are created through intensive training on high-performance computing hardware before being deployed on devices. However, even though the time required to execute a model is significantly shorter than to train it, the inherent computational complexity of modern models often poses challenges when deploying them to resource-constrained edge devices. To tackle this, researchers are shifting focus towards model compression and lightweight model design, ensuring that deep learning models operate efficiently on edge devices \citep{kumar2021mobihisnet,zhang2021csl,huang2023ffkd}. 
%Our discussion will primarily center on the deployment and acceleration of these models on such devices.

In recent years, with the advancement of edge devices and the rapid development of DL technologies, we have entered a new era marked by ubiquitous and continuously increasing levels of on-device intelligence. The amalgamation of edge computing and deep learning, often termed Edge DL \citep{shuvo2022efficient}, heralds new possibilities for various applications. Edge DL refers to the deployment of DL algorithms and models on edge devices, capitalising on the benefits of both local computation and artificial intelligence. This convergence has brought about advancements in autonomous driving and intelligent traffic control \citep{wang2021real,liu2023real,wan2022edge}, industrial automation \citep{yi2022defect,hang2022surface,chen2022detection}, agriculture monitoring \citep{dang2020uav,albattah2023custom}, retail monitoring \citep{lachhab2023deep,kanjula2022people}, wildlife conservation \citep{gotthard2023edge,arshad2020my}, smart cities \citep{avvenuti2022spatio,di2022embedded}, sport analytics \citep{cui2022application,cioppa2020context}, and medical diagnostics \citep{kara2023smart,iqbal2023ldmres, khan2024esdmr, matloob2024lmbis, khan2024lmbf}. Specifically, in medical diagnostics, edge computing can provide instantaneous feedback through medical image analysis and patient monitoring, enhancing diagnostic efficiency and patient safety \citep{greco2020trends,idlahcen2024exploring}. Edge DL represents a paradigm shift in the deployment of deep neural networks (DNNs), bringing artificial intelligence (AI) closer to the data source and facilitating real-time, autonomous decision-making that is sensitive to and informed by the surrounding environment.

According to recent reports, the global edge computing in healthcare market was valued at USD 5.28 billion in 2023,\footnote{Polaris Market Research: Edge Computing in Healthcare Market Share, Size, Trends. \url{https://www.polarismarketresearch.com/industry-analysis/edge-computing-in-healthcare-market}} and is projected to reach USD 12.9 billion by 2028, growing at a compound annual growth rate (CAGR) of 26.1\% from 2022 to 2028.\footnote{Markets and Markets: Edge Computing in Healthcare Market: Growth, Size, Share and Trends. \url{https://www.marketsandmarkets.com/Market-Reports/edge-computing-in-healthcare-market-133588379.html}} The adoption rate of DL algorithm-based medical devices has also seen a significant increase, demonstrating steady growth.\footnote{Intel: Healthcare and Life Sciences: Smart Healthcare That Moves the World. \url{https://www.intel.com/content/www/us/en/healthcare-it/edge-analytics.html}}$^,$\footnote{NVIDIA Developer: How Edge Computing is Transforming Healthcare. \url{https://developer.nvidia.com/blog/healthcare-at-the-edge/}} These advancements highlight the potential of DL in enhancing the efficiency, and accessibility of medical diagnostics.

\begin{figure}[!t]
    \centering
    \includegraphics[width=\textwidth]{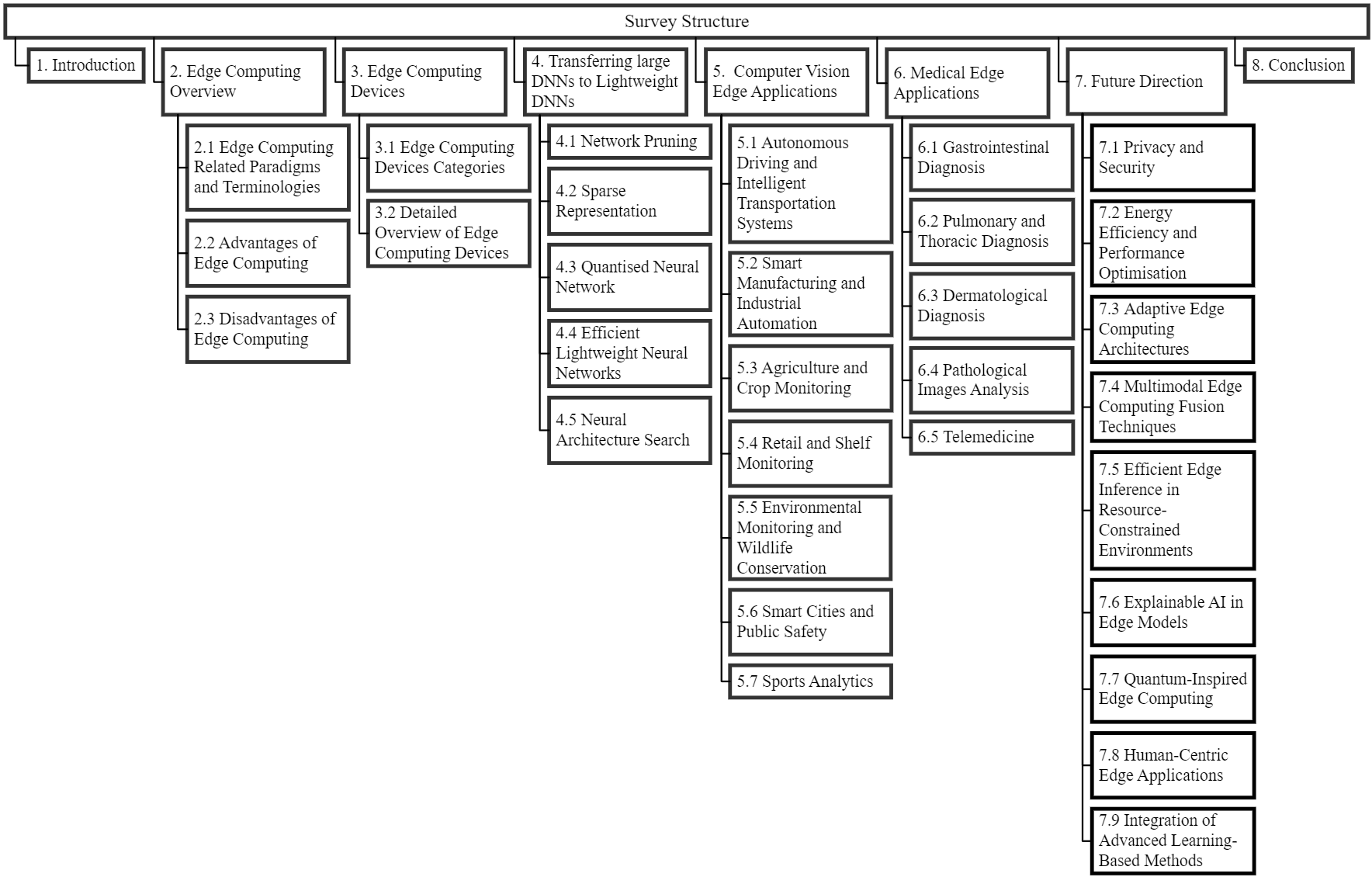}
    \caption{Schematic overview of the survey.}
    \label{fig:survey_structure}
\end{figure}

In this paper, we present a comprehensive survey of Edge DL with a focus on applications in computer vision in general and medical diagnostics in particular (Fig.~\ref{fig:survey_structure}). Providing a more in-depth discussion of the state-of-the-art specifically in these domains, our survey complements previous works exploring the fusion of edge computing and DL (Table \ref{fig:survey_summary}). We begin by elucidating the fundamental concepts, terminologies, and technical merits of edge computing (Section~\ref{sec:overview}), as well as categorising and exploring existing edge devices (Section~\ref{sec:devices}). Subsequently, our focus shifts to discussing strategies for model compression and lightweight model design (Section~\ref{sec:strategies}), ensuring efficient CNN operation on edge devices. Next, we review the applications of edge computing in computer vision in general (Section~\ref{sec:cvapps}) and medical diagnostics in particular (Section~\ref{sec:medapps}). Finally, we highlight pivotal research challenges and future opportunities (Section~\ref{sec:future}) and summarise the main conclusions (Section~\ref{sec:conclusion}). Building on existing literature, our survey highlights the important role of DL in advancing edge computing applications and aims to provide a comprehensive reference for researchers and practitioners.

\begin{table*}[!t]
\centering
\resizebox{\textwidth}{!}{%
\begin{tabular}{l>{\raggedright}p{9cm}ccccc}
\toprule
\multirow{2}{*}{\bf Paper} & \multirow{2}{*}{\bf Summary} & \multicolumn{5}{c}{\bf Scope} \\
\cmidrule{3-7} 
 & & \bf Hardware & \bf Lightweight & \bf Compression & \bf Vision & \bf Medical \\
\midrule
\cite{chen2019deep} & Reviews recent deep learning methods for edge computing, focusing on IoT applications and the benefits of edge over cloud computing. & \checkmark & \checkmark & \checkmark & \checkmark & \usym{2717} \\
\midrule
\cite{greco2020trends} & Discusses the role of IoT in the shift towards AI at the edge in  healthcare. & \usym{2717} & \usym{2717} & \usym{2717} & \checkmark & \checkmark \\
\midrule
\cite{liu2022bringing} & Reviews deep learning for edge AI, including model optimisation and potential future work. & \checkmark & \checkmark & \checkmark & \checkmark & \usym{2717} \\
\midrule
\cite{mendez2022edge} & Investigates edge intelligence, its motivations, challenges, and prospective evolution. & \checkmark & \usym{2717} & \checkmark & \checkmark & \usym{2717} \\
\midrule
\cite{murshed2021machine} & Details machine learning implementation at the edge, focusing on practical operational aspects. & \checkmark & \checkmark & \checkmark & \checkmark  & \usym{2717} \\
\midrule
\cite{wang2020deep} & Surveys deep learning in edge computing smart applications. & \usym{2717} & \usym{2717} & \usym{2717} & \checkmark & \usym{2717} \\
\midrule
\cite{wang2020convergence} & Explores the synergy of edge computing and deep learning, addressing custom frameworks and future research. & \checkmark & \usym{2717} & \checkmark & \checkmark & \usym{2717} \\
\midrule
Ours & Examines the deployment of computer vision and deep learning in edge computing with an emphasis on lightweight models and model compression, detailing advances, application, and enhancements in medical diagnostics and care. & \checkmark & \checkmark & \checkmark & \checkmark & \checkmark \\
\bottomrule
\end{tabular}
}
\vspace{0.5\baselineskip}
\caption{Related surveys on edge computing and deep learning. For each reference, the table summarises their focus areas and limitations, and indicates whether they discuss hardware, lightweight models, compression methods, computer vision applications, and medical applications.}
\label{fig:survey_summary}
\end{table*}

\section{Edge Computing Overview}
\label{sec:overview}
Edge computing is an emerging computing paradigm that has attracted a great deal of attention in recent years \citep{chen2019deep,mendez2022edge}. The proliferation of the Internet of Things (IoT), 5G, and other cutting-edge technologies has caused a rapid increase in data generation and consumption. In this setting, edge computing emerges as an effective strategy to handle the deluge of distributed data \citep{shi2016edge}. Unlike conventional cloud computing, edge computing brings data processing closer to its origin, thus achieving reduced latency and increased processing efficiency \citep{cao2020overview}. This approach is particularly relevant for latency-sensitive applications. For instance, autonomous vehicles require real-time data processing to make split-second decisions, ensuring safety and efficiency. Similarly, real-time patient monitoring systems in intensive care units rely on low-latency processing to provide immediate feedback and alerts. Furthermore, facilitating local processing on edge devices helps reduce data transmission and storage costs, simultaneously alleviating the computational burden on cloud infrastructures \citep{murshed2021machine}. In this section, we provide a succinct overview of salient edge computing techniques and exemplars, underscoring the paramount advantages of edge computing, especially in the realm of medical imaging, where these advantages are particularly pronounced.
\subsection{Edge Computing Related Paradigms and Terminologies}
First we delve into various paradigms and terminologies relevant to edge computing, providing a foundational understanding of the architectural frameworks and methodologies that characterise this computing paradigm. This paves the way for a deeper exploration of model compression, computer vision, and medical diagnostics in subsequent sections.

\bigskip
{\textbf{Cloudlets} \citep{babar2021cloudlet} are small-scale cloud service hubs located at the edge of the internet, closer to end-users and devices. They serve as a bridge between large cloud data centres and edge devices through small-scale cloud servers, offering reduced latency, improved bandwidth efficiency, and localised computing environments. Cloudlets can be seen as ``mini-clouds'' that provide a subset of the services of cloud data centres, with the added advantage of geographical proximity to end-users.

\bigskip
\textbf{Fog computing} \citep{bonomi2012fog} extends the functionalities of cloud computing through a decentralised architecture that provides localised data processing, storage, and analysis at the network edge. Implemented through a distributed network of nodes, which may include routers, gateways, and other networking devices with computing and storage capabilities, it offers a more distributed approach to deliver a wider range of services. Fog computing addresses the bandwidth limitations of traditional cloud architectures and complements Cloudlets within the overall framework of edge computing.

\bigskip
\textbf{Mobile (multi-access) edge computing (MEC)} facilitates real-time data processing at the network edge, reducing latency and bandwidth use. By bringing computational resources closer to mobile devices, MEC enhances user experiences and supports applications that require real-time processing. A recent review article \citep{shahzadi2017multi} dives into the architecture and applications of MEC, offering insight into its role in evolving network infrastructures.

\bigskip
{\textbf{Edge deep learning (Edge DL)} uses edge computing resources to perform deep learning inference and training tasks closer to data sources. This approach addresses the computational and latency challenges associated with centralised cloud-based deep learning frameworks. A recent survey \citep{wang2020convergence} provides a comprehensive overview of edge deep learning from a broad perspective, discussing the challenges and opportunities in this emerging field.

\bigskip
{\textbf{Edge analytics} is an approach to data analysis performed on data at the source of generation, such as sensors and other edge devices, rather than sending the data back to a centralised data store. This reduces latency and bandwidth usage while enabling real-time insight. Various issues, challenges, opportunities, promises, and future directions of edge analytics have recently been discussed in a detailed review \citep{nayak2022review}.

\bigskip
{\textbf{Cloud-edge collaboration} involves the integration of cloud computing and edge computing to take advantage of the strengths of both paradigms. This collaboration enables efficient data processing, analytics, and storage, optimises resource utilisation, and improves application performance. A recent comprehensive survey \citep{yao2022edge} delves into the various aspects of cloud-edge collaboration, including frameworks and architectures that facilitate seamless interaction between cloud and edge resources.

\bigskip
{\textbf{Edge model training} facilitates the creation and enhancement of machine learning (ML) models directly on edge devices. Examples include EdgeMove \citep{dong2023edgemove}, a scheme designed to accelerate model training between edge devices and servers by minimising communication overheads. Federated learning (FL), on the other hand, allows for model training on local data while preserving data privacy, albeit at the cost of computational and communication resources \citep{brecko2022federated,abreha2022federated}. Furthermore, distributed model training (DMT) optimises the distribution of training data among edge nodes, thus enhancing the efficiency of model training and the use of network resources \citep{hu2020distributed}.

\bigskip
{\textbf{Edge model deployment} is crucial for real-time processing. IBM underscores the necessity of a model management system to efficiently handle various models in edge computing.\footnote{IBM: IBM Edge Computing Solutions. \url{https://www.ibm.com/edge-computing}} Similarly, Microsoft Azure extends ML inference from the cloud to on-premise or edge scenarios through Azure Stack Edge, ensuring seamless deployment and operation of models at the edge.\footnote{Microsoft: Azure Stack Edge. \url{https://azure.microsoft.com/en-au/products/azure-stack/edge}} AdaptiveNet \citep{wen2023adaptivenet} enables post-deployment neural architecture adaptation for diverse edge environments to ensure stable service quality. The simplification of DL model deployment at the edge is further promoted by NVIDIA's Triton Inference Server, which standardises the deployment process across various devices.\footnote{NVIDIA: Simplifying AI Model Deployment at the Edge with NVIDIA Triton Inference Server. \url{https://developer.nvidia.com/blog/simplifying-ai-model-deployment-at-the-edge-with-triton-inference-server/}}
 
\bigskip
{\textbf{Edge model inference} is the execution of trained ML models on edge devices to deduce insights from new data \citep{jiang2018efficient}. This is crucial for real-time or near-real-time analytics, especially in scenarios where low latency and reduced data transmission are vital. The effectiveness of this approach hinges on the ability to execute complex computational tasks efficiently within the limited resource confines of edge devices. Accordingly, to optimise the performance of model inference, strategies such as model compression \citep{he2017channel,han2015deep,matsubara2022supervised,chen2022update,wang2020context} and adopting lightweight models \citep{mauri2022lightweight,koonce2021mobilenetv3,wang2021lightweight,paluru2021anam} are frequently implemented.

\subsection{Advantages of Edge Computing}
Edge computing, characterised by its decentralised data processing, represents a significant change from traditional cloud-centric models \citep{shi2016edge}. This paradigm is gaining prominence due to its potential impact across various domains, especially in environments with stringent requirements for real-time processing, data security, and bandwidth efficiency \citep{murshed2021machine}. The adoption of edge computing can bring about substantial improvements in processing speeds, security, network utilisation, scalability, operational reliability, and cost efficiency \citep{cao2020overview,mendez2022edge}. These attributes are particularly relevant in sectors such as healthcare, where fast, secure and reliable data processing is paramount. The following survey delineates these advantages, underscoring their implications and applications in medical and related fields.

\bigskip
{\textbf{Real-time data processing:} The primary advantage of edge computing is its ability to process data in real time at the data source. This is critical for applications that require swift responses. In healthcare, for example, immediate diagnostics and analysis in ambulances or intensive care units using edge devices can lead to interventions crucial for patient survival. Similarly, in medical imaging, processing images from MRI or CT scans directly on edge devices can accelerate the diagnostic process, potentially enabling quicker treatment planning. This rapid processing capability directly influences patient treatment outcomes, especially in time-sensitive scenarios \citep{sait2019mobile,liu2023edgemednet,chen2019multi,lingappa2022active}.

\bigskip
{\textbf{Data privacy and security:} In an era where data breaches are becoming more common, edge computing can offer a more secure alternative to traditional cloud computing. By processing and storing data locally, it helps reduce the exposure of data during network transmission. This is particularly important in industries with high data sensitivity, such as healthcare. Patient health information, medical histories, and diagnostic images contain highly sensitive information. Edge computing ensures that these data remain within the confines of the hospital's local network, significantly reducing breach risks and ensuring compliance with stringent health data regulations like the Health Insurance Portability and Accountability Act (HIPAA). Furthermore, local processing allows for quicker detection and response to potential security threats, enhancing the overall security framework. This local processing approach provides a reliable and robust framework for managing sensitive health data, safeguarding patient privacy and maintaining trust in healthcare systems \citep{cao2023privacy,singh2021securing,alwakeel2021overview}.

\bigskip
{\textbf{Bandwidth efficiency:} Edge computing can help mitigate network congestion by processing data locally, reducing the bandwidth required for data transmission. This efficiency is crucial in industries where large data files are common, such as healthcare. For example, transmitting high-resolution medical images, such as digital histopathology slides, can consume considerable bandwidth. With edge computing, these images are processed locally, eliminating the need for a constant, high-volume data transfer to the cloud. This not only conserves network bandwidth, but also ensures faster, more reliable medical imaging services, leading to more efficient healthcare delivery \citep{dong2020edge,dave2021benefits,zheng2021mobile}.

\bigskip
{\textbf{Scalability and flexibility:} Edge computing's scalability enables organisations to expand their computing capabilities as needed without significant infrastructure overhauls. This scalability is essential for industries experiencing rapid growth or facing fluctuating demands. In healthcare, the ability to handle varying patient loads and integrate new medical technologies is crucial. The flexibility of edge computing is evident in scenarios like remote monitoring and care, supporting an increase in chronic disease treatment and enhancing healthcare access, especially in remote or rural areas. Processing data from IoT, e-health devices, and wearable technology at the edge allows healthcare providers to offer more effective continuous care and management of chronic diseases \citep{sun2020edge,oueida2018edge}.

\bigskip
{\textbf{Reliability and accessibility:} Edge computing can provide more reliable data processing capabilities, especially in environments with limited or inconsistent internet connectivity. This reliability is crucial in many sectors, particularly in healthcare, where consistent access to patient data and medical applications can be lifesaving. For example, in remote or rural healthcare settings, reliable access to medical records and diagnostic tools is essential for patient care. Edge computing enables this by allowing local data processing and analysis, ensuring uninterrupted healthcare services despite connectivity issues. This local processing capability is particularly valuable in telemedicine and remote patient monitoring, where reliable data access is key to providing quality care to patients in remote locations \citep{srivastava2023federated,ullah2021multi,abdellatif2021medge}.

\bigskip
{\textbf{Cost-effectiveness:} Finally, edge computing offers cost-effective solutions by reducing the need for extensive data transmission and the dependency on centralised cloud storage. Particularly in the healthcare sector, this approach can save costs associated with managing and storing large amounts of medical data on cloud servers. Cloud services typically charge based on usage, including storage space and data transmission volumes. By keeping an appropriate amount of data storage and processing at the edge, hospitals and healthcare providers can establish more economical and efficient data processing schemes. In addition, economies of scale make the costs of deploying and maintaining edge computing hardware continuously decrease. Moreover, edge computing can be integrated with cloud computing to balance the initial investment and maintenance costs of local devices. As edge computing improves operational efficiency and response speed, healthcare professionals can gain faster access to critical information, make timely treatment decisions, and reduce treatment delays. In emergency situations, this rapid response capability can save lives and reduce long-term treatment costs. In summary, this localised data processing method  
can help optimise the use of IT resources in healthcare, making healthcare services more efficient and cost-effective \citep{d2019combining,gu2022low,tang2023efficient,mahenge2019mobile,jebadurai2021green}.

\subsection{Disadvantages of Edge Computing}

\bigskip
\textbf{Limited computational resources:} Compared to centralised cloud servers, edge devices often have limited resources, including processing power, memory, and storage space. These limitations can hinder the ability to run complex and computationally intensive deep learning models on edge devices \citep{varghese2020survey, murshed2021machine}. This issue is particularly significant in medical diagnostics, where the data involved is typically very large. For instance, medical imaging data such as MRI or CT scans often have extremely high resolution and detail, resulting in large data volumes. Moreover, due to the limited computational power, running large deep learning models for medical image analysis on edge devices may not be feasible. This limitation necessitates the use of data and model compression techniques or lightweight models, which may compromise the accuracy and effectiveness of the diagnostic process \citep{ou2021polyp,goceri2021diagnosis}.

\bigskip
\textbf{Heterogeneity of hardware and software:} Edge computing devices come in a wide variety, with significant differences in performance and capabilities across different platforms, as well as varying requirements for software compatibility. This heterogeneity increases the complexity of development and maintenance, requiring specialised optimisation and adaptation strategies to ensure efficient operation across different devices. For example, while field-programmable gate arrays (FPGAs) offer flexibility and lower power consumption, their programming and optimisation are challenging. Although there are solutions to mitigate these issues, such as Open Neural Network Exchange (ONNX),\footnote{ONNX. \url{https://onnx.ai/}} ExecuTorch,\footnote{ExecuTorch. \url{https://docs.pytorch.org/executorch/stable}} and TensorFlow Lite,\footnote{TensorFlow Lite. \url{https://www.tensorflow.org/lite}} which facilitate model conversion and optimisation across different hardware platforms, these solutions still face challenges in medical scenarios. Specifically, medical image data not only have high resolution but also vary in type, and devices like FPGAs and application-specific integrated circuits (ASICs) may encounter software compatibility and driver issues when processing these images, further increasing the difficulty and cost of application development and implementation \citep{solanki2021performance,gtifa2023integrating,tabassum2023brain}.

\bigskip
\textbf{Standardization issues:} Due to hardware differences and lack of uniform standards among edge devices, inconsistent computational results may be produced, affecting the overall system's accuracy and reliability \citep{feng2022benchmark}. In medical environments, ensuring the standardisation of edge device hardware and software protocols is crucial to guarantee the consistency and accuracy of diagnostic results, as well as to comply with regulatory requirements. Standardising edge devices not only helps improve system integration efficiency but also ensures the accuracy and reliability of diagnostic outcomes, thereby enhancing the overall quality of medical diagnostics. Additionally, standardisation can reduce development and maintenance costs, enabling healthcare institutions to deploy and manage edge computing devices more effectively. It also simplifies compliance with medical data regulations like HIPAA and General Data Protection Regulation (GDPR), further improving the efficiency and quality of medical services.

\bigskip

\section{Edge Computing Devices}
\label{sec:devices}
Edge computing devices serve as the core components of edge computing frameworks and play a crucial role in their implementation. The performance and diverse functionality of these devices directly impact the feasibility and efficiency of edge computing solutions. Different edge computing scenarios require various types of edge devices, ranging from simple microcontrollers to computing platforms capable of processing large deep learning models. These devices together form a complex and feature-rich technological ecosystem. Therefore, a thorough understanding of the classification, specifications, application scenarios, and cost-effectiveness is essential for the design and implementation of edge computing solutions in medical diagnostics.

In this section, we present a classification of edge computing devices based on their computational capabilities, dividing them into low-end, mid-range, and high-performance devices. Each category of devices is designed for specific application scenarios, from executing basic data preprocessing tasks to supporting complex deep learning model training. Subsequently, we introduce some typical devices and discuss their performance and purposes. We emphasise the importance of edge device diversity and demonstrate how to fully leverage the potential of edge computing by selecting the appropriate devices.

\begin{figure}[!t]
    \centering
    \includegraphics[width=0.95\textwidth]{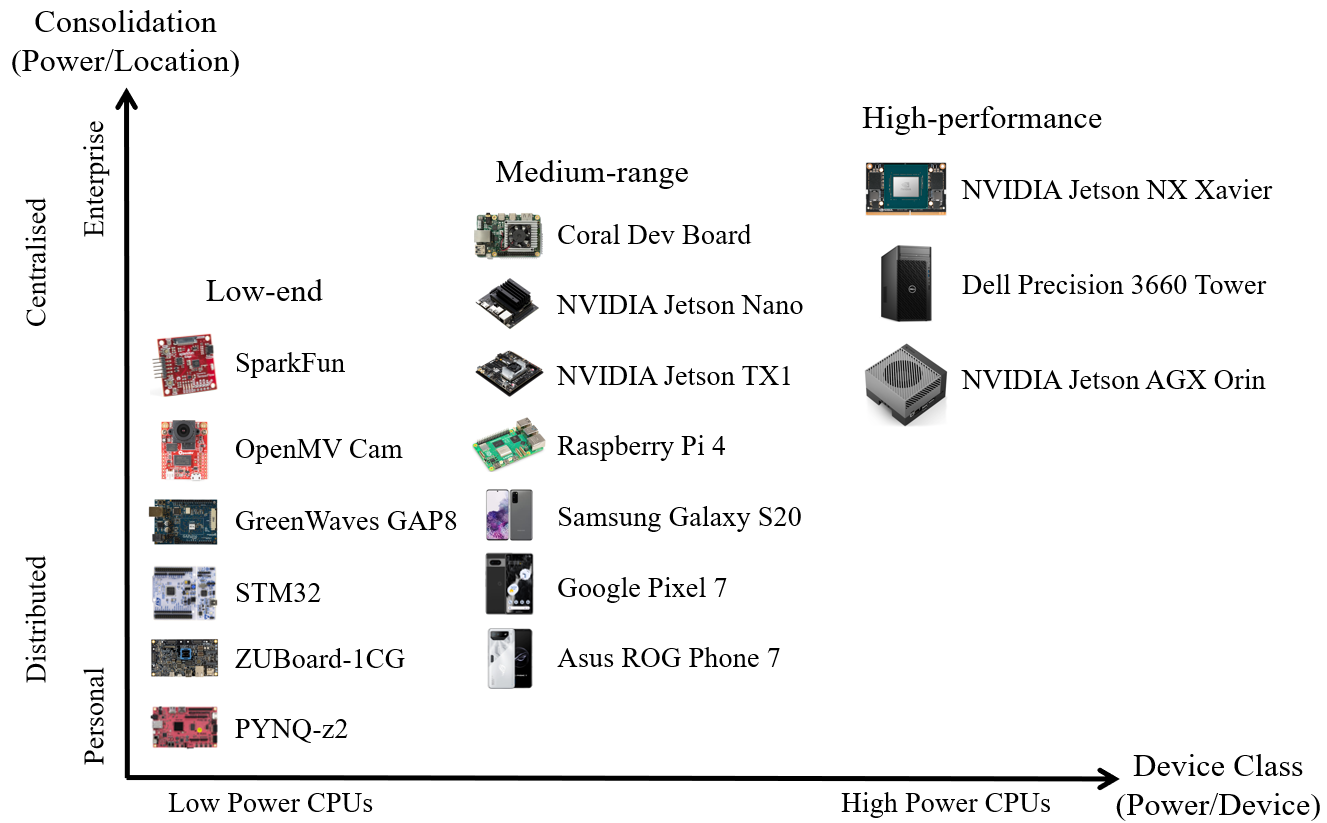}
    % \vspace{-1.5\baselineskip}
    \caption{Three categories of edge computing devices.}
    \label{fig:edge_computing}
\end{figure}

\subsection{Edge Computing Device Categories}
We propose to categorise edge computing devices according to their computational capabilities into low-end, medium-range, and high-performance devices (Fig.~\ref{fig:edge_computing}) and briefly discuss each category.

\bigskip

\textbf{Low-end edge devices} are characterised by their limited computational resources and energy efficiency, making them suitable for lightweight applications \citep{dutta2021tinyml}. These devices typically lack dedicated multiple graphics processing units (GPUs) or advanced central processing units (CPUs). Due to their computational constraints, they are more suited for inference tasks than for training DL models. These devices usually have power efficient processors, such as the ARM Cortex M series\footnote{ACM: Cortex M Series. \url{https://www.arm.com/products/silicon-ip-cpu/cortex-m/cortex-m4}} or the lower-end Cortex A series,\footnote{ACM: Cortex A Series. \url{https://www.arm.com/products/silicon-ip-cpu/cortex-a/cortex-a720}} and give priority to minimising power usage and achieving small form factors. Memory capacities are quite constrained, often ranging from a few hundred kilobytes (kB) to a few megabytes (MB) and may not include advanced accelerators. Therefore, these devices are highly suitable for performing basic edge computing tasks in environments with limited resources. Sensor data preprocessing, fundamental DL inference, and low-complexity IoT applications are a few examples \citep{nunez2021energy,shen2022big,dutta2021tinyml}. Due to their energy efficiency and affordability, these devices are also well suited for projects that have limited financial resources or require installation in remote locations with a lack of power supply \citep{ray2022review,schizas2022tinyml}. This makes them particularly valuable in lightweight medical diagnostic applications. For example, in wearable health monitoring systems, low-end devices can effectively handle sensor data preprocessing, enabling real-time health tracking in remote or resource-constrained environments. Their high energy efficiency and affordability make them ideal for tasks such as basic medical data acquisition and simple inference.

\bigskip

\textbf{Medium-range edge devices} offer a balanced level of computational power, making them capable of handling more advanced tasks compared to low-end devices. They often come with multicore processors and some level of GPU acceleration, which allows them to perform more computationally intensive tasks \citep{jolles2021broad,zhao2015exploring,mao2017survey}. Although they may lack the substantial computational resources required to train DL models, their enhanced processing capabilities make them well suited for more advanced inference tasks. Unlike low-end devices, these devices typically have memory capacities ranging from one to several gigabytes (GB). For performance enhancement, certain devices in this category incorporate specialised AI accelerators \citep{sipola2022artificial}. These devices exhibit versatility and are well suited for a wide array of AI applications, including image recognition \citep{chavan2021plant,sati2021face,kang2018joint}, object detection \citep{nazeer2022real,mittapalli2023deep,kaymak2018implementation}, and the creation of compact DL models \citep{momin2023lightweight,heo2020lightweight,gonzalez2021disease} for use in robotics \citep{alexey2021autonomous,tang2017real,mathe2022review}, smart cameras \citep{kyrkou2020yolopeds,karaman2021development}, and other edge computing scenarios requiring a moderate amount of computational capability \citep{mittal2019survey,miori2017platform,lertsinsrubtavee2017picasso,zhao2018ecrt}. In the medical field, medium-range devices are particularly well-suited for deploying AI models that perform image-based diagnostics, such as automated pathology screening. Devices such as the NVIDIA Jetson Nano, Raspberry Pi, and smartphones have already been utilised in portable medical imaging devices to execute pretrained diagnostic models in real-time, striking a good balance between performance and power consumption. \citep{wang2019recognition,raghavan2020initial,goceri2021diagnosis,paluru2021anam,shrivastava2023skin}

\bigskip

\textbf{High-performance edge devices} are designed to handle complex computational tasks, including DL model training. They feature robust architectures with powerful CPUs, GPUs, and significant memory resources. Unlike low-end and medium-range devices, high-performance edge devices can handle training and inference tasks effectively, making them suitable for scenarios that require high computational power. The memory capacities of these devices are substantial, ranging from several GB to terabytes (TB), allowing them to efficiently process intricate models and datasets. These devices provide exceptional performance in applications that require extensive AI tasks, such as training DL models, handling sophisticated computer vision tasks, and executing advanced AI applications \citep{kirillov2023segment,ma2024segment,ho2020denoising}. They are suitable for use in sophisticated robots \citep{krupnik2023fine,wang2023dexgraspnet,makoviychuk2021isaac} and data centres where top-notch performance is crucial \citep{angus2022real,oro2011real,barisoni2020digital,shen2019nexus}. These devices can handle large medical images (such as MRI or CT scans) and perform in-depth image analysis in real-time. Enabling local training of deep learning models, they can leverage test-time domain adaptation and continual learning to accommodate the ever-growing medical data, ensuring higher diagnostic accuracy and reliability. This is crucial in handling complex medical scenarios, especially when faced with constantly evolving patient data or environments.

\bigskip

This discussion underscores the fact that medium-range and low-end edge devices are typically more constrained in their training capabilities because of limited computational resources. However, their compact size and portability make them suitable for deployment in space-constrained \citep{dutta2021tinyml,xie2018video,lachhab2023deep,achakir2023automated} or mobile scenarios \citep{shahzadi2017multi,abbas2017mobile,mach2017mobile}. Moreover, although they may not be able to effectively handle training of DL models, they are generally capable of achieving (near) real-time inference, a critical aspect for certain computer vision applications \citep{nazeer2022real,tang2017real,ray2022review}. The ability to process data locally also enhances privacy and security by minimising data transmission, thus reducing the risk of data leakage or tampering \citep{cao2023privacy,singh2021securing,alwakeel2021overview,ranaweera2021survey,zhang2018data,ali2021multi}.

\begin{table}[!t]
\renewcommand{\arraystretch}{1.1}
\resizebox{\textwidth}{!}{%
\begin{tabular}{lllllllcc}
\toprule
\bf Type & \bf Devices & \bf Operating System & \bf Framework & \bf Core Language & \bf CPU & \bf Accelerator & \bf Memory & \bf Price (USD) \\
\midrule
 & SparkFun Apollo3 Blue & - & TensorFlow & Python & ARM Cortex-M4F & - & 384 KB & $\sim$\$20 \\
\cmidrule{2-9}
 & OpenMV Cam H7 R2 & MicroPython  & - & MicroPython & ARM Cortex-M7 & - & 1 MB & $\sim$\$80 \\
\cmidrule{2-9}
 & Eta Compute ECM3532 & - & - & C/C++ & \begin{tabular}[c]{@{}l@{}}ARM Cortex-M3 \end{tabular}& - & 256 KB & $\sim$\$60 \\
\cmidrule{2-9}
Low-End & GreenWaves GAP8 & \begin{tabular}[c]{@{}l@{}}FreeRTOS,\\ PULP OS,\\ PMSIS\end{tabular} & \begin{tabular}[c]{@{}l@{}}TensorFlow, Pytorch, 
 \\ Keras, ONNX \end{tabular} & Python, C/C++ & Nona-Core RISC-V & - & 16 MB & $\sim$\$100 \\
\cmidrule{2-9}
 & BeagleBone AI & Linux &  \begin{tabular}[c]{@{}l@{}}TensorFlow, Pytorch, 
 \\ Caffe, ONNX \end{tabular} & Python, C/C++ & \begin{tabular}[c]{@{}l@{}}Dual ARM Cortex-A15 + \\ Dual ARM Cortex-M4 \end{tabular} & - & 1 GB & $\sim$\$300 \\
\cmidrule{2-9}
 & STM32MP157F-DK2 & Linux  & ONNX & Python, C/C++ & \begin{tabular}[c]{@{}l@{}}Dual Cortex‑A7 + \\ ARM Cortex‑M4 \end{tabular}& - & 4 GB & $\sim$\$200 \\
\cmidrule{2-9}
 & PYNQ-Z2 & Linux & \begin{tabular}[c]{@{}l@{}}TensorFlow, Pytorch, 
 \\ ONNX \end{tabular} & Python &  \begin{tabular}[c]{@{}l@{}}Dual ARM Cortex-A9 \end{tabular} & - & 512 MB & $\sim$\$120 \\
 \cmidrule{2-9}
 & ZUBoard-1CG & Linux & \begin{tabular}[c]{@{}l@{}}TensorFlow, Pytorch, 
 \\ ONNX \end{tabular} & Python & \begin{tabular}[c]{@{}l@{}}ARM Cortex-A53 +\\  ARM Cortex-R5F \end{tabular}& - & 1 GB & $\sim$\$150 \\
\midrule
 & Google Coral Dev Board & Linux & TensorFlow & Python, C++ &\begin{tabular}[c]{@{}l@{}} Quad ARM Cortex-A53 + \\ARM Cortex-M4F\end{tabular} & \begin{tabular}[c]{@{}l@{}}GC7000 Lite Graphics + \\ Google Edge TPU\end{tabular} & \begin{tabular}[c]{@{}l@{}}1 GB / \\4 GB \end{tabular}& $\sim$\$130 \\
\cmidrule{2-9}
 & NVIDIA Jetson Nano & Linux & \begin{tabular}[c]{@{}l@{}}TensorFlow, Pytorch, Keras,\\ JAX, PaddlePaddle, MXNet, \\ Caffe, ONNX \end{tabular} & Python & Quad ARM Cortex-A57 & NVIDIA Maxwell 128 CUDA Cores & 4 GB & $\sim$\$150 \\
\cmidrule{2-9}
 & NVIDIA Jetson TX1 & Linux & \begin{tabular}[c]{@{}l@{}}TensorFlow, Pytorch, Keras,\\ JAX, PaddlePaddle, MXNet, \\ Caffe, ONNX \end{tabular} & Python & Quad ARM Cortex-A57 & NVIDIA Maxwell 256 CUDA Cores & 4 GB & $\sim$\$200 \\
\cmidrule{2-9}
Medium-Range & Raspberry Pi 5 & Linux & \begin{tabular}[c]{@{}l@{}}TensorFlow, Pytorch,\\ Caffe, Keras \end{tabular}& Python, C++ & Quad ARM Cortex-A76 & Broadcom VideoCore VII & \begin{tabular}[c]{@{}l@{}} 2 GB / \\ 4 GB / \\ 8 GB \end{tabular} & $\sim$\$80 \\
\cmidrule{2-9}
 & NVIDIA Jetson TX2 & Linux & \begin{tabular}[c]{@{}l@{}}TensorFlow, Pytorch, Keras,\\ JAX, PaddlePaddle, MXNet, \\ Caffe, ONNX \end{tabular} & Python & \begin{tabular}[c]{@{}l@{}}Quad ARM Cortex-A57 + \\NVIDIA Dual Denver 2\end{tabular}& NVIDIA Pascal 256 CUDA Cores & 8 GB & $\sim$\$250 \\
\cmidrule{2-9}
 & Samsung Galaxy S20 & Android & \begin{tabular}[c]{@{}l@{}}TensorFlow, Pytorch, Caffe, \\Keras, ONNX \end{tabular} & Java, Python & Qualcomm SM8250 & Qualcomm Adreno 650 & 8 GB & $\sim$\$250 \\
\cmidrule{2-9}
 & Google Pixel 7 & Android & \begin{tabular}[c]{@{}l@{}}TensorFlow, Pytorch, Caffe, \\Keras, ONNX \end{tabular} & Java, Kotlin, Python & Google Tensor G2 & Mali-G710 MP7 & 8 GB & $\sim$\$350 \\
\cmidrule{2-9}
 & Asus ROG Phone 7 & Android &\begin{tabular}[c]{@{}l@{}}TensorFlow, Pytorch, Caffe, \\Keras, ONNX \end{tabular} & Java Python & Qualcomm SM8550 & Qualcomm Adreno 740 & 16 GB & $\sim$\$800 \\
\midrule
 & Jetson NX Xavier & Linux & \begin{tabular}[c]{@{}l@{}}TensorFlow, Pytorch, Keras,\\ JAX, PaddlePaddle, MXNet, \\ Caffe, ONNX \end{tabular} & Python & \begin{tabular}[c]{@{}l@{}}6-Core NVIDIA Carmel  \end{tabular}& NVIDIA Volta 384 CUDA Cores & 16 GB & $\sim$\$700 \\
\cmidrule{2-9}
% High-Performance & Jetson AGX Xavier & Linux & \begin{tabular}[c]{@{}l@{}}TensorFlow, Pytorch, Keras,\\  JAX, PaddlePaddle, MXNet\end{tabular} & Python & \begin{tabular}[c]{@{}l@{}}6-Core NVIDIA Carmel Arm \\ 64-bit 6MB L2 + 4MB L3 \end{tabular}& NVIDIA Volta 512 CUDA Core & 32 GB & \$1,299 \\
\cmidrule{2-9}
High-Performance & Jetson AGX Orin & Linux & \begin{tabular}[c]{@{}l@{}}TensorFlow, Pytorch, Keras,\\ JAX, PaddlePaddle, MXNet, \\ Caffe, ONNX \end{tabular} & Python & \begin{tabular}[c]{@{}l@{}}12-Core Arm Cortex-A78AE \end{tabular}& NVIDIA Ampere 2048 CUDA Cores & 64 GB & $\sim$\$1,000 \\
\cmidrule{2-9}
 & Dell Precision 3660 Tower & Linux, Windows & \begin{tabular}[c]{@{}l@{}}TensorFlow, Pytorch, Keras,\\ JAX, PaddlePaddle, MXNet, \\ Caffe, ONNX \end{tabular} & Python & Intel Core i9-13900K & NVIDIA RTX A4000 & 32 GB & $\sim$\$3,000 \\

\bottomrule
\end{tabular}}
% \vspace{-0.2\baselineskip}
\caption{Examples of edge computing devices categorised by computational capabilities (Prices are approximate and may vary by region and retailer).}
\label{tab:edge_devices}
\end{table}

\subsection{Detailed Overview of Edge Computing Devices}

Following the above categorisation, we now present concrete examples of edge computing devices across the spectrum of computational capabilities, from low-end devices to high-performance devices (Table \ref{tab:edge_devices}). We summarise the fundamental hardware specifications of these devices, including the CPU, accelerator, and memory attributes, providing a clearer perspective on their computational capacities and potential use cases.

\bigskip

{\bf Low-end edge computing devices} such as the SparkFun Edge,\footnote{SparkFun Edge. \url{https://github.com/sparkfun/SparkFun\_Edge}}\label{SparkFun} STM32 microcontroller series,\footnote{ST: STM32. \url{https://www.st.com/en/microcontrollers-microprocessors/stm32-arm-cortex-mpus.html}} XUP PYNQ-Z2 board,\footnote{AMD: XUP PYNQ-Z2. \url{https://www.xilinx.com/support/university/xup-boards/XUPPYNQ-Z2.html}} and Intel Neural Compute Stick 2 (NCS2),\footnote{Intel: Neural Compute Stick 2. \url{https://www.intel.com/content/www/us/en/developer/articles/tool/neural-compute-stick.html}} have shown significant multifunctionality and applicability. The STM32 series of microcontrollers is widely used in embedded systems, especially in the TinyML (tiny machine learning) application domain \citep{cum2022neural,schizas2022tinyml,dutta2021tinyml}. The SparkFun Edge board optimised for ultra-low power consumption is ideal for edge DL applications where efficiency is paramount. Built around the ARM Cortex-M4F\footnote{ARM: Cortex-M4. \url{https://developer.arm.com/Processors/Cortex-M4}} and integrated with TensorFlow Lite for microcontrollers, it supports DL tasks directly on the device. Its on-board sensors and bluetooth connectivity make it perfect for IoT and wearable devices, enabling real-time data processing and communication in compact, power-sensitive projects.

The STM32 X-CUBE-AI extension package\footnote{ST: AI Expansion Pack for STM32Cube. \url{https://www.st.com/en/embedded-software/x-cube-ai.html}} further expands the capabilities of STM32, allowing users to easily deploy trained neural networks to microcontrollers. This feature makes the STM32 an ideal choice for tasks such as image recognition and video analysis. In industrial automation and smart home systems, STM32 combined with deep learning algorithms can enhance efficiency and accuracy, optimising monitoring and control processes. STM32 also supports various communication protocols, such as SPI, I2C, UART, and CAN, making it an ideal choice for integrating various peripherals and modules. Additionally, STM32 microcontrollers support multiple programming environments, including Eclipse-based STM32CubeIDE and Arduino, offering significant flexibility and ease of use.

Leveraging the STM32H743VI microcontroller with a Cortex-M7 processor, the OpenMV Cam H7 R2\footnote{OpenMV Cam H7 R2. \url{https://openmv.io/products/openmv-cam-h7-r2}} is a pivotal tool in edge-based machine vision applications. Its design facilitates real-time image processing, enabling tasks such as face recognition and object tracking. The device integrates a camera module for direct image capture and supports multiple peripherals for expanded functionality. Compatible with the OpenMV IDE, it offers a Python-based development environment, which streamlines the deployment of complex vision algorithms. 

Another commonly used edge device is the AUP PYNQ-Z2 board.\footnote{AUP PYNQ-Z2. \url{https://www.amd.com/en/corporate/university-program/aup-boards/pynq-z2.html}} PYNQ-Z2, based on the Xilinx Zynq SoC,\footnote{Xilinx Zynq SoC. \url{https://www.xilinx.com/products/silicon-devices/soc/zynq-7000.html}} combines the powerful functionality of an ARM processor with the flexibility of a field-programmable gate array (FPGA). It provides significant hardware acceleration for advanced image processing and complex signal processing. A notable feature of PYNQ-Z2 is its support for AMD Vitis AI,\footnote{Vitis AI. \url{https://www.xilinx.com/products/design-tools/vitis/vitis-ai.html}} designed to accelerate AI inference on FPGAs and other programmable logic devices. This allows optimising and deploying trained DL models to the PYNQ-Z2 board, enabling efficient edge AI applications. It makes PYNQ-Z2 particularly advantageous in AI projects that require custom hardware acceleration, such as real-time video analysis and intelligent sensor applications. Additionally, PYNQ-Z2 supports Python programming and the open source PYNQ framework, greatly simplifying the development process and making it an ideal choice for researchers and developers for rapid prototyping.

The Intel Neural Compute Stick 2 (NCS2) is focused on enhancing the DL inference capabilities of edge devices. Equipped with the Movidius Myriad X Vision Processing Unit (VPU),\footnote{Movidius Myriad X VPU. \url{https://www.intel.com/content/www/us/en/products/sku/125926/intel-movidius-myriad-x-vision-processing-unit-4gb/specifications.html}} it accelerates the execution of neural network models without significant increases in power consumption, which is crucial for applications requiring real-time image and video analysis, such as intelligent surveillance and automated detection systems.

Furthermore, the Eta Compute ECM3532\footnote{Eta Compute ECM3532. \url{https://media.digikey.com/pdf/Data\%20Sheets/Eta\%20Compute\%20PDFs/ECM3532_AI_Sensor_PB_1.0.pdf}} combines an ARM Cortex M3 processor, optimising energy efficiency and computational performance in edge AI applications. This dual-core approach enables the device to support advanced deep learning tasks, within the stringent power constraints typical of wearable and IoT devices. GreenWaves GAP8\footnote{GreenWaves GAP8. \url{https://greenwaves-technologies.com/low-power-processor}} stands out for its ultra-low power consumption and ability to perform embedded DL tasks efficiently on the edge. Utilising an 8-core computational cluster with a hardware accelerator, it is optimised for processing image and audio algorithms, including CNN inference, with exceptional energy efficiency. The BeagleBone AI,\footnote{BeagleBone AI. \url{https://www.beagleboard.org/boards/beaglebone-ai}} designed around the Texas Instruments AM5729 Sitara processor,\footnote{Texas Instruments AM5729. \url{https://www.ti.com/product/AM5729}} offers a platform for developers to explore AI integration in edge computing. It features a dual ARM Cortex-A15 processor,\footnote{ARM: Cortex-A15. \url{https://developer.arm.com/Processors/Cortex-A15}} supported by Embedded Vision Engines (EVEs) for DL, and extensive connectivity options including Gigabit Ethernet and WiFi. The ZUBoard 1CG,\footnote{ZUBoard 1CG. \url{https://www.tria-technologies.com/product/zuboard-1cg/}} serves as a versatile platform for edge computing applications such as embedded vision. Moreover, it offers high-speed storage and wireless connectivity options, satisfying the demanding requirements of industrial, healthcare, and multimedia applications.

\bigskip

{\bf Medium-range edge computing devices} include Google Coral Dev Board, a compact, powerful platform designed for Edge DL, which integrates Google's Edge Tensor Processing Unit (TPU) coprocessor\footnote{TensorFlow models on the Edge TPU. \url{https://coral.ai/docs/edgetpu/models-intro}} and is capable of performing fast DL inferencing on small form factor devices. This makes it ideal for prototyping AI products and solutions that require processing efficiency. The board supports TensorFlow Lite models, facilitating the development of AI applications such as healthcare, retail, and the smart home, by enabling local real-time processing of deep learning workloads \citep{imran2020embedded,winzig2022edge}.

The NVIDIA Jetson series,\footnote{NVIDIA Jetson Modules. \url{https://developer.nvidia.com/embedded/jetson-modules}} including Jetson Nano, TX1, and TX2, are key devices designed for different levels of edge computing needs. Jetson Nano, the base model in the series, is mainly suitable for lightweight machine vision and data processing tasks, such as simple object recognition and video stream processing. Its low power consumption and compact design make it an ideal choice for deployment in resource-constrained environments. Jetson TX1 and TX2 offer more powerful computing capabilities, suitable for applications that require higher image processing capacity, such as in augmented reality (AR) and virtual reality (VR), autonomous vehicle perception systems, drone navigation systems, chest CT, and dermatology detection \citep{mittal2019survey,cass2020nvidia,wang2019recognition,shrivastava2023skin,abubeker2023b2}.

The Raspberry Pi series,\footnote{Raspberry Pi. \url{https://www.raspberrypi.com}} a widely used single-board computer, is preferred for its cost-effectiveness, low power consumption, and robust community support. Despite its limited processing capabilities, it is suitable for simple applications, such as sensor data processing and lightweight image recognition tasks \citep{zhao2015exploring,jolles2021broad,kaymak2018implementation,paluru2021anam}.

Additionally, smartphones represent a typical example of mid-range edge devices, equipped with processors capable of efficiently running DL models for various tasks including image processing and video analysis. Integrated with dedicated GPUs, digital signal processors (DSPs), and DL accelerators such as neural processing units (NPUs), they offer excellent task processing capabilities and efficient energy management. The Qualcomm SM8250 and SM8550\footnote{Snapdragon 865 5G Mobile Platform. \url{https://www.qualcomm.com/products/mobile/snapdragon/smartphones/snapdragon-8-series-mobile-platforms/snapdragon-865-5g-mobile-platform}} are top-notch smartphone chipsets that can provide devices with advanced inference performance for DL models. In addition, they support 5G connectivity, enabling faster internet speeds and improved network performance. This facilitates effective support for distributed edge training and inference paradigms, such as federated learning.

Quectel's SC66 smart module\footnote{Quectel LTE SC66 series. \url{https://www.quectel.com/product/lte-sc66-smart-module-series}} is a multi-functional and widely applicable edge module. Equipped with Qualcomm's SDM660 chipset\footnote{Qualcomm-SDM660.\url{https://www.qualcomm.com/products/technology/processors/application-processors/sdm660}} and the Snapdragon Neural Processing Engine (SNPE),\footnote{SNP Engine. \url{https://docs.qualcomm.com/nav/home/index_SNPE.html?product=1601111740010412}} it is designed for high data rate, multimedia capabilities, and advanced deep learning-based use cases. SC66 supports various features, such as quick charge technology, making it highly suitable for industrial and consumer-grade applications. Combining high-speed wireless connectivity and an embedded Global Navigation Satellite System (GNSS) receiver, it is capable of serving a wide range of edge applications \citep{lestariningati2018mobile,sadique2013secure,mwansa2022augmented,cum2022neural}.

Furthermore, the Apple A16 Bionic chip,\footnote{Apple debuts iPhone 14 Pro and iPhone 14 Pro Max. \url{https://www.apple.com/au/newsroom/2022/09/apple-debuts-iphone-14-pro-and-iphone-14-pro-max/}} introduced in the iPhone 14 Pro models, is built on 4 nm process technology, offering improved performance and efficiency compared to its predecessors. It features a 6-core CPU with two high-performance and four efficiency cores, a 5-core GPU for improved graphics, and a 16-core neural engine for advanced DL tasks, doubling the DL capabilities to 17 trillion operations per second. This chip significantly boosts performance for real-time DL-based image and video processing, while optimising power consumption for extended battery life.

\bigskip

{\bf High performance edge computing devices} include the NVIDIA Jetson Xavier series\footnote{Jetson Xavier series. \url{https://www.nvidia.com/en-au/autonomous-machines/embedded-systems/jetson-xavier-series/}} and Jetson Orin series.\footnote{Jetson Orin series. \url{https://www.nvidia.com/en-au/autonomous-machines/embedded-systems/jetson-orin/}} The Jetson Xavier and Orin series uses the NVIDIA Volta and Ampere GPU architecture, featuring CUDA (compute unified device architecture) cores and Tensor cores, as well as larger memory capacity, all designed specifically for DL acceleration. These devices are suited for running more complex DL models and allow model fine-tuning during deployment \citep{bhardwaj2022unsupervised,kortli2022deep}. A performance comparison of the YOLOv3 model on NVIDIA Jetson Xavier NX, NVIDIA Jetson Nano, and Raspberry Pi 4 + NCS2 (Table \ref{tab:model_compare}) highlights the advantages of high-performance devices like the Jetson Xavier NX in terms of FPS and inference time, while also showcasing the balance of performance and energy efficiency offered by mid-range devices like the Jetson Nano.

% The Jetson Orin series offers an advanced level of performance, with more CUDA and Tensor cores, as well as larger memory capacity. 

\begin{table}[!t]
\renewcommand{\arraystretch}{1.1}
\color{black}
\resizebox{\textwidth}{!}{%
\begin{tabular}{llcccccc}
\toprule
\bf Model & \bf Devices & \bf Mean Confidence (\%) & \bf FPS & \bf CPU Usage (\%) & \bf Memory Usage (GB) & \bf Energy Consumption (W) & \bf Inference Time (s) \\
\midrule
 & Raspberry Pi 4 + NCS2 & 99.3 & 2.5 & 4.3 & 0.33 & 6.0 & 690  \\
\cmidrule{2-8}
\cmidrule{2-8}
YOLOv3 & Jetson Nano & 99.7 & 1.7 & 26.5 & 1.21 & 7.9 & 967  \\
\cmidrule{2-8}
 & Jetson Xavier NX & 99.7 & 6.1 & 22.5 & 1.51 & 15.2 & 256  \\
\bottomrule
\end{tabular}}
% \vspace{-0.5\baselineskip}
\caption{Performance comparison of the YOLOv3 model on Raspberry Pi 4 + NCS2, Jetson Nano, and Jetson Xavier NX. The tests were conducted on video data consisting of 1596 frames with a frame size of 768×436 \citep{feng2022benchmark}.}
\label{tab:model_compare}
\end{table}

The Dell Precision 3660 Tower\footnote{Dell Precision 3660 Tower Workstation \url{https://www.dell.com/en-au/shop/dell-desktop-computers/precision-3660-tower-workstation/spd/precision-3660-workstation}} exemplifies edge devices of workstation type, which typically offer exceptional performance and high configurability. It typically offer exceptional performance and high configurability.Equipped with the latest generation Intel Xeon processors\footnote{Intel Xeon \url{https://www.intel.com/content/www/us/en/products/details/processors/xeon.html}} and NVIDIA Quadro GPUs,\footnote{NVIDIA Quadro GPUs \url{https://www.nvidia.com/en-us/design-visualization/quadro/}} it can meet the training and deployment needs of resource intensive DL applications. Additionally, it provides flexible expansion options and comprehensive security features, including the Trusted Platform Module (TPM) 2.0 and hardware security modules,\footnote{TMP \url{https://support.microsoft.com/en-us/topic/what-is-tpm-705f241d-025d-4470-80c5-4feeb24fa1ee}} to protect sensitive data. Furthermore, some workstations support multi-GPU configurations. These workstations are specifically designed to handle highly parallel computing tasks, such as large-scale DL model training. Integrating multiple high-performance GPUs, such as NVIDIA's Tesla\footnote{NVIDIA Tesla GPUs \url{https://www.nvidia.com/en-gb/data-center/tesla-v100/}} or Quadro series, these workstations can significantly accelerate the training and inference of the DL model. Workstations that support multiple GPUs also offer high-bandwidth memory configurations and high-speed data transfer interfaces to ensure efficient data transfer between GPUs, maximising parallel computing efficiency. Furthermore, these workstations are usually equipped with advanced cooling systems and power management features to ensure stability and reliability during extended periods of operation.

This section categorises edge computing devices from low-end to high-performance, emphasising their computational capabilities and hardware specifications to highlight their applicability across various applications. With advancements in technology, not only traditional embedded systems and specialised computing platforms but also an increasing number of smartphones, laptops, automobiles and even smart home devices have been optimised for deep learning, extending their use into the domain of edge computing \citep{murshed2021machine}. This diversity ensures that users can select the most appropriate technology based on their specific computational needs and application requirements, effectively leveraging the power of edge computing in diverse environments.

\section{Transferring Large DNNs to Lightweight DNNs}
\label{sec:strategies}
The need to implement DNNs on devices with limited resources has led to the creation of lightweight neural network architectures \citep{khan2022neural,khan2024esdmr,farooq2024lssf}. To achieve an optimal balance between computational efficiency and model complexity, these architectures are well-suited for edge devices. Compared to their larger counterparts, lightweight neural networks provide a number of benefits, including a smaller memory footprint, accelerated inference speed, and reduced energy consumption \citep{khan2022t, javed2024advancing,naqvi2023glan,khan2022mkis}. We examine the main methods (Fig.~\ref{fig:DNNToLWDNN}) to obtain lightweight neural networks suited for implementation in edge devices. These methods include network pruning, sparse representation, quantised neural networks, efficient lightweight neural networks, and neural architecture search (see Table~\ref{tab:Comparison_large_and_lightweight} for a comparison of these methods, highlighting their respective advantages, disadvantages, and potential applications).
%Through a concise overview of recent advances in these fields, our objective is to provide an understanding of the most recent methods by which substantial DNNs can be converted to their more lightweight counterparts, thereby facilitating the efficient use of edge computing resources. 

\begin{figure}[!t]
    \centering
    \includegraphics[width=\textwidth]{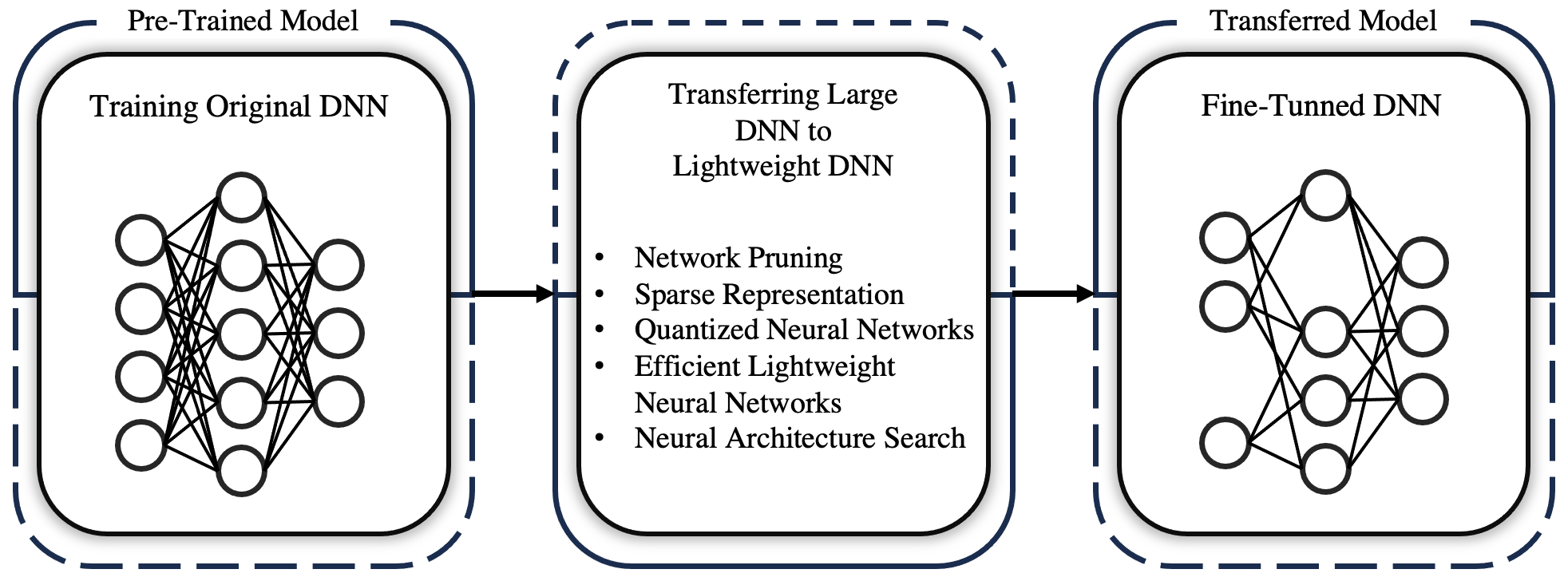}
    \caption{Methods to convert large DNNs to Lightweight DNNs.}
    \label{fig:DNNToLWDNN}
\end{figure}

\subsection{Network Pruning}
Neural networks enable systems to autonomously discern patterns from data \citep{lecun2015deep}. However, achieving high performance in DNN models often entails a trade-off, as it leads to an increased number of neurons and synaptic connections, which in turn significantly increases the computational and spatial complexity of neural network models. Various studies \citep{han2015deep, he2017channel, he2018amc, hu2016network} reveal a substantial presence of redundant neurons and connections within numerous DNN architectures. Identification and removal of these redundant elements have been shown to significantly decrease both computational demands and model size while maintaining accuracy. Addressing this challenge, pruning is a technique designed to automatically pinpoint redundant neurons and connections, facilitating model compression and acceleration. We discuss four different types of pruning often used for model compression, namely weight pruning, neuron pruning, channel pruning, and filter pruning.
\begin{table}[!t]
\centering
\resizebox{\textwidth}{!}{%
\begin{tabular}{l>{\raggedright}p{6cm}>{\raggedright}p{6cm}>{\raggedright\arraybackslash}p{6cm}}
\toprule
\textbf{Method} & \textbf{Advantages} & \textbf{Disadvantages} & \textbf{Applications} \\
\midrule
Network Pruning & Reduces model size and complexity; Maintains accuracy; Suitable for post-training optimisation & Can be computationally intensive; Requires retraining; Pruned networks might need specialised hardware for efficient inference & Image classification, object detection, and scenarios with over-parameterized models. \\
\midrule
Sparse Representation & Reduces memory and computational requirements; Improves generalisation; Makes models more interpretable & Determining optimal sparsity level is challenging; May require specialised algorithms for training and inference & Applications where interpretability and efficiency are crucial, such as medical imaging and real-time analytics. \\
\midrule
Quantised Neural Networks & Significant reduction in memory and computation; Faster inference; Lower energy consumption & Potential loss of accuracy; Requires quantisation-aware training; Bit-width selection is nontrivial & Edge devices, mobile applications, and real-time systems requiring fast and efficient inference. \\
\midrule
Efficient Lightweight Neural Networks & Designed for efficiency from the ground up; High accuracy with low computational cost; No need for post-processing optimisation & May not perform as well as larger networks on very complex tasks; Design is task specific & Mobile and embedded devices, IoT applications, and scenarios with stringent resource constraints. \\
\midrule
Neural Architecture Search (NAS) & Automates the design of optimal architectures; can achieve state-of-the-art performance; considers multiple objectives (accuracy, latency, etc.) & Computationally expensive; may require significant resources and time; Complexity of search space can be high & Diverse applications including image recognition, language processing, and any task needing tailored architectures. \\
\bottomrule
\end{tabular}}
\caption{{Comparison of methods for converting large DNNs to lightweight DNNs.}}
\label{tab:Comparison_large_and_lightweight}
\end{table}

\subsubsection{Weight Pruning}
Weight pruning in CNNs is a technique to reduce the size of the model by eliminating certain weights (parameters) while maintaining performance to some extent. In a neural network, weights represent the strength of connections between neurons. Pruning involves identifying and removing some of these weights, effectively setting them to zero, and then retraining the network to recover performance. Unstructured pruning \citep{han2015deep} aims to reduce the complexity and energy consumption of large DNNs on edge devices, specifically during the inference phase. This technique achieves reductions of 9$\times$ and 13$\times$ in complexity in AlexNet \citep{krizhevsky2012imagenet} and VGG16 \citep{simonyan2014very}, respectively. It has three main stages. First, a network is trained to acquire knowledge about the connection using conventional training techniques \citep{bottou2010large}. Afterwards, a pre-established threshold is used to remove connections with weights below the threshold. Finally, the weights acquired during the initial training process are utilised to establish the starting weights of the pruned neural network. Pruning leads to the weight matrix becoming sparse as numerous weights are eliminated. A compressed sparse row or column structure is utilised to effectively store this sparse matrix.

\begin{figure}[!t]
    \centering
    \includegraphics[width=0.8\textwidth]{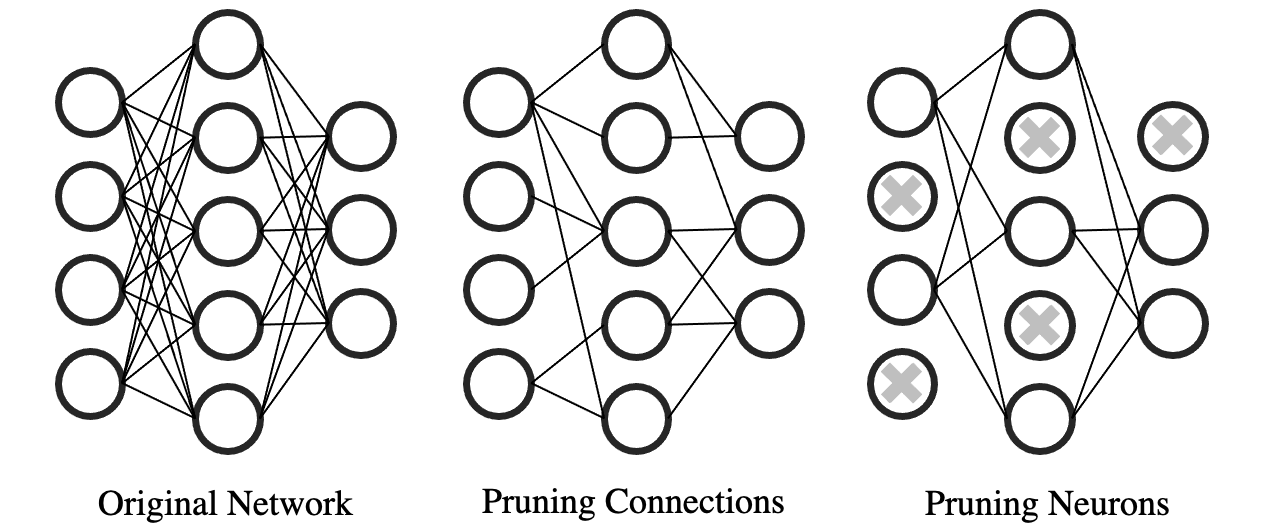}
    \caption{Difference between pruning connections and pruning neurons.}
    \label{fig:NeuronPrune}
\end{figure}

\subsubsection{Neuron Pruning}
Neuron pruning goes beyond weight pruning by not only removing individual weights but entire neurons (or filters) along with their associated weights (Fig.~\ref{fig:NeuronPrune}). Neurons in CNNs correspond to feature maps or filters that capture specific patterns or features in the input data. Similar to weight pruning, neuron pruning is typically performed iteratively. After pruning, the model needs to be fine-tuned to recover any performance loss caused by the removal of neurons. Neuron pruning uses certain criteria to determine which neurons are less important or redundant and can be pruned from the network. To address problems with unstructured pruning \citep{han2015deep} and improve the computational and space efficiency of DNN models, a structured pruning technique can be used \citep{hu2016network}. Contrary to weight-based pruning, this technique specifically targets neurons that exhibit a high frequency of zero activations after the ReLU (rectified linear unit) layer. The technique is based on the recognition of significant duplication of neurons across DNN models. Redundancy in neural networks not only leads to higher computational and space requirements but also worsens overfitting. The average percentage of zeros (APoZ) metric can be used to measure the proportion of zero activations in a neuron following the ReLU layer.

\subsubsection{Channel Pruning}
Weight pruning can decrease the space complexity of a neural network model. However, implementing the resulting sparse structure on hardware platforms might be difficult, and accelerating the inference process usually requires the use of specialised hardware. To overcome this, channel pruning \citep{he2017channel} can be used as an alternate technique for structured pruning specifically for convolutional layers \citep{han2015deep}. Channel pruning is distinct from weight pruning, as it introduces structured sparsity by directly reducing the number of channels in a feature map rather than inducing unstructured sparsity in convolutional layers. Structured sparsity allows for efficient implementation on both CPU and GPU platforms. Inference-based channel pruning \citep{he2017channel} aims to minimise the reconstruction error of a layer's feature map. The optimisation process comprises two primary stages: channel selection and feature map rebuilding. In the first stage, the most representative channels are selected, and redundant channels are eliminated to decrease the complexity of the model. To address pruning mistakes, a linear least-squares approach is used to reconstruct the output feature map using the pruned feature map data. This bipartite procedure guarantees that the pruned model maintains its prognostic effectiveness while reducing computational and memory demands.

Channel pruning enhances the scalability of neural networks by reducing the number of channels in each layer, leading to a more compact model that can be efficiently deployed across various hardware platforms. \citet{he2017channel} demonstrated that channel pruning could reduce the number of parameters in ResNet-50 by 50\% while maintaining 99\% of its original accuracy on ImageNet. This reduction in complexity makes the pruned model highly scalable and suitable for deployment on both CPUs and GPUs.

The structured sparsity introduced by channel pruning makes it flexible for implementation on different types of hardware without requiring specialized inference engines. \citet{li2020eagleeye} applied channel pruning to MobileNetV1, achieving a $2\times$ reduction in the number of FLOPs while maintaining a top-1 accuracy of 70.9\% on ImageNet. This flexibility allows the pruned model to be used in a wide range of applications, from mobile devices to high-performance servers.

Recent advancements in channel pruning have focused on optimizing models for edge devices by further reducing computational complexity and memory usage. \citet{he2020learning} proposed a learning-based channel pruning method that achieved a 60.8\% reduction in the number of parameters for ResNet-50, while maintaining 98\% of its original accuracy on ImageNet. This makes it highly suitable for real-time applications on edge devices such as smartphones and IoT devices.

Channel pruning has been combined with other compression techniques, such as quantisation and knowledge distillation, to create highly efficient models for deployment in resource-constrained environments. \citet{bai2023unified} integrated channel pruning with quantisation, achieving a top-1 accuracy of 69.86\% on ImageNet for a pruned and quantised version of ResNet-34, with a 5$\times$ reduction in the model size. This hybrid approach ensures that the model remains efficient and accurate and suitable for mobile and embedded applications.

Furthermore, channel pruning has been applied to object detection models, where maintaining high accuracy is crucial while reducing the size of the model and the computational cost. \citet{xu2022cap} applied channel pruning to YOLOv3 maintaining 39.8\% of the original mean average precision (mAP) in the COCO data set. This makes the pruned model highly efficient for real-time object detection tasks on edge devices.

\begin{figure}[!t]
    \centering
    \includegraphics[width=0.7\textwidth]{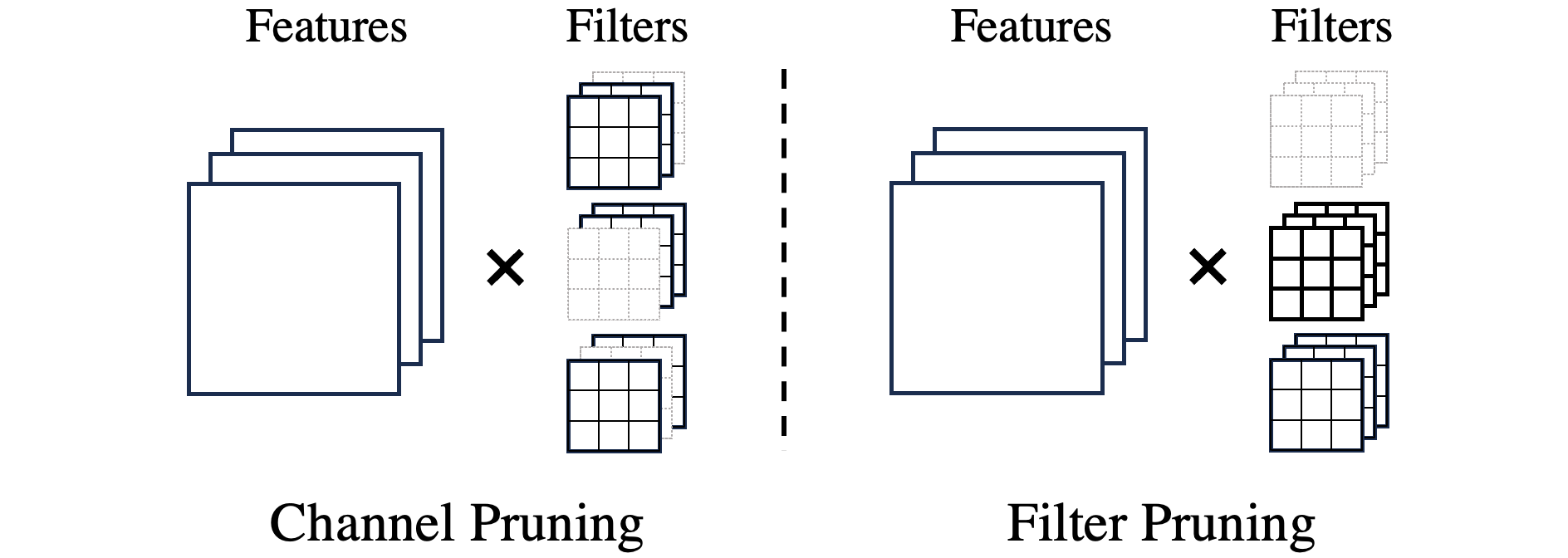}
    \caption{Channel pruning versus filter pruning.}
    \label{fig:FilterVsChannel}
\end{figure}

\subsubsection{Filter Pruning}
Filter pruning involves identifying and eliminating redundant or less impactful filters within the network without compromising overall performance (Fig.~\ref{fig:FilterVsChannel}). Filter pruning is motivated by the recognition that neural networks often contain filters that contribute minimally to the network's representational capacity. Removing these less crucial filters can result in a more streamlined model with improved efficiency, making it particularly valuable for deployment in resource-constrained environments such as edge devices or mobile applications. A common strategy involves evaluating the importance of filters based on metrics such as weight magnitudes, activation patterns, or gradient information during the training process. Filters identified as less critical are subsequently pruned from the network. Another approach incorporates regularisation techniques tailored to encourage sparsity in the network. L1 regularisation, for example, penalises nonessential parameters, making them more likely to be pruned during the training phase. Deep compression with filter pruning has been shown to result in significant model compression without sacrificing accuracy, particularly in tasks such as image classification \citep{han2015deep}. Others have formulated filter pruning as an optimisation problem and efficiently solved it through a three-step algorithm \citep{he2017channel}. Recent advances in filter pruning include the AMC (AutoML for Model Compression) framework, which leverages reinforcement learning to dynamically determine which filters to prune \citep{he2018amc}. The choice of a specific pruning method often depends on the specific application, resource constraints, and the desired trade-off between model size and performance.

In recent times, there have been numerous groundbreaking developments in filter pruning methods that aim to improve the effectiveness and efficacy of deep neural networks. Soft filter pruning is one such technique; it permits dynamic modifications to be made during training, allowing previously pruned filters to be reinstated in subsequent training epochs. The method illustrated by \citet{he2019soft} provides greater versatility and adaptability when it comes to managing a wide range of dynamic and changing datasets. Moreover, \citet{yang2017designing} have devised an energy-aware pruning approach that prioritises the optimisation of neural network energy consumption and computational efficiency. This characteristic makes the strategy highly suitable for implementation in environments with energy restrictions, such as mobile devices.

Genetic algorithms have been used to optimise filter selection in a more exploratory way, further expanding the scope of filter pruning \citep{kim2018ga}. This methodology emulates a natural selection process in which only the most efficient filters are preserved, revealing distinctive network configurations that traditional techniques may fail to identify \citep{kim2018ga}. In addition, \citet{lu2022bayesian} have proposed a Bayesian sparsity pruning technique that employs probabilistic models to evaluate the redundancy of filters. This approach is particularly valuable in situations where there are ambiguous or noisy data. It offers a statistically reliable framework for making pruning decisions \citep{lu2022bayesian}. 

Filter pruning improves the scalability of neural networks by reducing the number of filters, leading to a more compact model that can be efficiently deployed across various hardware platforms. \citet{he2017channel} demonstrated that filter pruning could accelerate ResNet-50 by 2$\times$ while maintaining 98.6\% of its original accuracy on ImageNet. This reduction in complexity makes the pruned model highly scalable and suitable for deployment on both CPUs and GPUs. The structured sparsity introduced by filter pruning makes it flexible for implementation on different types of hardware without requiring specialised inference engines. \citet{li2020eagleeye} applied filter pruning to MobileNetV2, achieving a $2\times$ reduction in the number of FLOPs while maintaining a top-1 accuracy of 70.6\% on ImageNet. This flexibility allows the pruned model to be used in a wide range of applications, from mobile devices to high-performance servers.

Recent advances in filter pruning have focused on optimising models for edge devices by further reducing computational complexity and memory usage. \citet{he2020learning} proposed a learning-based filter pruning method that achieved a 60.3\% reduction in FLOPs for ResNet-56, while maintaining 99\% of its original accuracy on CIFAR-10. This makes it highly suitable for real-time applications on edge devices, such as smartphones and IoT devices. Filter pruning has been combined with other compression techniques, such as quantisation and knowledge distillation, to create highly efficient models for deployment in resource-constrained environments. \citet{bai2023unified} integrated filter pruning with quantisation, achieving a top-1 accuracy of 74.69\% on ImageNet for a pruned and quantised version of ResNet-101, with a 5.3$\times$ reduction in the model size. This hybrid approach ensures that the model remains efficient and accurate and suitable for mobile and embedded applications.

\begin{figure}[!t]
    \centering
    \includegraphics[width=0.9\textwidth]{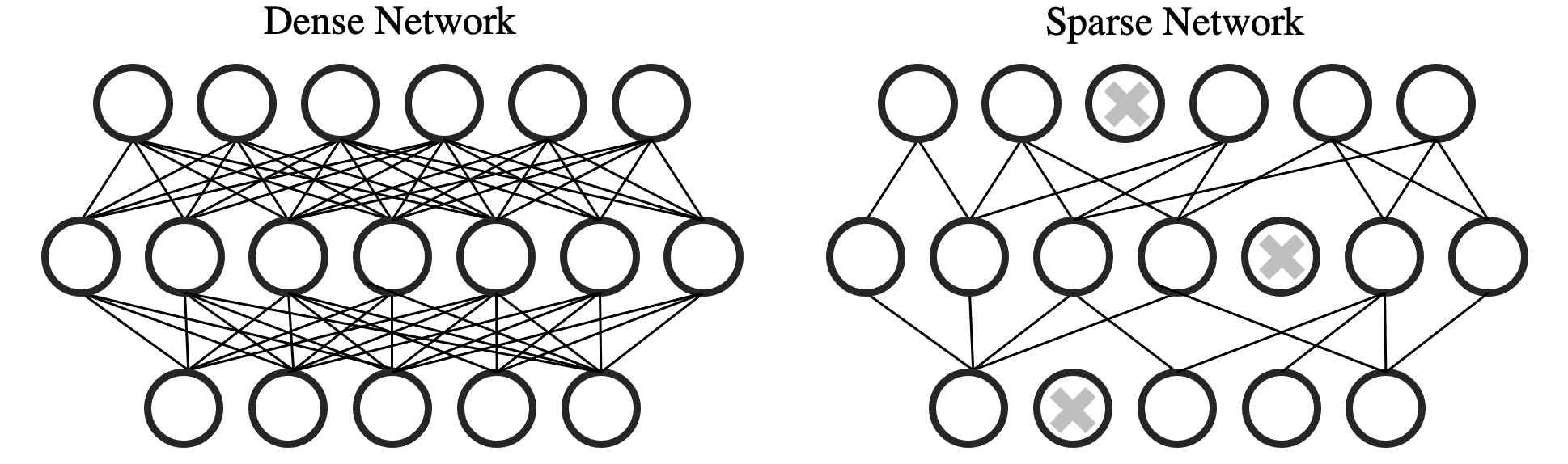}
    \caption{Dense versus sparse representation.}
    \label{fig:Sparse}
\end{figure}
\subsection{Sparse Representation}
Sparse representation, characterised by the presence of a relatively small number of nonzero elements (Fig.~\ref{fig:Sparse}), has emerged as a crucial aspect of neural network optimisation. The motivation behind promoting sparsity in neural networks lies in the observation that not all connections and activations are equally essential for effective learning and representation. Sparse representations facilitate a more compact and interpretable model, reducing redundancy, and improving generalisation.

Various techniques have been proposed to induce sparsity in neural networks during training and inference. A common technique is to apply L1 regularisation in the training process to penalise nonessential parameters and encourage the model to set many of them to zero. Another technique is the use of activation functions, such as ReLU, that naturally lead to sparse activations and can contribute to a sparser network.

The introduction of sparsity in neural networks offers several advantages. First, sparse networks have lower computational and memory requirements, making them more efficient and suitable for deployment in resource-constrained environments. Second, sparse models are more interpretable, as nonzero weights and activations can be directly associated with influential features and connections. And third, sparsity acts as a form of regularisation, preventing overfitting and resulting in a more robust model.

Despite the advantages, sparse representation in neural networks poses challenges, such as determining an optimal level of sparsity and addressing potential performance trade-offs. The ongoing research aims to develop more efficient pruning algorithms, adaptive sparsity-inducing techniques, and strategies to mitigate sparse model drawbacks.

Another progress is the emergence of structured pruning approaches that not only eliminate individual weights but also prune entire filters or layers in a network. This strategy preserves structural integrity while substantially decreasing the complexity. \citet{qi2021efficient} applied structured pruning to ResNet-50, achieving a $3\times$ reduction in the number of parameters and a $2\times$ reduction in FLOPs while maintaining 97\% of the original accuracy on ImageNet. This makes the model highly efficient for real-time image classification tasks on edge devices.

Furthermore, scholars have investigated the possibility of incorporating sparsity into additional network optimisation techniques. \citet{hu2023bag} implemented a hybrid approach of quantisation and sparsity on MobileNetV2, achieving a top-1 accuracy of 73.1\% on ImageNet with reduction in model size and inference latency. This makes it suitable for deployment in mobile devices for applications such as real-time image recognition and augmented reality.

Furthermore, an innovative approach to dynamically modify sparsity levels during training was proposed by \citet{sanh2020movement} in the form of an adaptive sparsity method. It used movement pruning on BERT, achieving substantial reductions in model size while maintaining much of the original performance on tasks such as SQuAD, MNLI, and QQP. This makes adaptive sparsity a promising direction for reducing the memory footprint of BERT-like models on resource-constrained devices.

Furthermore, certain works concentrate on the theoretical aspects of sparsity, in addition to these algorithmic developments. \citet{galanti2023norm} conducted theoretical research on the limitations of generalisation errors in sparse neural networks. Their study sheds light on how sparsity enhances the resilience and generalisation capabilities of models.

\subsection{Quantised Neural Network}

Quantised neural networks (QNNs) are networks in which the precision of the model parameters is reduced to a discrete set of quantised values (Fig.~\ref{fig:Quantised}). Quantisation involves mapping high-precision weights to lower bit-width representations, such as 8-bit integers. QNNs offer a trade-off between model accuracy and computational efficiency, making them particularly attractive for resource-constrained environments.

Several quantisation techniques have been developed to reduce the bit-width of weights and activations in neural networks. The most common technique is to reduce the precision of the weights. For example, weights originally represented as 32-bit floating-point numbers can be quantised to 8-bit integers. In addition to quantising the weights, the activations can be quantised as well, further contributing to the overall compression of the model. The most extreme form of quantisation is to binarise the weights and activations (1-bit). Although challenging, binary quantisation significantly reduces memory requirements and computational complexity. Finally, mixed precision quantisation allows different layers of the neural network to be quantised to different bit-widths based on their sensitivity to precision, optimising the trade-off between accuracy and efficiency.

The use of QNNs has several advantages. Quantisation substantially decreases the memory requirements of neural network models, allowing deployment on memory-constrained devices. Furthermore, with lower bit-width representations, quantised models lead to faster inference times, making them suitable for real-time applications. In addition, quantised models require less computational resources, resulting in reduced energy consumption, which is particularly crucial for edge devices and mobile applications.

On the other hand, QNNs come with challenges and considerations. Reduction in precision can lead to loss of model accuracy, especially for tasks that require fine-grained distinctions. To mitigate accuracy degradation, it is crucial to train models with quantisation in mind and employ quantisation-aware training techniques \citep{9181327, shen2021quantizationaware}. Choosing the appropriate bit-width for quantisation is a nontrivial task and often depends on the specific characteristics of the neural network and the target platform.

Recent research has focused on developing advanced quantisation methods and training strategies to overcome the challenges associated with QNNs. Techniques like mixed-precision quantisation \citep{wang2019haq}, learning-aware quantisation \citep{Esser2020learned}, and hardware-aware quantisation \citep{huang2024ohq} aim to push the boundaries of model efficiency while maintaining acceptable levels of accuracy.

An approach worth mentioning is adaptive quantisation, which modifies the bit-widths dynamically throughout the training process in response to the weight and activation distribution. The aforementioned approach enables greater precision in regulating the quantisation procedure, potentially resulting in improved efficacy when faced with intricate tasks \citep{dong2019hawq}.  The utilisation of quantisation intervals that are learnt during training as opposed to fixed is an additional significant development. The method outlined by \citet{lee2018retraining} allows the quantisation procedure to adjust to the unique attributes of the data, potentially improving the reliability and precision of the quantised model.

Moreover, researchers have also investigated the use of quantisation with other methods for compressing networks, such as pruning and knowledge distillation. \citet{kim2021pqk} proposed an integrated framework that integrates various strategies to significantly reduce the size of the model and computing needs, while maintaining a reasonable level of accuracy.

Research on application-specific quantisation techniques has recently gained attention. In the realm of natural language processing, specific quantisation strategies have been created to address the distinct difficulties presented by huge language models. \citet{frantar2022gptq} presented a technique tailored for transformer designs that effectively manages the trade-off between efficiency and performance in these computationally demanding models.

Mixed precision quantisation has been widely adopted for deploying models on mobile and edge devices due to its balance between accuracy and efficiency. \citet{wang2019haq} demonstrated that a mixed-precision quantised version of  ResNet-50 achieved a top--1 accuracy of 75. 3\% on ImageNet while reducing model size by $10\times$ and inference latency compared to the full-precision model. This makes it suitable for real-time image recognition tasks on mobile devices. \citet{zafrir2019q8bert} applied quantization techniques to BERT, a popular transformer model, reducing its memory footprint by $4\times$ and maintaining 98\% of its original accuracy on the GLUE benchmark. This makes it feasible to deploy BERT on mobile devices for real-time text processing tasks. \citet{huang2024ohq} implemented a hardware-aware quantised version of MobileNetV2, achieving a top-1 accuracy of 72.2\% on ImageNet with only 220 million FLOPs and significant energy savings. This approach is particularly beneficial for edge devices with limited power resources.

\begin{figure}[!t]
    \centering
    \includegraphics[scale=0.33]{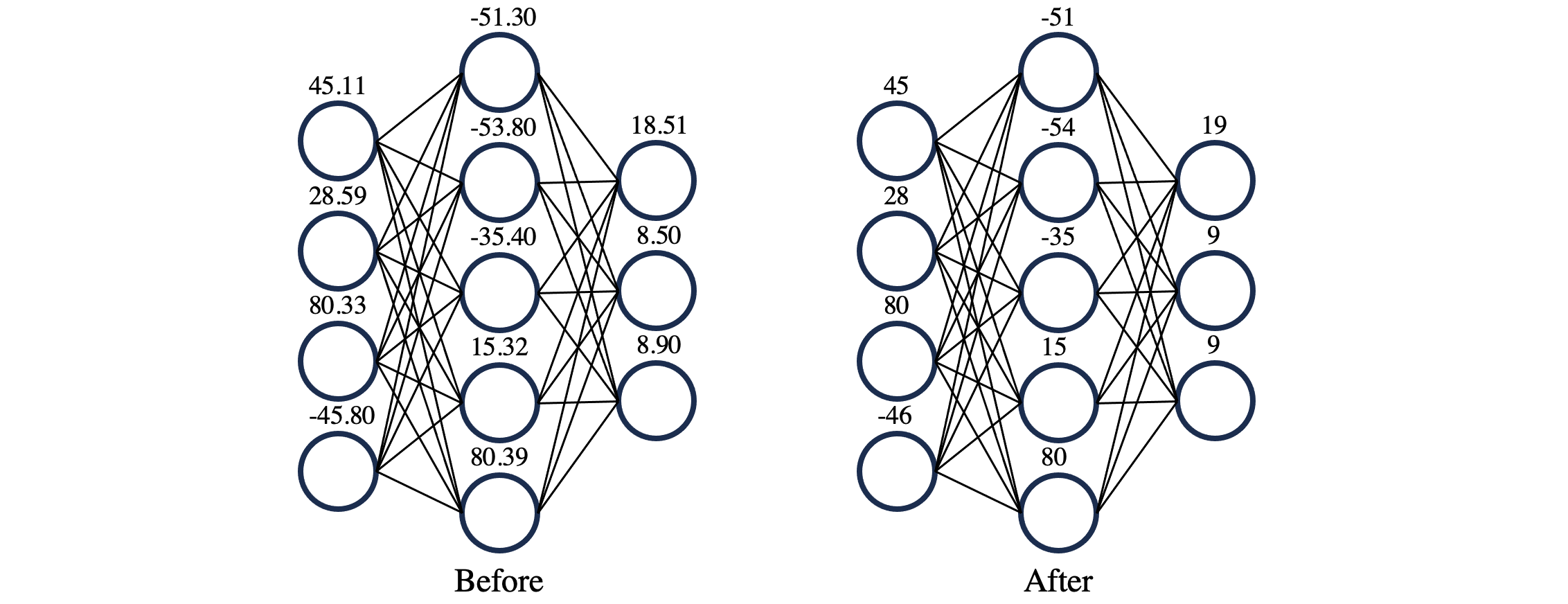}
    \caption{Quantisation of a neural network. Here, the original model parameter values before quantisation are floating-point numbers, which are quantised to integers.}
    \label{fig:Quantised}
\end{figure}

\subsection{Efficient Lightweight Neural Networks}
Many neural network architectures exist nowadays that are lightweight by design rather than by reducing a given large network using pruning, sparsification, or quantisation. Here, we survey the most prominent and efficient lightweight neural networks.

\subsubsection{MobileNet}
MobileNet \citep{howard2017mobilenets} is a family of neural network architectures designed for mobile and embedded devices. The first iteration, MobileNetV1, was introduced in 2017, pioneering the use of depthwise separable convolutions for efficient model design. It struck a balance between the size of the model and its accuracy, making it suitable for on-device tasks. MobileNetV2 followed in 2018, improving performance through inverted residual blocks and linear bottlenecks. It offered better accuracy and efficiency, particularly for real-time applications on mobile hardware. MobileNetV3 continued this trend in 2019 with optimisations such as the h-swish activation function, further improving both accuracy and efficiency. These models are well-regarded for their practicality in various mobile vision tasks.

\citet{howard2017mobilenets} demonstrated that MobileNetV1 achieved a top-1 accuracy of 70.6\% on ImageNet with 569 million FLOPs, significantly reducing computational cost compared to traditional convolutional neural networks. MobileNetV2 further enhanced scalability, achieving a top-1 accuracy of 72.0\% on ImageNet with only 300 million multiply-add operations (MAdds) \citep{sandler2018mobilenetv2}. \citet{howard2019searching} demonstrated that MobileNetV3-Large achieved a top-1 accuracy of 75.2\% on ImageNet with  only 219 million MAdds, making it highly suitable for mobile and edge applications such as real-time image recognition and augmented reality. \citet{hashmi2020efficient} used MobileNetV2 to detect pneumonia in chest X-rays. MobileNetV2 achieved a precision of 96.08\% and accuracy of 96.71\% while significantly reducing inference time compared to previous models, making it practical for real-time diagnostic tools on edge devices.

MobileNet variants have exhibited considerable success in various practical domains, highlighting their performance in edge-specific environments. In agriculture, \citet{lu2023improved} employed MobileNetV2 for the real-time detection of plant diseases utilising images obtained from smartphones. This application demonstrates MobileNetV2's efficacy in low-latency, on-device processing, allowing farmers to make prompt, informed decisions in the field without reliance on high-performance computing resources. The model's capacity to operate on devices with constrained computational resources while preserving high accuracy allows implementation in rural or resource-deficient environments. Likewise, \citet{yang2023real} utilised MobileNetV3 on drones for instantaneous object detection. Their study illustrated that the lightweight architecture of MobileNetV3 enabled drones to execute precise detection tasks without overburdening onboard processors, thereby ensuring extended flight durations and enhanced power efficiency.

In healthcare, in addition to pneumonia detection, MobileNet has been utilised in numerous diagnostic instruments. For example, \citet{ragab2022covid} utilised MobileNetV2 to identify anomalies in chest X-rays within the framework of a comprehensive mobile health initiative. This application highlights MobileNet's efficacy in providing real-time diagnostic insights in regions with restricted access to conventional medical imaging tools. Furthermore, \citet{caesarendra2022embedded} employed MobileNetV3 for real-time ECG monitoring on wearable devices, demonstrating its capacity to manage continuous time-series data with minimal latency, rendering it appropriate for the early identification of cardiac events in routine environments.

MobileNet variants have been extensively utilised in robotics within autonomous systems. For instance, \citet{aulia2024new} utilised MobileNetV2 for real-time object recognition on mobile robots, illustrating its applicability for edge devices with constrained computational resources and battery longevity. The model's low memory usage and fast inference speed enabled the robots to execute real-time decisions independently of cloud processing, thereby enhancing the system's reliability and responsiveness in dynamic settings. The literature case studies demonstrate the adaptability of MobileNet variants to diverse real-world edge applications, confirming their versatility and efficacy in both industrial and consumer sectors.

\subsubsection{EfficientNet}
EfficientNet \citep{tan2019efficientnet} takes a different approach to optimisation. Using compound scaling, EfficientNetV1 uniformly scales network width, depth, and image resolution, thus creating a spectrum of models ranging from EfficientNetB0 to EfficientNetB7. EfficientNetV1 is typically synonymous with the foundational model architecture outlined in the paper. This methodology quickly garnered attention for its ability to strike a balance between efficiency and accuracy, particularly excelling in image classification tasks.

The variations within the EfficientNet family, from B0 to B7, diverge in their architectures, depth, and computational complexity. Beginning with B0, the models progressively increase in size and complexity. EfficientNetB0 is the smallest and simplest model, having fewer layers and parameters. On the contrary, EfficientNetB7 emerges as the largest, featuring a deeper and wider architecture capable of capturing intricate data patterns.

Typically, transitioning from B0 to B7 results in improved performance in tasks such as image classification or object detection. However, this improvement comes at the expense of increased model capacity, manifested in larger parameter counts and computational requirements. The decision on which variant to employ depends on the task-specific demands, available computational resources, and the desired balance between the size, speed, and accuracy of the model.

%\citet{tan2020efficientnetlite} demonstrated that EfficientNet-Lite0 achieved a top-1 accuracy of 75.1\% on ImageNet with only 5.4 million parameters and 390 million FLOPs, making it highly suitable for mobile and edge applications such as real-time image recognition and augmented reality. \citet{tan2021efficientnetv2} reported that EfficientNetV2-S achieved a top-1 accuracy of 83.9\% on ImageNet while being $2\times$ faster to train compared to its predecessor EfficientNetV1. This makes EfficientNetV2 highly applicable for edge devices where fast deployment and inference are crucial. \citet{wang2020efficientnet} used EfficientNet for detecting diabetic retinopathy in retinal images. EfficientNet-B3 achieved a sensitivity of 92.1\% and specificity of 90.3\% while significantly reducing inference time compared to previous models, making it practical for real-time diagnostic tools on edge devices. \citet{li2021efficientnet} applied EfficientNet-B4 for object detection and scene understanding in autonomous vehicles. The model provided state-of-the-art accuracy while maintaining low latency and energy consumption, crucial for real-time decision-making in autonomous driving.

 The compound scaling strategy of EfficientNet has demonstrated significant efficacy across numerous practical applications, where the trade-off between computational efficiency and accuracy is paramount. For instance, \citet{elmoufidi2022diabetic} illustrated the application of EfficientNetB3 in healthcare, particularly for the detection of diabetic retinopathy in retinal images. Their study demonstrated that the lightweight architecture of EfficientNet attained high sensitivity (98.1\%) and specificity (98.3\%) while reducing inference time, making it suitable for deployment on edge devices like portable diagnostic tools, where rapid and precise decision-making is crucial. In another study, \citet{tan2021efficientnetv2} utilised EfficientNetV2 in autonomous vehicles for object detection and scene comprehension, where real-time performance and minimal latency are essential for safety and functionality. EfficientNetV2 delivered exceptional accuracy while optimising energy consumption, enabling efficient operation on the constrained processing power of embedded systems in vehicles.

To further illustrate EfficientNet's relevance to edge scenarios, \citet{tan2019efficientnet} presented EfficientNet, which was explicitly designed for mobile and edge applications by reducing model complexity while maintaining performance. EfficientNet-B0 achieved a top-1 accuracy of 77.1\% on ImageNet with merely  5.3 million parameters and 390 million FLOPs, making it exceptionally suitable for applications like real-time image recognition on smartphones or augmented reality systems, where power efficiency and rapid inference are paramount.} Furthermore, \citet{tan2021efficientnetv2} showed that EfficientNetV2-S enhanced accuracy, achieving 83.9\% top-1 accuracy on ImageNet, while significantly decreasing training duration, rendering it beneficial for contexts necessitating swift deployment, such as in retail, healthcare, and autonomous systems, where fast training and inference are paramount.

\subsubsection{Group Convolutional Networks}
Group-convolutional networks (Table~\ref{GroupConv}), with their various iterations and innovative approaches, offer efficient alternatives to traditional deep neural networks. They are well suited for a wide range of applications, from mobile and edge devices to real-time tasks, where computational efficiency is a critical factor. The choice of architecture depends on the specific resource constraints and performance requirements of a given project.

ShuffleNet \citep{zhang2018shufflenet} is a family of neural network architectures that revolutionised the efficiency of models. ShuffleNetV1, introduced in 2017, leveraged group convolutions and channel shuffling to significantly reduce computational costs while maintaining competitive accuracy. It became the go-to choice for mobile and embedded devices where computational resources are limited. ShuffleNetV2, released in 2018, further improved efficiency by introducing residual connections, offering an even better balance between accuracy and computational efficiency. These models are known for their ability to provide impressive performance on resource-constrained platforms. ShuffleNet effectively balances accuracy and computational cost by using efficient architectural strategies. This approach ensures that the model can achieve high accuracy while minimising resource usage. In their experiments, \citet{zhang2018shufflenet} found that ShuffleNet consistently outperformed other lightweight models in terms of accuracy-efficiency trade-offs. For example, ShuffleNetV2 achieved a top-1 accuracy of 69.4\% on ImageNet with only 146 million FLOPs, significantly reducing computational cost while maintaining competitive performance compared to models such as MobileNetV2 and CondenseNet.

CondenseNet \citep{huang2017condensenet}, introduced in 2018, follows a two-step pipeline to prune and condense neural networks. Removing less important connections significantly reduces the size of the model while maintaining competitive accuracy. It is a highly efficient architecture, well-suited for applications with strict resource constraints. CondenseNet balances accuracy and computational cost effectively by pruning less important connections and condensing the network structure. This approach ensures that the model can achieve high accuracy while minimising resource usage. In their experiments, \citet{huang2017condensenet} found that CondenseNet consistently outperformed other pruned models in terms of accuracy-efficiency trade-offs. For example, CondenseNet achieved atop-1 accuracy of 73.8\% on ImageNet with only 4.8 million parameters, significantly reducing computational cost while maintaining competitive performance compared to models like MobileNet and ShuffleNet.

MixNet \citep{tan2019mixnet}, unveiled in 2019, uses mixed depthwise convolutions to reduce computation while preserving accuracy. It offers a flexible architecture that can be easily customised to meet different resource constraints. MixNet has earned recognition for achieving state-of-the-art efficiency-accuracy trade-offs, making it a versatile choice for various applications. MixNet balances accuracy and computational cost effectively employing mixed depth-wise convolutions. This approach ensures that the model can achieve high accuracy while minimising resource usage. In their experiments, \citet{tan2019mixnet} found that MixNet consistently outperformed other lightweight models in terms of accuracy-efficiency trade-offs. For example, MixNet-M achieved a top-1 accuracy of 77.0\% on ImageNet with only 360 million FLOPs, significantly reducing computational cost while maintaining competitive performance compared to models such as MobileNetV3 and EfficientNet-Lite.

GhostNet \citep{han2020ghostnet}, developed in 2020, introduces ``ghost'' modules to produce lightweight feature maps, significantly reducing the number of parameters and computations. This innovative approach allows GhostNet to maintain competitive performance while being highly efficient, particularly suited for tasks where computational resources are limited. GhostNet balances accuracy and computational cost effectively by generating more feature maps from fewer computations. This approach ensures that the model can achieve high accuracy while minimising resource usage. In their experiments, \citet{han2020ghostnet} found that GhostNet consistently outperformed other lightweight models in terms of accuracy-efficiency trade-offs. For example, GhostNet achieved a top-1 accuracy of 73.9\% on ImageNet with only 141 million FLOPs, significantly reducing computational cost while maintaining competitive performance compared to models like MobileNetV2 and ShuffleNet.

\begin{table}[!t]
\centering
\resizebox{\textwidth}{!}{%
\begin{tabular}{l>{\raggedright}p{6cm}>{\raggedright}p{6cm}>{\raggedright\arraybackslash}p{6cm}}
\toprule
\textbf{Method} & \textbf{Advantage} & \textbf{Disadvantage} & \textbf{Characteristic} \\
\midrule
ShuffleNet & Significantly reduces computational costs. & May have slightly lower accuracy compared to more resource-intensive models. & Leverages group convolutions and channel shuffling for efficiency. Improved in ShuffleNet\-V2 with residual connections. \\
\midrule
CondenseNet & Greatly reduces model size while maintaining competitive accuracy. & Pruning and condensing process may require careful tuning. & Follows a two-step pipeline for pruning and condensing networks. Highly efficient. \\
\midrule
MixNet & Uses mixed depthwise convolutions for efficiency-accuracy trade-offs. & Customisation may require expertise in architecture design. & Offers flexibility and customisation to meet various resource constraints. Achieves state-of-the-art trade-offs. \\
\midrule
GhostNet & Introduces ``ghost'' modules for lightweight feature maps and reduced computation. & Adaptation of the ``ghost'' modules may be task-specific. & Maintains competitive performance with a significant reduction in parameters and computation. The application is well suited for limited computational resources. \\
\midrule
DiCENet & Dynamic inference channels adjust the number of active channels at runtime for efficient computation. & May require additional complexity for dynamic channel management. & Designed for real-time and mobile applications, prioritising adaptability and efficiency. \\
\midrule
MicroNet & Optimised for ultralow resource constraints. & May have limitations in handling complex tasks due to resource constraints. & Includes MicroNetV1 and MicroNetV2 for edge and IoT applications. Provides options for highly resource-constrained environments. \\
\bottomrule
\end{tabular}}
\caption{Summary of group convolutional networks.}
\label{GroupConv}
\end{table}

DiCENet \citep{mehta2020dicenet}, incorporates dynamic inference channels to adjust the number of active channels at runtime. This adaptability significantly reduces the computational and memory footprint, allowing for varying resource constraints. DiCENet is designed for real-time and mobile applications, prioritising efficiency without compromising performance. DiCENet balances accuracy and computational cost effectively by dynamically adjusting its inference channels. This approach ensures that the model can maintain high accuracy while minimising resource usage. In their experiments, \citet{mehta2020dicenet} found that DiCENet consistently outperformed other dynamic models in terms of accuracy and efficiency trade-offs. For example, DiCENet achieved a top-1 accuracy of  75.7\% on ImageNet with significantly lower computational cost compared to static models, making it an attractive choice for resource-constrained applications.

MicroNet \citep{banbury2021micronets} is a collection of neural network architectures optimised for ultra-low resource constraints. These models provide a range of options for resource-constrained environments, making them highly suitable for edge and IoT applications. MicroNet balances accuracy and computational cost by employing efficient architectural strategies tailored for low-resource environments. In their experiments, \citet{banbury2021micronets} found that MicroNet models consistently outperformed other ultra-low-resource architectures in terms of accuracy while maintaining minimal computational requirements. Their results show that MicroNet can match or surpass other competing  lightweight models like MCUNet \citep{lin2020mcunet} and EfficientNet-Lite \citep{tan2019efficientnet}.

\subsubsection{Squeeze \& Excitation}
Squeeze \& excitation (SE) networks, including SqueezeNet, SENet, and SqueezeNeXt (Table~\ref{SEFamily}), have made significant contributions to the field of deep learning. SqueezeNet focusses on model compression and efficiency, while SENet uses crucial attention mechanisms that enhance model performance, and SqueezeNeXt combines the best of both worlds, offering a balance between computational efficiency and accuracy, which makes it a valuable choice for various applications with limited computational resources.

SqueezeNet \citep{iandola2016squeezenet}, introduced in 2016, is a pioneering SE network that stands out for its exceptional compression and efficiency. It achieves these characteristics by using small 1$\times$1 convolutions, known as ``fire'' modules, to reduce the number of parameters while preserving the representational power of the network. This is a particularly popular choice for deployment on resource constrained devices, such as smartphones and IoT devices. Although it focuses on model compression, it does not incorporate the attention mechanisms of later SE models. SqueezeNet balances accuracy and computational cost, making it a practical choice for resource-constrained environments. In their experiments, \citet{iandola2016squeezenet} found that SqueezeNet maintained competitive accuracy levels while drastically reducing the number of parameters. For example, SqueezeNet achieved a $50\times$ reduction in parameters and a $2\times$ reduction in inference time compared to AlexNet, making it suitable for real-time applications.

SENet \citep{hu2018senet}, introduced in 2017, addresses the limitations of conventional neural networks by introducing attention mechanisms. It employs a ``squeeze'' step to capture global statistics from feature maps, and an ``excitation'' step to adaptively recalibrate the importance of different channels. This self-attention mechanism significantly increases model performance by allowing the network to focus on relevant information and ignore less informative features. SENet has become a critical advancement in computer vision, achieving state-of-the-art accuracy in various image classification tasks. The squeeze-and-excitation mechanism is flexible and can be applied to various types of neural network, including CNN and RNN. This flexibility allows SENet to be used in a wide range of applications. \citet{hu2018senet} showed that SENet could be effectively applied to different architectures such as ResNet, Inception, and MobileNet. For example, SENet improved the top-1 accuracy of MobileNetV2 from 70.6\% to 74.7\% on the ImageNet dataset.

SqueezeNeXt \citep{gholami2018squeezenext} builds on the concepts of SqueezeNet and SENet. Introduced in 2017, it combines the efficiency of SqueezeNet with the SENet attention mechanism. The network employs parallel paths with varying channel sizes to capture a wide range of information, promoting both efficiency and accuracy. By integrating channel-wise attention mechanisms inspired by SENet, SqueezeNeXt effectively boosts model performance. This architecture has become a popular choice for applications where a balance between computational efficiency and accuracy is essential, particularly in scenarios with limited computational resources. SqueezeNeXt balances accuracy and computational cost by leveraging the efficiency of SqueezeNet and the performance boost from SENet's attention mechanisms. In their experiments, \citet{gholami2018squeezenext} found that SqueezeNeXt consistently outperformed other lightweight models in terms of accuracy and efficiency trade-offs. For instance, SqueezeNeXt-23 achieved a top-1 accuracy of 67.18\% on ImageNet with only 2.4 million parameters and 1.5 billion FLOPs, making it an attractive choice for resource-constrained environments.

\begin{table}[!t]
\centering
\resizebox{\textwidth}{!}{%
\begin{tabular}{l>{\raggedright}p{6cm}>{\raggedright}p{6cm}>{\raggedright\arraybackslash}p{6cm}}
\toprule
\textbf{Method} & \textbf{Advantage} & \textbf{Disadvantage} & \textbf{Characteristic} \\
\midrule
SqueezeNet & Exceptional model compression and efficiency. & Lacks attention mechanisms for enhanced feature selection. & Utilises ``fire'' modules for model compression. Ideal for resource constrained devices. \\
\midrule
SENet & Uses attention mechanisms for improved feature selection. & May be computationally more expensive due to attention mechanisms. & Employs ``squeeze'' and ``excitation'' steps for adaptive recalibration of channel importance. Achieves state-of-the-art accuracy. \\
\midrule
SqueezeNeXt & Combines efficiency of SqueezeNet with the SENet attention mechanisms. & Slightly more complex compared to SqueezeNet. & Integrates parallel pathways with varying channel sizes and channel-wise attention mechanisms for a balance between efficiency and accuracy. \\
\bottomrule
\end{tabular}}
\caption{Summary of squeeze \& excitation networks.}
\label{SEFamily}
\end{table}

\subsubsection{Mobile Transformers}
Mobile transformers are transformer-based neural network architectures designed to be efficient and lightweight, making them suitable for deployment on mobile devices with limited computational resources. Transformers, originally introduced for natural language processing tasks, have proven to be powerful for a wide range of applications, including computer vision.

MobileViT \citep{mehta2022mobilevit} combines the strengths of lightweight CNNs and heavy weight self-attention-based vision transformers (ViT) to create a lightweight and low-latency network for mobile vision tasks. It introduces a unique vision transformer designed for mobile devices, treating transformers as convolutions for a different perspective on global information processing. Experimental results show that MobileViT outperforms CNN-based networks (specifically MobileNetV3) and ViT-based networks (specifically DeIT) in various tasks and datasets. In the ImageNet-1k dataset, MobileViT achieves a top-1 accuracy of 78.4\% with approximately 6 million parameters, surpassing MobileNetV3 by 3.2\% and DeIT by 6.2\% with a similar parameter count. In object detection on MS-COCO, MobileViT is 5.7\% more accurate than MobileNetV3 with a comparable number of parameters.

EdgeViT-S \citep{pan2022edgevits} address the computational and model size challenges associated with self-attention-based ViTs for mobile devices. They introduce a new family of lightweight ViTs with a focus on low on-device latency and high energy efficiency. EdgeViT-S incorporate a local-global-local (LGL) information exchange bottleneck that efficiently integrates self-attention and convolutions. This design allows them to compete with the best lightweight CNNs in terms of the trade-off between accuracy and on-device efficiency. EdgeViT-S are positioned as Pareto-optimal models, excelling in both accuracy-latency and accuracy-energy trade-offs. Models demonstrate strict dominance over other ViTs in almost all cases and compete with the most efficient CNNs when evaluated based on on-device latency and energy efficiency. EdgeViT-S are designed to efficiently balance accuracy and computational cost. The LGL bottleneck enables models to achieve high accuracy while minimising computational overhead. In their experiments, \citet{pan2022edgevits} found that EdgeViT-S consistently outperformed other lightweight models in terms of accuracy-latency and accuracy-energy trade-offs. For example, EdgeViT-S achieved a top-1 accuracy of 81.0\% on ImageNet with a latency of 85.3$\pm$3.9 ms on Samsung Galaxy S21, positioning it as a leading choice for mobile vision applications.

Mobile transformer architectures, including MobileViT and EdgeViTs, have proven effective in numerous real-world applications. For example, \citet{george2024edgeface}  utilised mobile transformers for facial recognition tasks in low-power smart devices, including smartphones and smart doorbells. The model's transformer-based architecture showed exceptional performance in facial recognition with minimal power consumption, rendering it an ideal solution for battery-operated edge devices. This application demonstrates the capacity of mobile transformers to perform complex vision tasks, such as facial recognition, in real-time without relying on cloud processing, thereby safeguarding privacy and security. Mobile transformers have been employed in healthcare for remote patient monitoring systems. \citet{park2022self} implemented a mobile transformer-based model for the real-time analysis of medical images, specifically for identifying abnormalities in chest X-rays and CT scans.

Another practical application is in autonomous retail, where mobile transformers are utilised for object detection in cashierless checkout systems. \citet{pan2022edgevits} illustrated the application of EdgeViTs for real-time product identification with negligible latency in low-power embedded systems deployed in retail settings. These systems depend on efficient and precise object detection to guarantee seamless customer experiences, and the transformer-based architecture enabled them to surpass conventional CNN models while preserving a minimal computational footprint.} Mobile transformers have also been explored for augmented reality (AR) applications, especially for real-time object tracking. \citet{chen2022edgevit} utilised EdgeViT to facilitate augmented reality systems in tracking multiple objects within dynamic environments with minimal latency. The low latency and high accuracy of mobile transformers demonstrate their potential to improve mobile gaming and augmented reality experiences on smartphones, which often have constrained computational resources.

\subsection{Neural Architecture Search}
Neural architecture search (NAS) is a subfield of DL that focusses on automating the design of neural network architectures \citep{Elsken2019}. Traditional neural network architecture design often involves human expertise and manual trial and error, which can be time consuming and may not always yield the best results. NAS aims to address these limitations by using search algorithms to discover optimal neural network architectures for specific tasks. We briefly discuss some notable NAS approaches and architectures (Table~\ref{tab:NAS}). For a comprehensive overview of NAS, the readers can refer to the survey by \citep{Elsken2019}.

PNASNet (Progressive Neural Architecture Search Network) \citep{liu2018progressive}, introduced in 2017, is a NAS method that employs a progressive search strategy. PNASNet's progressive search strategy starts with a smaller, simpler network and incrementally increases its complexity. This method ensures that the search process remains computationally feasible even as the architecture grows. For example, PNASNet achieved competitive performance in the CIFAR-10 dataset with significantly fewer computational resources compared to other NAS methods. Specifically, PNASNet was able to achieve a top-1 error rate of 3.41\% on CIFAR-10 while usingonly 400 GPU hours for the architecture search, compared to NASNet's 2,000 GPU hours \citep{liu2018progressive}. One of the strengths of PNASNet is its ability to balance accuracy with computational cost. During the search process, both these factors are considered, ensuring that the resulting architecture is not only accurate, but also efficient. For example, in the study by \citet{liu2018progressive}, PNASNet achieved a top-1 accuracy of 82. 9\% in CIFAR-10 with significantly lower computational resources compared to other architectures such as  AmoebaNet \citep{real2019regularized}, which required much higher computational costs to achieve similar accuracy levels.

MNASNet (Mobile Neural Architecture Search Network) \citep{tan2019mnasnet}, proposed in 2019, is designed for mobile and edge devices with limited computational resources. MNASNet's architecture search is guided by a multi-objective optimisation framework that explicitly incorporates latency into the optimisation process. This ensures that the resulting architectures are scalable and can be efficiently deployed on devices with varying computational capabilities. For example, \citet{tan2019mnasnet} demonstrated that MNASNet achieved a top-1 accuracy of 75.2\% on ImageNet with 312 billion MAdds, making it more efficient than MobileNetV2 \citep{sandler2018mobilenetv2}.

FBNet (Facebook Network) \citep{wu2019fbnet}, also from 2019, is an AI-assisted NAS approach developed by Facebook that targets efficient neural network architectures. FBNet's architecture search uses a factorised approach that decouples the selection of operations from their connectivity. This makes the search space more manageable and allows for efficient exploration of potential architectures. For instance, \citet{wu2019fbnet} demonstrated that FBNet achieved significant improvements in both performance and efficiency compared to manually designed models. Specifically, FBNet achieved a top-1 accuracy of 74.9\% on ImageNet with 375 million FLOPs, which is $1.5\times$ more efficient than MobileNetV2 \citep{sandler2018mobilenetv2}.

AmoebaNet \citep{real2019amoebanet}, introduced in the same year, is known to achieve state-of-the-art results in NAS using an evolutionary algorithm. Explores a wide search space, allowing for the discovery of complex and effective architectures. Together with other NAS approaches, it has contributed to the advancement of DL by automating the design of neural networks and improving their performance in various tasks. NAS continues to play a key role in obtaining networks that are more efficient, effective, and adaptable to various applications and hardware constraints. For example, \citet{real2019amoebanet} demonstrated that AmoebaNet achieved a top-1 accuracy of 83.9\% on ImageNet, which was state of the art at the time. The scalability of AmoebaNet is evident in its ability to generate architectures that perform well on both small datasets like CIFAR-10 and large datasets like ImageNet. AmoebaNet achieves a high level of accuracy using an evolutionary algorithm that optimises both performance and computational cost. For example, AmoebaNet achieved a top-1 accuracy of 96.7\% on CIFAR-10 and 83.9\% on ImageNet.

In real-world applications, edge computing models have to trade-off various aspects such as accuracy, latency, memory consumption, and power efficiency. An example is the work of \cite{cereda2023deep}, in which a lightweight model for autonomous drone navigation was designed using NAS. This model adeptly balanced precision and computational efficiency, enabling drones to execute real-time object detection and navigation in dynamic environments without dependence on cloud-based processing, thereby illustrating the relevance of NAS to practical edge computing applications. \cite{odema2021eexnas} used NAS in a study optimising models for wearable medical technology. The architecture, developed via NAS, was specifically designed for real-time monitoring of physiological signals, including heart rate and oxygen levels, on edge devices. Through the implementation of techniques such as quantization and network pruning, they effectively diminished the model's size and power consumption while preserving high accuracy, making it suitable for continuous monitoring in low-power settings such as smartwatches and fitness trackers.} In the automotive industry, NAS was used by \cite{gupta2024efficient} to optimise models for advanced driver assistance systems (ADAS) and autonomous driving. Their NAS-generated models were implemented on in-vehicle edge devices, where the capacity to process visual information efficiently and rapidly was essential. The research indicated that NAS could generate architectures that satisfied the stringent latency and energy efficiency criteria essential for real-time decision-making in vehicles.

\begin{table}[!t]
\centering
\resizebox{\textwidth}{!}{%
\begin{tabular}{l>{\raggedright}p{6cm}>{\raggedright}p{6cm}>{\raggedright\arraybackslash}p{6cm}}
\toprule
\textbf{Method} & \textbf{Advantage} & \textbf{Disadvantage} & \textbf{Characteristic} \\
\midrule
PNASNet & Progressive search strategy. & Computational cost may still be high. & Uses progressive growth in architecture. \\
\midrule
MNASNet & Designed for mobile and edge devices. & May require reinforcement learning expertise. & Uses reinforcement learning for search. \\
\midrule
FBNet & Factorised search space for flexibility. & Trade-off between performance and efficiency. & Focuses on balancing performance and efficiency. \\
\midrule
AmoebaNet & Achieved state-of-the-art results. & Evolutionary algorithm can be computationally intensive. & Uses an evolutionary algorithm for search. \\
\bottomrule
\end{tabular}}
\caption{Summary of NAS methods.}
\label{tab:NAS}
\end{table}

\section{Computer Vision Edge Applications}
\label{sec:cvapps}
With the advancement of computer vision, lightweight models, and model optimisation techniques, an increasing number of edge applications are being explored across various types of edge devices. Here we survey several primary application domains (Fig.~\ref{fig:application_example1}): autonomous driving and intelligent transportation, smart manufacturing and industrial automation, agriculture and crop monitoring, retail and shelf monitoring, environmental monitoring and wildlife conservation, smart cities and public safety, and sports analytics. Emphasis is placed on applications that leverage the real-time processing capabilities of edge devices to enhance operational efficiency and decision-making. By reviewing nonmedical applications of edge deep learning, we aim to highlight their potential contributions to advancing medical diagnostics.

\subsection{Autonomous Driving and Intelligent Transportation Systems}
Autonomous driving and intelligent transportation are critical domains that leverage advanced edge computer vision technologies to improve vehicle autonomy and traffic management. These technologies contribute significantly to real-time object detection, tracking and classification, which are vital to the safety and efficiency of transportation systems.

\begin{figure}[!t]
    \centering
    \includegraphics[width=\textwidth]{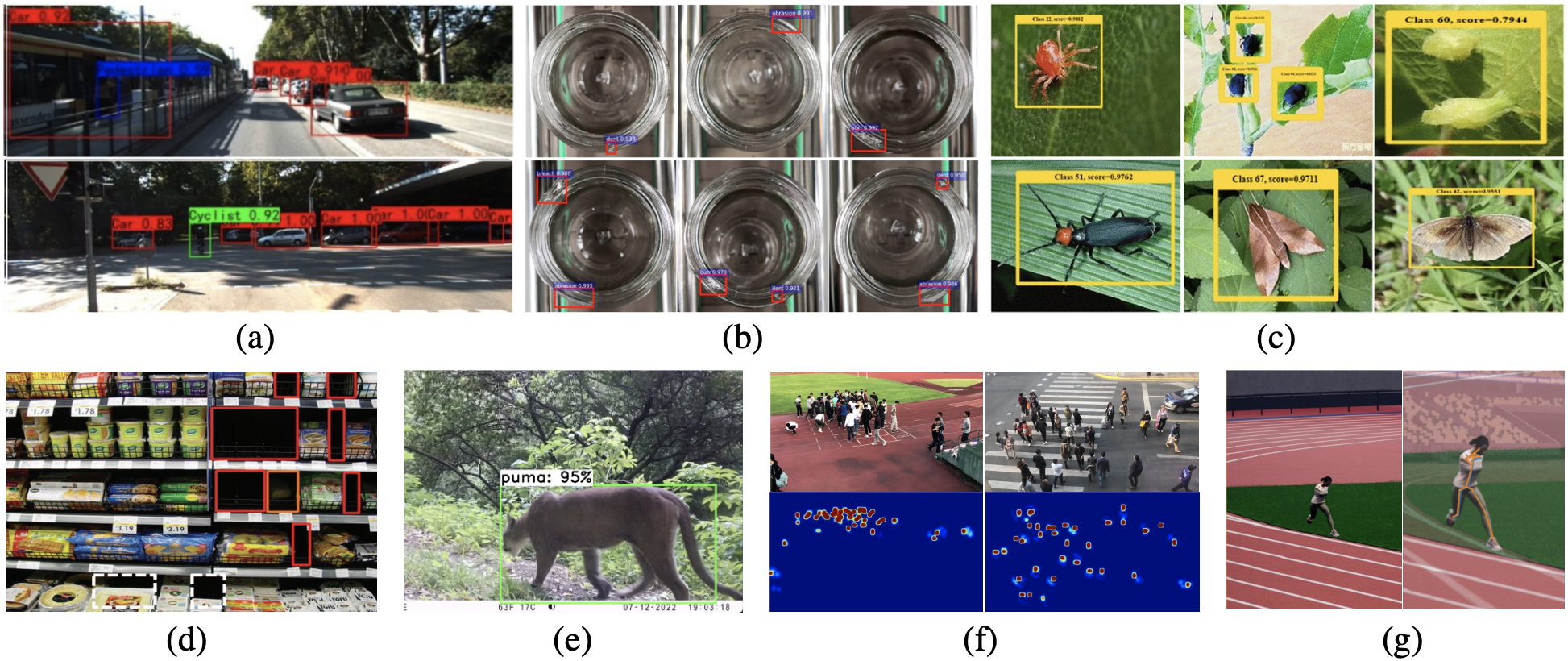}
    \caption{Examples of computer vision edge applications. (a) Detecting complex dynamic driving environments by fusing RGB-camera and LiDAR data \citep{liu2023real}. (b) Detecting defects such as cracks and dents on seal surfaces of containers in filling lines using a lightweight network based on MobileNet \citep{li2018research}. (c) Identifying and classifying pests in crops via drones \citep{albattah2023custom}. (d) Monitoring out-of-stock situations in retail environments using mobile robots \citep{de2023autonomous}. (e) Detecting wildlife using lightweight models deployed on Raspberry Pi \citep{tulasi2023smart}. (f) Improving automatic people counting in smart city environments through surveillance videos \citep{avvenuti2022spatio}. (g) Analyzing human motion through edge photographic equipment \citep{baumgartner2023monocular}.}
    \label{fig:application_example1}
\end{figure}

\cite{wang2021real} proposed an improved object detection method based on YOLOv4 \citep{bochkovskiy2020yolov4} tailored for autonomous vehicles, focusing on the trade-off between speed and accuracy. Their method improved detection accuracy and inference speed by incorporating modifications in the backbone, neck, and predictor head of the YOLOv4 architecture. It showed a notable increase in average accuracy in the KITTI \citep{geiger2013vision} and BDD \citep{yu2018bdd100k} datasets, demonstrating its potential for real-time applications in autonomous driving. Additionally, \cite{mauri2022lightweight} proposed a lightweight CNN based on YOLOv5\footnote{Ultralytics YOLOv5. \url{https://github.com/ultralytics/yolov5}} for real-time 3D object detection in road and railway environments, improving on real-time embedded constraints and enhancing its applicability.

Another approach, by \cite{liu2023real}, tackled the challenges of object detection by fusing data from RGB cameras and LiDAR. Their method uses the richness of semantic information from cameras with the accurate depth data from LiDAR sensors, improving the reliability of object detection units. The proposed siamese network structure and feature-layer fusion strategy significantly enhanced detection performance, particularly in complex and dynamic driving environments.

In the domain of traffic monitoring, \cite{wan2022edge} introduced an edge computing-based video segmentation method. They optimised traditional video processing techniques to reduce redundancy and enhance real-time performance. By implementing spatiotemporal interest points and multimodal linear feature combinations, their method efficiently segments traffic videos, enabling effective monitoring and management of urban traffic.

Moreover, \cite{fernandez2021robust} developed TrafficSensor, a robust real-time traffic surveillance system using DL. The system uses a calibrated camera to track and classify vehicles on highways under various conditions, including poor lighting and adverse weather. Using advanced neural networks and tracking algorithms, TrafficSensor provides reliable data for infrastructure planning and traffic management, showcasing the capabilities of edge computing in intelligent transportation systems. Furthermore, the method introduced by \cite{zou2022real} involves shifting video analysis tasks to edge devices. This aims to facilitate low-latency and high-accuracy experiences in real-time traffic monitoring for connected vehicle networks.

\subsection{Smart Manufacturing and Industrial Automation}
In modern manufacturing, edge computer vision has made significant strides in smart manufacturing and industrial automation. These advances are crucial to improving the precision, efficiency, and safety of manufacturing processes. A core aspect of this technological evolution is the development of sophisticated real-time defect detection systems, critical to maintaining product quality and optimising manufacturing workflows.

\cite{yi2022defect} tackled the challenge of detecting bead-toe defects in tyre X-ray images. They proposed an innovative lightweight semantic segmentation network that first extracts texture features from various tyre regions and then employs a decoder to fuse these features. This effectively reduces the dimensions of feature maps to pinpoint bead-toe positions. 

For the detection of surface defects in sanitary ceramics, \cite{hang2022surface} proposed a lightweight real-time defect detection network that uses MobileNetV3 \citep{koonce2021mobilenetv3} as the backbone. The network achieves multiscale detection of surface defects through a multilayer feature pyramid and combines a region proposal network with an anchor-free method. The detection head incorporates a channel attention structure and a low-level mixed feature classification strategy is employed to achieve higher accuracy in defect classification. This method demonstrated significant improvements, including at least 22.9\% faster detection and 35.0\% higher average precision, while reducing memory consumption by at least 8.4\% compared to classical detection methods such as SSD \citep{liu2016ssd}, YOLO V3 \citep{redmon2018yolov3}, and Faster R-CNN \citep{girshick2015fast}.
 
Furthermore, \cite{chen2022detection} presented an improved YOLOv3 model to detect surface defects on polarisers. They modified the model by replacing DarkNet53 \citep{redmon2016you} with MobileNet \citep{howard2017mobilenets} as the backbone, significantly reducing the number of network parameters and enhancing detection speed. The introduction of the Mixed Convolution Efficient Attention (MECA) module further improved detection accuracy. The modified network showed a detection speed of 121 frames/second and an mAP of 89.8\%, demonstrating a 44\% increase in speed compared to the standard YOLOv3.

To detect defects in industrial insert moulding using X-ray images, \cite{wang2021lightweight} developed a lightweight deep network based on the YOLOv5 model,  incorporating the Ghost module to reduce the model size and a transformer module for enhanced spatial multiheaded attentional feature extraction. The network achieved an mAP of 93.6\% and showed robustness under different luminance and noise conditions.

\cite{xia2019ssim} proposed SSIM-NET, a new print circuit board (PCB) defect detector combining the structural similarity index (SSIM) and MobileNetV3 \citep{howard2019searching}. This two-stage detection algorithm first uses SSIM to identify suspicious regions, less susceptible to environmental factors such as variation in illumination. In the second stage, MobileNetV3 with a binary loss and focal loss is used to classify these regions, significantly reducing computational costs. The approach achieved 12$\times$ speed increase over Faster-RCNN without sacrificing accuracy.

Additionally, \cite{dai2020soldering} introduced an integrated detection framework for solder joint defects in PCBs using automatic optical inspection (AOI). They employed a generic DL method for localisation that is easily adaptable to various PCB configurations and soldering technologies. For classification, they introduced an active learning method to minimise the labeling workload. Their approach demonstrated fast, accurate localisation, and high classification accuracy with minimal user input.

To achieve accurate and real-time surface defect detection, \cite{li2018research} optimised the SSD network structure by integrating it with MobileNet, resulting in a method called MobileNet-SSD. This method was particularly effective in detecting typical defects such as breaches, dents, and burrs on container sealing surfaces in filling lines. 

\cite{bonam2023lightweight} investigated the application of lightweight CNN models for product defect detection in the manufacturing industry, focusing on utilising edge deep learning to address the issues of labour intensity, error-proneness, and unreliability in detecting defects in fabrics, surfaces, and castings. Specifically, the authors employed lightweight models such as MobileNetV2, ShuffleNetV2, and CondenseNetV2 for defect detection in these areas and successfully deployed these models on edge devices with limited computational capabilities, such as the NVIDIA Jetson Nano. The results demonstrated that these lightweight models achieved high accuracy and efficiency in defect detection, with MobileNetV2 achieving a test accuracy of 96.87\% on the fabric defect dataset, ShuffleNetV2 achieving 99\% on the surface defect dataset, and CondenseNetV2 achieving 98.08\% on the casting defect dataset. This showcases the potential of lightweight CNN models to enhance the efficiency and reliability of defect detection in manufacturing environments.

These edge DL algorithms are increasingly being integrated with existing manufacturing processes, thereby enhancing precision and efficiency. However, this integration also presents certain challenges. Ensuring compatibility with legacy systems can be complex, often requiring modifications or upgrades to existing infrastructure. Additionally, the initial cost of implementing and deploying these edge deep learning algorithms can be substantial. Despite these challenges, the benefits of edge DL, including reduced waste, improved product quality, and increased production efficiency, make it an ideal choice for modern manufacturing \citep{nain2022towards}. Moreover, some continual learning approaches \citep{zhang2023toward,dekhovich2024continual,chang2024unified} confer additional advantages over traditional methods, making edge DL particularly well-suited for application in the manufacturing sector.

\subsection{Agriculture and Crop Monitoring}
Edge computer vision is also crucial in precision agriculture as it enables the monitoring of crops and the optimisation of agricultural techniques. Unmanned aerial vehicles (UAVs) equipped with advanced electronics and cameras capture live images of agricultural fields. Computer vision algorithms then analyse these photographs to assess the condition of crops (identifying overall health and areas of stress, disease, or nutrient deficiencies), detect pests (enabling early intervention and targeted treatments), and optimise watering (based on soil moisture levels and crop health indicators). This empowers farmers to make decisions based on data, resulting in higher crop production and better use of resources.

\cite{dang2020uav} introduced an innovative method for the automatic detection and classification of diseases in radish fields using UAVs. It uses UAV-mounted cameras to capture high-quality field images, which are analysed by combining colour and texture feature extraction. The method employs K-means clustering to segment radish regions, followed by the use of a fine-tuned GoogleNet to detect early stages of Fusarium wilt. This offers a rapid and accurate alternative to manual inspection methods.

To identify and categorise insect pests in crops, \cite{albattah2023custom} introduced an automated system based on UAVs to address the challenges of manual inspection and the need for timely pest management. They employed a lightweight UAV and a custom CornerNet \citep{law2018cornernet} with DenseNet100 \citep{huang2017densely} as the foundational network. Specifically, the method involves three stages: acquiring the region of interest through sample annotations for model training, applying DenseNet100 for deep keypoint computation in the custom CornerNet, and finally employing the CornerNet model to identify and categorise various insect pests. The DenseNet network enhances feature representation, helping CornerNet to effectively detect insect pests as paired key points.} The method was evaluated using the standard IP102 benchmark dataset \citep{wu2019ip102}, which demonstrated its effectiveness and accuracy in identifying target insects, thus providing an essential tool to strengthen crop management and food yield through timely pest detection and intervention.

\cite{albuquerque2020water} proposed a DL method designed to automatically identify water from aerial footage captured by UAVs using the Mask R-CNN \citep{he2017mask}. The method aims to improve the inspection and maintenance of irrigation systems, potentially reducing time and costs. The ability to accurately detect water in image frames allows for early identification of malfunctioning irrigation nozzles, crucial to correctly implementing crop field hydration plans. Such malfunctions can lead to insufficient or excessive watering, compromising the effectiveness of irrigation plans. Preliminary results demonstrated the feasibility of using advanced UAVs and neural networks in smart irrigation systems, offering a promising solution to ensure crop quality and productivity amid increasing food demand.

The application of UAV-acquired multispectral and thermal infrared imagery in precision irrigation management in a Cabernet Sauvignon orchard was showcased by \cite{zuniga2017high}. They evaluated the efficacy of these images in assessing various subsurface irrigation configurations at different depths and rates compared to standard surface irrigation. While no significant differences in fruit yield were observed between different irrigation techniques (pulse versus continuous) or depths, the study did find significant yield variations caused by  deficient irrigation. Strong correlations were seen between vegetation indices (NDVI, GNDVI) and canopy temperature with both fruit yield and leaf stomatal conductance, demonstrating the potential of UAV-acquired imagery in real-time crop stress assessment. This work emphasises the efficacy of using thermal imagery as a rapid tool to estimate leaf stomatal conductance, crucial to optimising irrigation schedules and enhancing water use efficiency in viticulture.

An innovative method combining UAVs, multispectral imaging, and YOLOv3 to evaluate phenotypic traits in citrus crops was introduced by \cite{ampatzidis2019uav}. The method overcomes the limitations of traditional plant breeding evaluation by offering low-cost, automated, high-throughput phenotyping. Using YOLOv3, the method detects, counts, and geolocates trees and tree gaps, categorises trees based on canopy size, develops individual tree health indices, and evaluates citrus varieties and rootstocks. In a study involving 4,931 citrus trees, the method achieved high precision and recall rates of 99.9\% and 99.7\%, respectively, for tree detection and counting, 85.5\% overall accuracy for canopy size estimation, and 100\% precision and 94.6\% recall for detecting and locating tree gaps. The method significantly advances genomic selection and cultivar development in agriculture. The authors did not further optimise the YOLOv3 model, but rather deployed it directly on UAVs, demonstrating the feasibility of using an existing efficient model on resource-constrained edge devices. This highlights the inherent balance between performance and computational efficiency of YOLOv3.

It is noteworthy that the effectiveness of edge deep learning methods in different agricultural environments may vary due to factors such as crop type, environmental conditions, and specific agricultural practices \citep{mcenroe2022survey}. For instance, using drones in dense crop canopies may require more sophisticated methods or higher resolution images to maintain detection accuracy \citep{su2023ai}. Nevertheless, these edge deep learning methods generally contribute to enhanced productivity and resource efficiency. Moreover, these methods demonstrate the potential for adaptability and scalability of drone-based solutions across various agricultural settings, ranging from small farms to large-scale operations. Particularly, large-scale farms can leverage economies of scale to reduce deployment costs and further enhance operational efficiency through standardisation.

\subsection{Retail and Shelf Monitoring}
Retailers utilise computer vision technology to monitor shelves and manage inventory. Cameras are installed on shelves to capture photos in real-time. Computer vision algorithms analyse the photos to monitor stock levels, detect misplaced items, and generate warnings for restocking. This optimises inventory management and increases the overall shopping experience for customers. We discuss several notable studies that have explored the integration of computer vision for retail and shelf monitoring applications.

\cite{lachhab2023deep} proposed the use of edge computer vision techniques, particularly edge-based object counting models, to automate the traditional manual inventory management process. Automation includes capturing images of fruits and vegetables shelves, identifying boxes and their categories, and then using DL counting models to estimate the number of items in each box. This process aims to optimise store operations through continuous monitoring and analysis. The study employed object detection and density estimation methods, evaluating object counting approaches in four data scenarios: supervised learning, semi-supervised learning, few-shot learning, and zero-shot learning. Key findings include the YOLO model (especially YOLOv5) performing well in supervised learning due to its balance between speed and size. In semi-supervised learning, the Efficient-Teacher method enhanced the performance of the YOLO model using limited labeled data. Zero-shot learning, particularly the CLIP-Count method, is recommended in environments with limited data but ample computational resources, striking a balance between speed and error rate. 

Advances in edge computer vision were also utilised by \cite{kanjula2022people}, who introduced a pioneering AI-based people counting system for retail analytics. This cost-effective solution calculates conversion rates by correlating the number of visitors with actual transactions, providing valuable insights for retail optimisation, such as security checks and intelligent queuing. The project utilised Intel's OpenVINO toolkit for optimising neural network inference, demonstrating how advanced sensor technology and edge AI can transform the traditional retail environment. This approach not only improved the accuracy of people counting but also aligned with the trend of integrating AI-driven solutions into retail to achieve growth equivalent to online data analysis capabilities.

\cite{fan2021cmss} presented a low-power, cost-effective DL method using small IoT cameras to monitor the shelf status in retail environments, termed as CMSS (Camera Monitoring Store Shelves). This method addressed the inefficiency of manual inspection for restocking shelves, crucial for maintaining turnover in retail settings. The system operates on ultra-low-power FPGA chips, ensuring minimal power consumption at milliwatt levels, with total system power consumption below 6 mW. Their experimental results demonstrated the effectiveness of the system, with a maximum lateral recognition distance of 40 centimeters on shelves of the same width. Compared to existing methods, this solution offers significant advantages, including lower cost, reduced power consumption, and ease of maintenance, making it a highly feasible option for large-scale retail applications.

A sophisticated mobile object recognition system tailored for retail environments utilising video frame detection was proposed by \cite{mustafa2005detecting}. The system was specifically designed to monitor and detect activities such as the movement of shopping carts and the operation of cash drawers by analysing the boundaries of these objects. The system, manually trained through pre-recorded video sequences, reliably identifies specific actions, aiding in the surveillance of potential suspicious activities. It emphasises the importance of understanding the specific context of the monitored activities, involving either human or nonhuman entities, and adapts to various surveillance requirements, including single or multiple camera setups. This method provides retail store owners with a practical solution to effectively filter and analyse surveillance footage, thereby enhancing the security and operational oversight in retail environments.

To address out-of-stock (OOS) issues in retail supermarkets, \cite{achakir2023automated} introduced an edge computer vision system based on Faster R-CNN and the monocular depth estimation model MiDas to identify OOS products by analysing images of store shelves. The integration of Affine-SIFT descriptors with Faster R-CNN improved the accuracy of detecting products, especially in challenging categories like coffee and wine. This automated method, tested in multiple stores, not only accurately detects stockouts but also facilitates timely restocking, ultimately improving sales and customer satisfaction. The system demonstrated a significant shift from manual to automated inventory monitoring, highlighting the potential of DL technologies in optimising retail operations and inventory management.

\cite{de2023autonomous} proposed ROSCH, a humanoid autonomous mobile robotic platform based on the robotic operating system (ROS) framework, designed to detect OOS items on supermarket shelves. Specifically, ROSCH utilises the humanoid robot Pepper, powered by the NVIDIA Jetson Xavier NX, to identify empty and partially empty shelves using the YOLOv6 model. YOLOv6 \citep{li2022yolov6} is a single-stage anchor-free object detection model known for its excellent generalisation capabilities and fast convergence, particularly when training data is limited. To further enhance inference speed, the researchers optimised the model using the TensorRT framework. Integrating navigation with YOLOv6, ROSCH autonomously navigates supermarket aisles and notifies human operators to expedite restocking. The ROSCH system was tested in a supermarket in Salerno, Italy, demonstrating impressive performance, with its OOS detection being eight times faster than manual detection. The system aims to reduce sales losses due to OOS situations and improve the overall efficiency of supermarket operations.

Edge deep learning technology offers numerous benefits for retail environments, such as real-time inventory monitoring, which helps optimise inventory management processes while reducing labour costs. However, the initial investment in infrastructure, including devices and software, can be substantial. Despite this, the potential return on investment (ROI) remains considerable, as these technologies can enhance sales through improved inventory management and customer insights, ultimately optimising store operations and increasing profitability \citep{kellermayr2022digitalization}. Furthermore, it supports unmanned retail by providing more accurate shelf monitoring and automated customer services, offering customers a more efficient and convenient shopping experience. \citep{xu2020design}

\subsection{Environmental Monitoring and Wildlife Conservation}
Edge computer vision is also used in the field of environmental monitoring and wildlife conservation. Wildlife is documented through the use of camera traps (i.e., remotely triggered cameras) placed in natural habitats, which noninvasively collect both images and videos (see Fig.~\ref{fig:environmental_monitoring} for a camera trap-based edge device setup used for monitoring wildlife and conserving natural habitats). Computer vision systems deployed at the camera traps process these data to monitor animal movements, recognise endangered species, and detect any abnormal behaviour. Real-time monitoring facilitates ecological studies and improves the efficacy of wildlife protection efforts.

\begin{figure}[!t]
    \centering
    \includegraphics[width=0.7\textwidth]{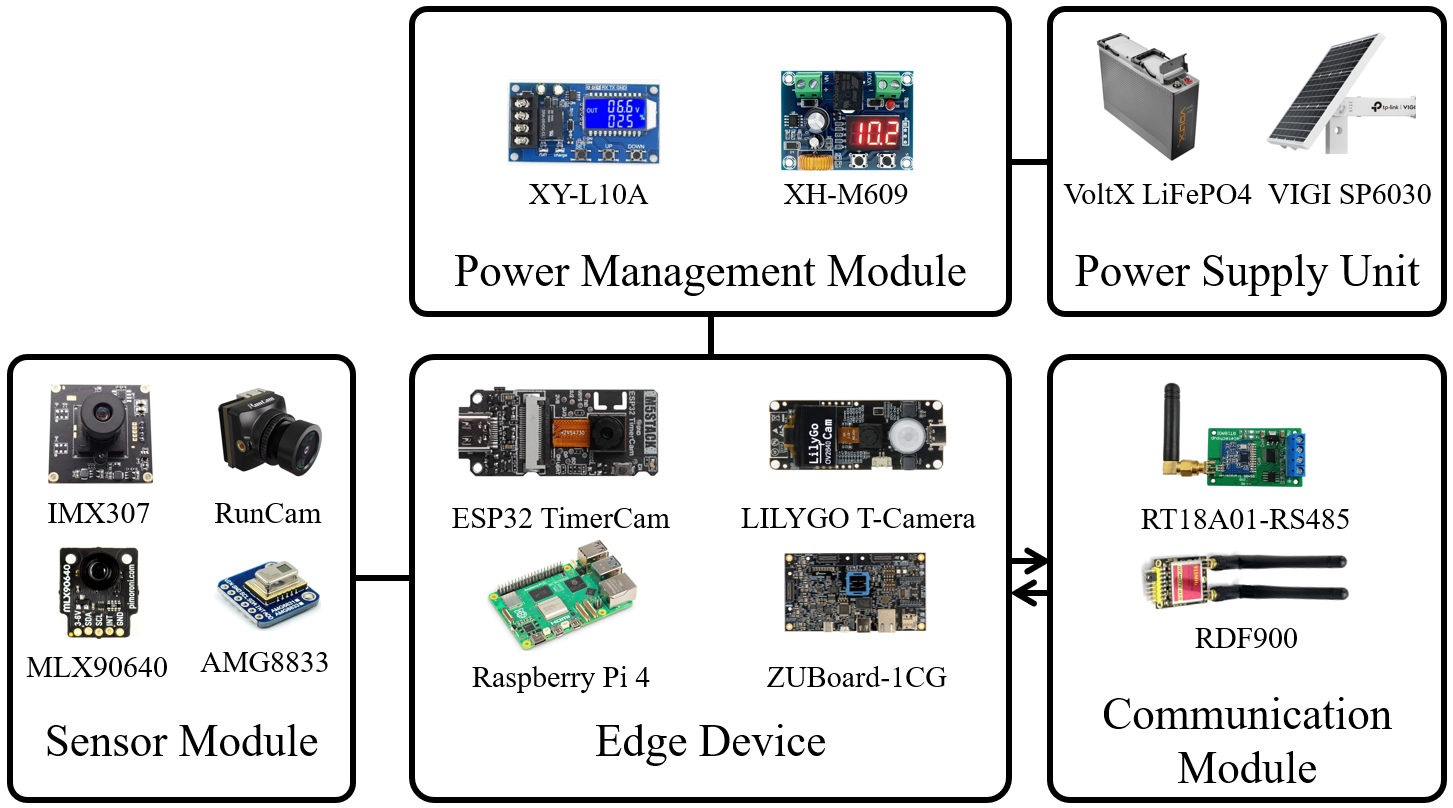}
    \caption{Schematic of an edge device for wildlife monitoring and conservation.}
    \label{fig:environmental_monitoring}
\end{figure}

\cite{gotthard2023edge} used edge computer vision for wildlife conservation by employing object detection and classification models on edge devices equipped with camera sensors to monitor endangered species and detect potential intruders, primarily in the Ngulia sanctuary in Africa. The study evaluated three object detection models (SSD, FOMO, MobileNetV2, and YOLOv5) on three different microcontrollers. A key innovation is the deployment of wireless updates, enabling edge devices in remote locations to collect field data and iteratively improve through an active learning pipeline. This approach demonstrates the feasibility and effectiveness of edge computer vision in real-world conservation efforts.

\cite{simoes2023deepwild} developed a framework that employs MegaDetector, a computer vision model created by Microsoft’s AI for Earth project, which is widely used in camera trap-based environmental monitoring and wildlife conservation. MegaDetector utilises a two-stage Faster R-CNN process to identify and classify objects in images by first detecting regions containing objects and then classifying these objects with confidence scores. The authors fine-tuned the Faster R-CNN model with an Inception-ResNet-v2 backbone to detect and classify 13 wildlife species relevant to the Parc National du Mercantour (PNM) in France. Under various environmental conditions, DeepWILD accurately detected, classified, and counted these 13 species. At an IoU of 0.5, the method achieved a mean Average Precision (mAP) of 96.88\%, significantly enhancing the efficiency and accuracy of wildlife population estimation.

Focusing particularly on accurately determining the population density of deer in specific areas, \cite{arshad2020my} introduced a novel method for wildlife monitoring and counting. Their method, utilising CNNs, edge computing, and online tracking, addressed various challenges in automatic animal counting, such as wildlife movement, light fluctuation, and the issue of recounting the same animal in different images. The authors highlighted the shortcomings of traditional wildlife monitoring methods, including the high cost and logistical challenges of manual observation, and the limitations of trap cameras in manual data extraction and their inability to distinguish repeated appearances of the same animal. To overcome these issues, \citet{arshad2020my} proposed a video-based detection and tracking system capable of operating under various lighting conditions, accurately counting and tracking the movement of deer within the camera's field of view. This method not only enhances the accuracy of wildlife counting but also provides valuable data for understanding animal activities and behaviors, proving to have immense potential in conservation and ecological research.

\cite{tulasi2023smart} presented an energy-efficient solution to enhance wildlife monitoring by deploying a lightweight network on existing camera traps. Specifically, their method employs the Raspberry Pi Zero 2W for on-site data processing and combines it with a detection and classification two-stage pipeline based on MobileNetV2. This approach offers timely wildlife detection and alerts, addressing major challenges of traditional camera traps, such as delayed data retrieval and processing. This not only improves the accuracy and efficiency of wildlife monitoring but also provides a scalable and sustainable solution for ecological research and conservation efforts, demonstrating its practical application through successful field deployment.

Another method for wildlife monitoring and analysis based on camera trap network captures was proposed by \cite{gupta2022deep}. Their method addresses the challenge of cluttered images in camera trap data, which typically leads to low detection rates and high false discovery rates in animal monitoring. To address this, the authors leveraged a camera trap database containing candidate animal proposals generated using multilevel graph cuts in the spatiotemporal domain. These proposals were then validated to differentiate animals from the background. The core of their method is the use of self-supervised CNNs to develop an animal detection model. Their extensive experimental results show that the proposed detection model achieves a high precision  on standard camera trap datasets.

Focusing on the needs of animal conservation, particularly for endangered elephants in regions like India, \cite{verma2018wild} presented a solution using deep learning models to prevent human-elephant collisions (HEC). They developed an automated early warning system based on visual cues, employing transfer learning models such as ResNet50, MobileNet, and InceptionV3. These models were tested on a comprehensive dataset, showing high accuracy in detecting elephants, potentially preventing tragic incidents on forest-crossing railway tracks. This study offers a low-cost, accurate, and efficient method for elephant detection, aligning with efforts to mitigate the ecological imbalance caused by increasing wildlife mortality rates.

In conclusion, edge deep learning offers innovative and effective tools for wildlife monitoring and conservation. Various studies have demonstrated its potential in enhancing the accuracy, efficiency, and scalability of monitoring systems, particularly in remote or challenging environments. However, ethical considerations and the potential impacts on wildlife behaviour must be taken into account to ensure that conservation efforts do not inadvertently harm the species they aim to protect. Noninvasive techniques, such as camera traps, can minimise human disturbance, thereby avoiding stress or changes in animal behaviour and movement patterns.

\subsection{Smart Cities and Public Safety}
Edge computer vision also plays a significant role in improving public safety and urban planning within the framework of smart cities. Surveillance cameras, equipped with edge devices, employ real-time video stream analysis to identify abnormal behaviours, monitor traffic patterns, and improve overall security. The decentralised model guarantees prompt reactions to crises and facilitates municipal planning using real-time data. Several studies have explored the diverse applications and advantages of deploying edge computer vision systems in urban environments.

\cite{avvenuti2022spatio} introduced a neural network based on spatiotemporal attention for improving automatic people counting in urban smart city environments using surveillance video. Their method effectively counts and precisely locates people within video frames, embodying significant advancements in practical urban surveillance and crowd monitoring through DL technologies. Leveraging the temporal correlation between video frames, it significantly reduces errors on the FDST benchmark through self-attention connectives and attention-based temporal fusion layers.

Utilising cost-effective off-the-shelf hardware equipped with computer vision capabilities, \cite{di2022embedded} presented an embedded system for monitoring human activities and ensuring safety in critical environments, such as in the context of health emergencies or hazardous workplaces. Their DL-based embedded system performs tasks like people counting, social distance measurement, and personal protective equipment detection. Developed in response to challenges posed by the COVID-19 pandemic, the system operates both indoors and outdoors, reducing the need for manual supervision. Its effectiveness was validated through two novel datasets, one containing images from public squares in Pisa, Italy, and another featuring images with and without personal protective equipment. The results indicate that the system can accurately monitor compliance with safety rules, providing a practical solution for real-time safety monitoring in various scenarios.

Addressing the challenge of real-time facial recognition on resource-constrained devices, \cite{deng2023lightweight} developed an efficient, compact DL model inspired by FaceNet. The model employs one-shot or few-shot learning to achieve effective feature embedding. The study demonstrated the model's effectiveness in recognising occluded faces using grayscale input images in uncontrolled environments, making it suitable for real-time embedded applications like entrance security systems in urban settings.

\cite{raj2023ocularone} presented a study on how drones and computer vision technologies could improve the quality of life for visually impaired persons (VIPs) in smart cities. They explored the potential of these technologies to assist VIPs in navigating both indoor and outdoor environments, enhancing their mobility and safety. Given the large number of people affected by visual impairments globally, the study emphasises the need for innovative solutions to aid VIPs. With the rapid development of smart city infrastructures, the authors suggest drones equipped with advanced computer vision and navigation systems as a promising solution. These drones could serve as mobile assistive tools, replacing traditional methods like guide dogs, and offer additional features like hazard detection and collision avoidance. This approach highlights the integration of emerging technologies into urban environments, aiming to make cities more inclusive and convenient for VIPs.

A significant study on the urban heat island effect in the rapidly urbanising Pearl River Delta region in China was presented by \cite{chen2006remote}. Focusing particularly on the impact on regional climate and socio-economic development, they analysed land use/cover types and brightness temperature using satellite imagery, introducing the Normalised Difference Bareness Index (NDBaI) for improved land analysis. The study specifically examined the rapidly developing city of Shenzhen to understand how its expansion affects temperature distribution and the urban heat island effect. The study provides valuable insights into the impact of urbanisation on local climate, offering crucial data for urban planning and mitigating the urban heat island effects of rapidly developing cities.

\cite{cao2023improved} proposed an improved, lightweight, real-time detection algorithm for drone imagery, addressing the challenges of detecting small objects and reducing computational costs. The algorithm combines MobileNetV3 with the ECA attention mechanism and uses additional prediction heads to enhance small object detection. The algorithm showed outstanding performance and efficiency, making it a valuable tool for drones performing disaster search and rescue missions in urban areas.

\cite{muhammad2018early} proposed a CNN-based fire detection method for surveillance videos. This approach leverages DL to extract features from video frames, identifying early signs of fire. Subsequently, the authors optimised the method using MobileNet, improving inference speed and computational efficiency while reducing model parameters and memory usage, making it more suitable for resource-constrained edge devices \citep{muhammad2018convolutional}. The optimised method maintains high detection accuracy, significantly enhances real-time performance, and broadens its applicability. Experimental results demonstrate its effectiveness in various surveillance environments, quickly detecting flames and smoke during the early stages of fire, thus providing critical time for rescue operations.

\subsection{Sports Analytics}
In sports analytics, edge computer vision is increasingly utilised to provide real-time insights during events. Edge devices process live footage captured by cameras installed in stadiums to generate performance metrics, analyse game dynamics, and monitor player movements. This information improves the overall sports experience for coaches, analysts, and even spectators, as illustrated by several recent studies.

\cite{cui2022application} introduced a federated learning algorithm and a lightweight neural network-enhanced distributed computing model to address the limitations of centralised cloud computing in processing large-scale video surveillance data. This model processes video data at the edge, reducing transmission load and latency. The study also explored human action recognition, particularly in Taekwondo, using behavior detection algorithms combined with sensors to improve training quality. This approach aims to strengthen Taekwondo training methods and ensure safer practice routines. The study explores the feasibility and reliability of distributed computing and DL to sports training and intelligent video surveillance systems.

Focusing on improving the localisation of human-generated events, \cite{cioppa2020context} introduced a novel loss function for action discovery in sports videos, especially football. Their method considers the temporal context around actions and, when applied to SoccerNet, achieved a substantial improvement over the baseline and was applied to ActivityNet for general activity detection. Additionally, the method not only enhanced action recognition in SoccerNet but also demonstrated its versatility and effectiveness in broader activity localisation tasks. This provides advanced tools for deeper match analysis and automated content creation.

Addressing the limitations of monocular 3D human pose estimation (HPE) methods, \cite{baumgartner2023monocular} proposed a novel method for extracting 3D kinematic data from 2D sports videos. Their research introduced partial motion field registration for precise camera calibration, crucial for kinematic analysis in sports like middle and long-distance running. By generating a synthetic dataset in Unreal Engine 5, the study provided new benchmarks for evaluating 3D HPE technologies, paving the way for advanced, large-scale human motion analysis from existing video sources.

\cite{van2023accuracy} performed markerless motion capture using computer vision technology to analyse running techniques. They evaluated the accuracy of DeepLabCut and OpenPose in tracking sagittal plane hip, knee, and ankle kinematics during running and compared them to a marker-based gold standard system. They emphasised the potential of computer vision in enhancing the reliability and effectiveness of running technique analysis.

A four-point camera calibration method for sports video capture was proposed by \citet{zhang2023four}. Their method uses conditional generative adversarial networks (cGANs) to generate semantically segmented video frames and then estimates four key points from a single segmented frame using a regression network. Evaluation on various datasets, including the 2014 FIFA World Cup and National Basketball, showed that this method is superior in accuracy and computational efficiency, and suitable for real-time applications in sports video camera calibration.

\section{Medical Edge Applications}
\label{sec:medapps}
%The infusion of edge computing technologies has significantly improved diagnostic efficiency across various medical domains. The integration of edge devices ranging from low-end to high-performance, coupled with advancements in computer vision, model compression, and efficient lightweight models, has facilitated real-time analysis of medical images and other diagnostic data. This acceleration in the diagnostic process aids in timely medical interventions, creating a conducive environment for point-of-care applications.

\color{black}The application of edge computing technology in computer vision is becoming increasingly mature, demonstrating the unique advantages of edge computing in various scenarios. Recent advancements in edge computing within nonmedical domains have also impacted medical diagnostics, particularly through the adoption of optimised models and hardware initially designed for general computer vision tasks. For example, the YOLO series, as a highly efficient object detection algorithm, widely used in surveillance and autonomous driving, has now been successfully applied to medical tasks, such as polyp detection during colonoscopies \citep{ou2021polyp,redmon2016you}. Similarly, MobileNet, known for its efficiency on edge devices, has shown promising results in diagnosing diseases from medical images, achieving reduced latency without sacrificing accuracy \citep{sait2019mobile,goceri2021diagnosis}.

On the hardware side, edge devices such as the NVIDIA Jetson series and Raspberry Pi have proven capable of supporting medical diagnostic tasks that require real-time processing and low power consumption. For instance, the Jetson series has been utilised in computer vision-based frameworks for detecting skin and cervical cancers \citep{shrivastava2023skin,wang2019recognition}. Additionally, several approaches have integrated Raspberry Pi, smartphones, and lightweight networks for chest CT and dermatology detection, offering a low-cost and efficient diagnostic solution \citep{mieras2018development,dai2019machine,raghavan2020initial,masud2020light,goceri2021diagnosis}. The successful deployment of these devices highlights the potential of edge computing in medical scenarios, enabling efficient processing and diagnostics while delivering reliable results, even under resource-constrained conditions.

The exploration of nonmedical applications in the previous section underscores the potential of edge deep learning technologies. Autonomous driving and intelligent traffic control utilise edge devices to process sensor data locally, enabling real-time decision-making and adaptive responses. These methodologies, developed for efficient real-time processing, can be applied to medical diagnostics and emergency care, where immediate analysis of medical images or patient data is critical. Furthermore, sensor fusion techniques can assist in the multimodal data analysis and processing required in medical scenarios. Additionally, the long-term, low-power monitoring solutions used in agriculture can be adapted for continuous medical monitoring and remote diagnostics. In retail, edge deep learning supports intelligent inventory management and real-time customer interaction. Similarly, these methods can improve drug inventory management in medical facilities, ensuring efficient healthcare operations.

In summary, insights gained from nonmedical domains can inspire new approaches and improve existing methodologies in medical diagnostics, paving the way for a more integrated and effective use of edge deep learning in healthcare.\color{black}\ Here we discuss various examples of medical edge computing (Fig.~\ref{fig:application_example2}), including for the diagnosis of gastrointestinal, pulmonary and thoracic, and dermatological diseases, as well as for pathological images analysis and telemedicine applications, with a focus on ensuring real-time and efficient diagnostic services.

\subsection{Gastrointestinal Diagnosis}
Gastroenterological diagnostics has benefited from advances in computer vision and edge computing technologies, as these assist in the real-time analysis and interpretation of endoscopic images for early detection, treatment, and monitoring of gastrointestinal diseases. Currently, gastroenterological diagnostics primarily rely on computer-aided diagnosis (CAD) systems, deployed on intermediate edge devices, primarily identifying and classifying anatomical markers through analysis of endoscopic images.

\begin{figure}[!t]
    \centering
    \includegraphics[width=\textwidth]{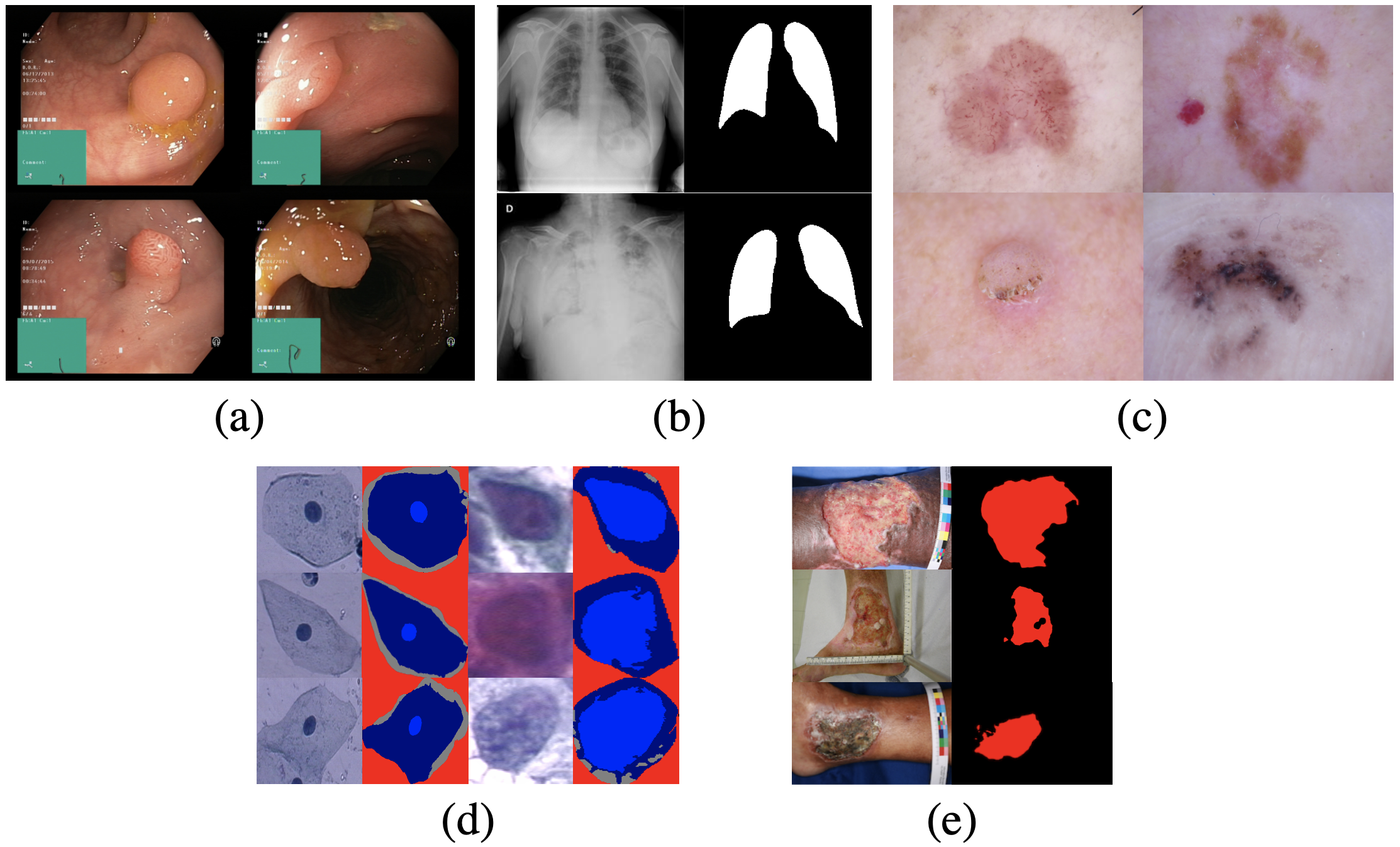}
    \caption{Examples of edge applications in the medical field based on computer vision. (a) Detecting intestinal polyps through endoscopes and DL-based computer-aided diagnosis \citep{urban2018deep}. (b) Rapid screening method for COVID-19 deployed on Intel/Movidius Neural Compute Stick 2 \citep{liu2023edgemednet}. (c) Fast identification of melanoma through mobile edge devices \citep{dai2019machine}. (d) Detection of cervical cancer cells through embedded edge devices \citep{wang2019recognition}. (e) Detection of chronic skin ulcers using portable edge devices \citep{chino2020segmenting}.}
    \label{fig:application_example2}
\end{figure}

Accurate polyp detection is crucial for the early identification of adenomas and for mitigating the risks associated with cancer progression. However, traditional colonoscopy examinations have a polyp miss rate as high as 25\% \citep{corley2014adenoma}. Factors such as the varying skill levels of gastroenterologists, along with physical and mental fatigue, contribute to these oversights, leading to significant variability in adenoma detection rates (ADR) among practitioners \citep{kumar2017adenoma,leufkens2012factors}. CAD systems employing DL algorithms can reduce the polyp-miss rate, particularly for endoscopists with lower ADR \citep{wang2019real,wang2020lower,goceri2024polyp}. 

However, a challenge in CAD system-based polyp detection is the difficulty of deploying models on resource-constrained endoscopic hardware for real-time clinical prediction. DL models are typically trained and tested on dedicated deep learning accelerators, making it impractical to deploy them directly on medical diagnostic devices. Medical diagnostic devices often suffer from low performance due to weaker DL accelerators or the absence of specialised AI accelerators. For models that are difficult to deploy on low-performance devices, forced deployment resulting in low frame rates per second (FPS) still severely affects the efficiency of CAD. Therefore, from a systems perspective, lightweight models play a crucial role in facilitating high-accuracy and real-time polyp diagnosis.

A representative approach is PolypSeg+ \citep{wu2022polypseg+}, which utilises a lightweight architecture to enhance real-time application in clinical settings. It incorporates Adaptive Scale Context (ASC) and Efficient Global Context (EGC) to improve feature differentiation and detail preservation. This method effectively addresses several challenges in polyp segmentation, such as significant variations in polyp size and shape, low contrast between polyps and surrounding tissue, and blurred polyp boundaries. The model has been tested on the Kvasir-SEG \citep{pogorelov2017kvasir} and CVC-EndoSceneStill \citep{vazquez2017benchmark} datasets, showing that PolypSeg+ outperforms existing models in both speed and accuracy. Additionally, \cite{ou2021polyp} have proposed a lightweight network based on YOLOv5. By targeting optimisations for endoscopic video inputs, the number of convolutional kernels is halved, and unnecessary large object detection heads are removed. The model achieves accuracy close to that of YOLOv3-spp, yet its size is only 1/30 of YOLOv3-spp.

Although wireless capsule endoscopy (WCE) is a crucial method for small-bowel investigation, its time consuming and tedious nature poses challenges for physicians \citep{wang2013wireless}. \cite{leenhardt2020cad} introduced CAD-CAP, a large multicentre database designed for the development of CAD tools for WCE image reading. It includes 100,000 annotated training images and 20,000 test images. \cite{segui2016generic} designed a lightweight CNN achieving a high accuracy of 96\% on CAD-CAP for the precise classification of nonpathological image features in the intestines, such as the wall of the bowel, bubbles, turbid substances, folds, and transparent spots.

% Furthermore, \cite{ding2019gastroenterologist} proposed a ResNet152-based CAD system to evaluate WCE images, utilising a dataset of 158,235 small-bowel WCE images from 77 medical centres and 1,970 patients for training to identify and classify small-bowel pathologies including inflammation, ulcers, polyps, lymphangiectasia, bleeding, vascular anomalies, protruding lesions, lymphoid hyperplasia, diverticula, and hookworms. The researchers compared the interpretations of 5,000 WCE videos of the small intestine by the system and 20 gastroenterologists. The system outperformed gastroenterologists in detecting anomalies in per-patient analysis (sensitivity 99.8\% versus 74.57\%; p\textless0.0001) and per-lesion analysis (sensitivity 99.90\% versus 76.89\%; p\textless0.0001). Also, the average reading time per patient for the system was significantly shorter at 5.9$\pm$2.23 minutes compared to the traditional reading by gastroenterologists of 96.6$\pm$22.53 minutes (p\textless0.001).

\cite{sahafi2022edge} introduced a capsule endoscopy device equipped with a Kendryte K210 chip, which enables real-time onboard analysis of gastrointestinal images through bidirectional communication with personal electronic devices such as smartphones or tablets. This innovation enhances the diagnostic efficiency of gastrointestinal diseases by employing a lightweight DNN to immediately identify abnormalities such as lesions or polyps during the examination. Moreover, the device also allows for task modifications and updates to the neural network via Bluetooth.

The Olympus EndoCapsule 10 System\footnote{ ENDOCAPSULE-10-System.\url{https://medical.olympusamerica.com/products/endocapsule-10-system}} sets a new standard for capsule endoscopy technology, integrating efficient image processing techniques with a user-friendly operating system. Equipped with a DL-based CAD system, Olympus ENDO-AID,\footnote{ENDO-AID.\url{https://www.olympus.com.au/medical/en/Products-and-Solutions/Products/Product/ENDO-AID.html}} it leverages advanced optical and computer vision technologies to deliver clear, detailed gastrointestinal images and evaluation reports, aiding physicians in the detection and assessment of various lesions, including inflammation, bleeding, or tumors in the small intestine. The EndoCapsule 10 features a highly sensitive camera capable of automatically adjusting lighting within the patient's body, ensuring high-quality images under any condition. Furthermore, the system provides an intuitive data management platform, supporting efficient review and analysis of image data by physicians, thereby further optimising the diagnostic process.

Volume Laser Endoscopy (VLE) offers wide-field imaging and can scan the esophageal wall layers up to 3 mm in depth at near-microscopic resolution, commonly used to detect and evaluate Barrett's Oesophagus (BE) \citep{swager2017computer}. VLE requires endoscopists to examine a large volume of image data in a short time (1,200 images from a 6 cm segment of the oesophagus in 90 seconds), which can be challenging. To address this, \cite{redmon2016you} proposed a miniaturised YOLO architecture based CAD method. Using partial residual networks achieved a detection accuracy of 98.23\% on resource-limited devices and reduced inference speed by a factor of 10 compared to standard YOLO \citep{selmanaj2021fast}.

% To address this, \cite{fonolla2019ensemble} developed a VGG16 based auxiliary diagnostic method, achieving a specificity of 0.85, a sensitivity of 0.95, and an area under the curve (AUC) of 0.96 on a validation dataset of 45 patients. 

Furthermore, \cite{le2022cnn} explored the potential of CNNs in classifying anatomical landmarks in upper gastrointestinal endoscopic images on edge devices. By employing quantisation techniques, they managed to reduce the memory and computational demands of the model, facilitating real-time diagnostics. Also, as the model allows processing more images per second, it potentially enables increasing the diagnostic accuracy. This work underscores the advantages of adopting lightweight techniques on edge devices to aid in medical diagnosis and treatment planning in gastroenterology.

\cite{kara2023smart} developed an intelligent handheld edge device equipped with a unique tactile sensing module and a dual-stage learning algorithm. This device is specifically designed for on-site diagnosis and histopathological assessment of resected colorectal cancer polyps, offering immediate insights into their texture and hardness. The practicality of this device demonstrates the transformative impact of edge computing in bridging the gap between data collection and interpretation in gastrointestinal diagnostics.

% To significantly accelerate real-time disease detection, \cite{pogorelov2016gpu} leveraged GPU acceleration. Similarly, using DL, both \cite{srivastava2019deep} and \cite{hmoud2021deep} made substantial strides in disease detection from gastrointestinal biopsy images and classification of gastrointestinal diseases, respectively. Additionally, \cite{oukdach2022gastrointestinal} showed the efficacy of transfer learning mechanisms in classifying gastrointestinal diseases. These advances demonstrate the capability of medium-range edge devices to handle more computationally intensive tasks, thus improving diagnostic accuracy and timeliness in gastrointestinal disease management.

\subsection{Pulmonary and Thoracic Diagnosis}
Diagnosis of lung and chest diseases are typically based on the analysis of chest X-ray (CXR) and CT scans. CXR imaging is the method of choice for screening of pneumonia, tuberculosis, emphysema, rib fractures, and cardiac abnormalities. Chest CT scans provide more detailed imaging for complex or difficult-to-diagnose cases, such as early-stage lung cancer, small lung nodules, pulmonary embolism, lung infections, and abnormalities of the thoracic organs. Local CAD systems not only enhance the confidence and accuracy of image interpretation, but also reduce the time taken for reading images. Moreover, in situations with limited resources or the need for rapid response, edge computing technology demonstrated advantages in quickly identifying diseases and promoting timely intervention.

Pneumonia, an acute disease, often requires rapid diagnosis and intervention. \cite{paluru2021anam} proposed Anam-Net, a lightweight CNN to segment anomalies in COVID-19 chest CT images, which can be deployed on embedded systems such as Raspberry Pi 4, NVIDIA Jetson Xavier, and Android applications for mobile devices. \cite{hou2020device} developed a method that employs subspace learning techniques for analysing CXR scans in edge devices, showing a significant reduction in computational and memory demands. This underscores the potential of subspace learning to facilitate efficient image analysis for pulmonary diagnosis on edge devices. 

Besides, \cite{abubeker2023b2} introduced B2-Net, a model designed to differentiate among normal pneumonia, bacterial pneumonia, and viral pneumonia in chest X-ray images. This approach leverages multiple ImageNet pretrained models integrated through Depth-Wise Convolution (DWC) and Squeeze-and-Excitation techniques, employing transfer learning for fine-tuning the models. The architecture of B2-Net is optimised for deployment on the Jetson Nano, offering rapid and accurate diagnostic support.

In addition, \cite{sait2019mobile} proposed a smartphone-based app for the preliminary detection of pneumonia using CXR images. The app uses a MobileNets model trained on a large dataset of CXR images from known pneumonia cases for detection. 
The app can quickly conduct a preliminary analysis of CXR images in conjunction with the camera of smartphone. In addition, the app includes an electronic diagnostic feature, allowing users to seek advice from qualified physicians based on the results. It also incorporates a respiratory pattern recorder module to enhance the app's predictive accuracy. 

Furthermore, EdgeMedNet proposed by \cite{liu2023edgemednet} is a lightweight U-Net architecture designed for efficient segmentation of medical images on edge devices. Tested on various thoracic imaging datasets, the framework demonstrated compelling performance in segmentation and classification tasks for pulmonary and thoracic diagnostics. Furthermore, to address clinical data security constraints, \cite{dong2023characterizing} integrated serverless edge computing with a browser-based medical imaging application (Fig.~\ref{fig:medical_application}). This integration aims to facilitate real-time image analysis while ensuring data security, which is crucial to the advancement of efficient and secure pulmonary and thoracic diagnostic methodologies on edge devices. \cite{france2020cluster} introduced a method employing cluster neural networks to harness the computational capabilities of edge devices to analyse thoracic imaging data. The method showcased an advancement in edge intelligence, enabling more sophisticated image analysis for pulmonary and thoracic diagnostics.

Regular low-dose CT screening for lung cancer can reduce the mortality rate by 20\% in high-risk groups \citep{national2011reduced,kauczor2015esr}. However, the number of CT scans can overwhelm radiologists, raising the false-positive rate. Thus, DL-based CAD may provide an alternative. \citep{masud2020light} proposed a lightweight CNN for mobile devices which achieved an impressive accuracy rate of 97.9\%. Additionally, \cite{raghavan2020initial} introduced a mobile low-dose CT screening device, aimed at improving opportunities for lung cancer screening, especially for uninsured and underinsured patients. The device, equipped with high-quality CT scanners and wireless internet connectivity, achieved lung cancer detection rates comparable to established trials such as the National Lung Screening Trial (NLST), even for patients not eligible for medical insurance. The study demonstrated its efficacy in reaching socio-economically disadvantaged groups, offering opportunities to detect early-stage lung cancer and increase survival rates at a lower cost per case. This mobile device represents a significant advancement in healthcare accessibility and has the potential to bring changes in lung cancer screening policies.

\begin{figure}[!t]
    \centering
    \includegraphics[width=0.6\textwidth]{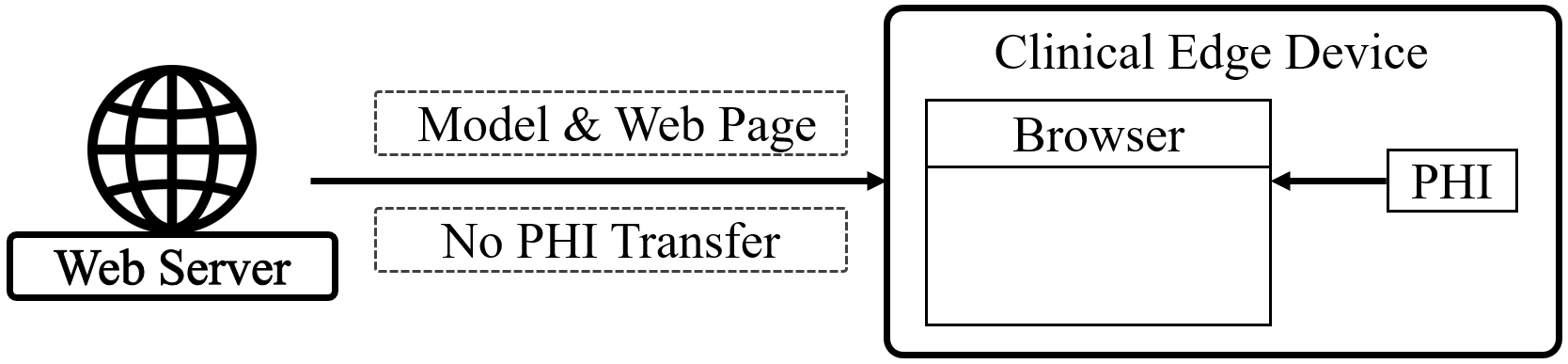}
    \caption{Browser-based pulmonary and thoracic diagnostics on clinical edge devices with secure patient health information (PHI) handling. The figure is adapted from the methodology presented in \cite{dong2023characterizing}.}
    \label{fig:medical_application}
\end{figure}

Several commercial edge devices have been developed specifically for pulmonary and thoracic diagnosis. GE Healthcare leveraged the Intel Distribution of the OpenVINO toolkit\footnote{OpenVINO™ Toolkit. \url{https://www.intel.com/content/www/us/en/developer/tools/openvino-toolkit/overview.html}} to improve the performance of pneumothorax detection algorithms in CXRs\footnote{Intel and GE Healthcare Partner to Advance AI in Medical Imaging. \url{https://www.intel.com/content/www/us/en/customer-spotlight/stories/ge-healthcare-medical-imaging.html}}. GE Healthcare also developed the portable X-ray device Optima XR240amx\footnote{Optima XR240amx. \url{https://www.gehealthcare.com/-/media/3072e4b953114c6cac1606e0e320e578.pdf?srsltid=AfmBOorU_H7B4HJ2ccuCE9xb82xvcLTsOQvhoEQImQeMD5JEyLwXNlGl}}, which performs real-time diagnostics using the multimodal DL-based Critical Care Suite\footnote{Critical Care Suit.\url{https://www.gehealthcare.com/en-ph/products/radiography/mobile-xray-systems/critical-care-suite}}. This suite comprises multiple DL models that work collaboratively to automatically detect and prioritise critical conditions such as pneumothorax. Specifically, the detection models focus on identifying and marking key abnormalities in the images, such as pneumothorax and lung nodules, while the segmentation models further isolate these abnormal regions, generating precise contours and area information. Moreover, the suite provides confidence scores for the detection and segmentation results, aiding clinicians in assessing the reliability of the diagnostic outcomes.

In addition, Butterfly Network\footnote{Butterfly iQ3. \url{https://www.butterflynetwork.com}} introduced a handheld ultrasound device that, through DL algorithms, offers various scanning modes and provides real-time image analysis and interpretation, enabling nonexpert users to perform ultrasound examinations.

\subsection{Dermatological Diagnosis}
The skin, as the largest organ of the body, serves as a vital barrier against environmental hazards like microbes, viruses, and pollutants. Skin diseases affect individuals across all age groups, stemming from various causes such as hereditary factors, lifestyle choices, and exposure to environmental elements. Among the prevalent skin disorders are acne, skin cancer, seborrheic keratosis, psoriasis, melanoma, and vitiligo. Given the ongoing and prevalent nature of skin diseases, their negative effects can significantly impair both the physical and psychological well-being of those afflicted.

Additionally, diseases such as melanoma can lead to severe consequences if not treated promptly. Diagnosing skin diseases from clinical images is particularly challenging due to the complexity, diversity, and similarity of these conditions. Moreover, manual diagnosis by medical experts is both time-consuming and subjective. Thus, accurate and timely diagnosis is essential for effective treatment and health management. The integration of edge computing in dermatological diagnostics has ushered in a new era of efficient and real-time analysis of skin images \citep{gocceri2020impact,goceri2021automated}. The ability to process and analyse data on edge devices close to the data source facilitates timely diagnostic feedback, thereby improving patient outcomes and optimising treatment strategies. We delve into several methods and frameworks that leverage edge computing for dermatological diagnosis.

\cite{goceri2021diagnosis} developed a lightweight network based on an improved MobileNet architecture and a hybrid loss function, specifically designed for efficient image classification on resource constrained devices such as smartphones. It enables mobile applications to diagnose five common skin diseases, including seborrheic dermatitis, rosacea, hemangioma, psoriasis, and acne vulgaris, with high accuracy. Additionally, \cite{mieras2018development} addressed the high prevalence of skin diseases in resource-poor areas by developing Skin\-App, a mobile app intended to assist healthcare workers in diagnosing and managing these diseases, including neglected tropical diseases (NTDs) with skin manifestations. Their research highlights the lack of healthcare personnel trained in dermatology in these areas and the potential of mobile health technology to bridge this gap.

In addition, \cite{shrivastava2023skin} proposed a lightweight model for distinguishing between benign and malignant skin lesions, tailored for the Jetson Nano platform. Utilising transfer learning from pretrained networks such as ResNet50 and MobileNet, this model achieves a classification accuracy of up to 93.3\% on the PH2 skin lesion dataset \citep{senan2021classification}. This approach demonstrates the integration of lightweight models with compact hardware platforms to provide portable and efficient healthcare solutions.

\cite{dai2019machine} presented an app for mobile devices for rapid detection and diagnosis of skin diseases. Furthermore, they highlighted the limitations of cloud-based ML, such as privacy concerns and latency. The authors deployed a pretrained neural network on mobile devices to ensure that all computations are performed locally, thereby reducing latency, saving bandwidth, and enhancing privacy. This method underscores the advantages of mobile edge computing in the diagnosis of skin diseases.

To facilitate early skin cancer detection in resource limited rural areas of developing countries, \cite{ngeh2020deep} developed a low-cost, user-friendly, and internet-independent prescreening device capable of classifying skin abnormalities and performing region segmentation. The device, powered by a Raspberry Pi and a CNN trained on the Skin Cancer MNIST dataset \citep{tschandl2018ham10000}, offers a practical solution for remote skin cancer assessment. Providing prescreening to identify high-risk individuals, it optimises the allocation of medical resources and improves early detection in underserved areas.

Transitioning to more advanced frameworks, \cite{shi2023skin} explored a federated contrastive learning framework for automatic skin lesion diagnosis. The proposed federated learning approach adeptly navigates the challenges of data silos while enhancing the model's diagnostic generalisability to unseen data. This work stands as a testament to the potential of federated learning in conjunction with edge computing to deliver robust and privacy-preserving dermatological diagnostic solutions.

The VISIA Complexion Analysis system\footnote{VISIA. \url{https://www.canfieldsci.com/imaging-systems/visia-complexion-analysis}} by Canfield Scientific integrates various imaging and DL technologies to provide a comprehensive analysis of skin health. Through RBX technology,\footnote{Canfield Scientific RBX® Technology Overview. \url{https://www.canfieldsci.com/research/stories/white-paper-rbx-technology-overview}} VISIA conducts in-depth inspections of wrinkles, spots, pores, textures, and subdermal issues that are invisible to the naked eye. The system not only aids in detecting and quantifying various skin conditions to develop targeted treatment plans but also enhances patient engagement through personalised reports and progress tracking. As a modular and adaptable platform, VISIA allows for ongoing updates and feature integrations, making it a crucial tool for comprehensive dermatological diagnosis and personalised skincare solutions.

% Furthermore, the AI-Skin framework proposed by \cite{chen2020ai}, embarked on a novel trajectory by integrating a self-learning mechanism with a wide data collection strategy within a closed-loop framework. This framework aimed at improving the recognition accuracy of skin diseases, thereby presenting a robust method for real-time, extendable, and individualised skin disease medical services.

\subsection{Pathological Images Analysis}
Microscopic examination of tissue slides (histopathology) is considered the gold standard for cancer diagnosis and prognosis. This process requires pathologists to identify subtle histopathological patterns within highly complex tissue images. It is time-consuming, subjective, and prone to considerable inter-observer and intra-observer variability. Hematoxylin and eosin (H\&E) staining is the most popular method for tissue staining. With the advent of whole-slide imaging (WSI) technology, a large number of H\&E stained tissues can be scanned, digitally represented, and archived. The analysis of WSIs in pathology using CAD systems is becoming a routine clinical practice. We focus on lightweight yet powerful models and introduce groundbreaking methodologies that have successfully harnessed edge computing for efficient and accurate analysis of histopathological images.

\cite{auguste2015mobile} developed an economical method for pathology in resource limited settings. Their method uses a standard optical microscope, an iPhone 5s, and a custom 3D-printed adapter to capture high-quality WSIs. Compatible with various stains including H\&E, this method is particularly cost-effective and portable, making it a viable option for medical facilities in developing countries where traditional WSI systems are prohibitively expensive. This innovation has significant potential to enhance diagnostic capabilities and improve healthcare outcomes in underserved areas.

Similarly, \cite{ramey2011use} evaluated the feasibility of interpreting WSIs in pathology, particularly frozen sections (FS), using a mobile viewing device (MVD). Their study involved scanning FS samples and assessing them on an iPad using the Interpath app. The study found a high concordance of 89\% between the initial FS diagnosis and the diagnosis made using the iPad, with the latter taking an average time of less than 2 minutes per slide. Set against the backdrop of increasing digitisation in the medical field, this study highlights the potential of using MVDs like iPads for the analysis of WSI images.

For the classification of histopathological images, \cite{datta2023novel} introduced ReducedFireNet. This lightweight model stands out for its compact size (merely 0.391 MB) and low computational demand (0.201 GFLOPS), making it exceptionally suitable for edge devices with limited processing capabilities. Despite its lightweight nature, the model achieved an impressive average accuracy rate of 96.88\% and an F1 score of 0.968 on the Malignant Lymphoma datasets, as per the settings of reference \citep{orlov2010automatic}. This dataset comprises 374 histopathological images, including types such as Lymphocytic Leukemia (CLL), Mantle Cell Lymphoma (MCL),  Follicular Lymphoma (FL) stained with hematoxylin and eosin (H\&E). The success of ReducedFireNet demonstrates its potential in facilitating timely and accurate disease diagnoses, especially in scenarios where rapid and efficient image analysis is critical.

To enable the detection of cervical cancer cells on embedded devices, \cite{wang2019recognition} proposed a lightweight network incorporating optimised convolution operations, model parameter compression, and enhanced feature expression depth in the network structure design. It achieves a 94.1\% accuracy rate on the NVIDIA Jetson TK1 embedded device, with fewer model parameters compared to ResNet18 and MobileNet.

There exist several edge devices employing DL for histopathological and cytopathological analysis. Cell Metric X,\footnote{Cell Metric® X. \url{https://www.aicompanies.com/cell-line-development/cell-metric-x}} a high-resolution imager driven by DL technology, enhances the efficiency and accuracy of cell line development processes. Maestro TrayZ's real-time cell impedance monitoring technology\footnote{Maestro TrayZ. \url{https://www.axionbiosystems.com/products/cell-analysis/maestro-trayz}} provides critical data for cell health and function, and offers real-time, noninvasive monitoring, widely used in cytotoxicity and cell growth studies. Digital pathology tools have radically changed nephropathology by improving the accuracy of disease detection and classification. zenCELL owl\footnote{zenCELL owl. \url{https://zencellowl.com}} is a compact automated microscope for monitoring cell cultures, essential for accurate and reliable cell analysis. And CytoPAN,\footnote{CytoPAN. \url{https://getzpharma.com/product/cytopan}} a portable image cytometer, shows exceptional diagnostic precision in identifying breast cancer subtypes, especially ER/PR and HER2 \citep{min2020cytopan}.

\subsection{Telemedicine}
With the rapid advancements in internet-of-things (IoT) and information and communication technologies (ICT), telemedicine is emerging as a promising healthcare service model, gradually becoming a vital component of the global healthcare sector \citep{azimi2018empowering,ye2023implications}. Telemedicine, powered by edge computing, presents an alternative to traditional diagnostic and preventive measures. In recent years, the evolution of 5G technology and mobile health applications has unveiled the potential of telemedicine in chronic disease management and remote diagnostics. We survey how edge computing empowers telemedicine, enhancing the accessibility and efficiency of healthcare services.

\cite{van2017validity} embarked on an innovative exploration of telemedicine capabilities, particularly in the context of managing diabetic foot ulcers (DFUs). They used images captured by iPhones for remote assessment of DFUs, carefully evaluating the accuracy and reliability of such digital diagnoses compared to traditional on-site clinical assessments conducted by experienced podiatrists. The study integrated a comprehensive analysis of clinical features and treatment decisions, and examined the key limitations of mobile image-based DFU assessments. Subsequently, \cite{yap2018new} introduced an iPad-based DFU diagnostic app, FootSnap, and tested its reliability across different operators and patients. The results indicated high intra-operator and inter-operator reliability, with Jaccard similarity index values ranging from 0.89 to 0.94, demonstrating that FootSnap can effectively monitor the condition of diabetic feet longitudinally.

Furthermore, \cite{goyal2018robust} proposed a DFU detection and localisation method based on Fast R-CNN, compiling a dataset of 1,775 DFU images annotated by medical experts and optimising the model through transfer learning. The authors showed that the model can achieve 91.8\% average precision on an NVIDIA Jetson TX2 with an inference speed of 48 ms per image. This demonstrates the effectiveness and practical potential of DL approaches for DFU diagnosis on medium-range devices.

To support remote medical practices in managing chronic skin ulcers, \cite{cazzolato2021utrack} launched the UTrack mobile app, which facilitates remote wound image capture, segmentation, measurement, and monitoring. It leverages an innovative unsupervised segmentation method, UTrack-Seg, using mobile device cameras to accurately segment and measure ulcers while storing data locally to protect privacy. The app demonstrates superiority in accuracy and speed, providing patients and healthcare providers with an effective tool for monitoring the progress of ulcer healing.

Focusing on cloud and edge computing for fall detection, \cite{mrozek2020fall} discussed a scalable telemedicine system for remote monitoring of the elderly. It achieves efficient monitoring and alert mechanisms through mobile IoT devices and shows how edge-based processing can reduce data transmission needs and storage consumption, offering practical solutions for monitoring indoor and outdoor activities of the elderly. This ensures timely assistance in the case of falls, a major health risk for the elderly, contributing to improved care and self-sufficiency.

For gait analysis, \cite{martini2022enabling} introduced a telemedicine app based on 3D HPE using the NVIDIA Jetson Xavier. It employs DL for markerless motion capture, balancing accuracy, portability, and privacy. The app is also suitable for monitoring the elderly, providing real-time and high-precision capabilities. It showcases the potential for remote gait analysis, meeting the demand for scalable, efficient telemedicine solutions.

\cite{han2022super} introduced a method named SR-Telemedicine to enhance video quality for telemedicine. It focuses on upgrading very low-resolution video chunks to high resolutions such as 720p or 1080p via a DL-based super-resolution model. The method aims to provide high-quality videos for physicians without requiring extensive network bandwidth. The approach leverages a scalable neural network model and a double deep Q-Network (DDQN) algorithm to dynamically adjust the video resolution and model scale based on network and computational capabilities, significantly improving the user's quality of experience by 17–79\% over baseline methods.

TytoCare\footnote{TytoCare. \url{https://www.tytocare.com}} is a versatile handheld examination device capable of capturing a variety of medical data, including heart and lung sounds, throat images, skin conditions, ear canal images, and body temperature. The device utilises advanced computer vision and DL technology to analyse collected data, ensuring diagnoses are comparable in accuracy to in-person consultations. Seamless integration with smartphones and tablets facilitates real-time communication between patients and healthcare providers, enabling immediate medical consultations and diagnoses. The platform supports wireless updates, allowing continuous improvements to its diagnostic algorithms and functionalities to meet evolving healthcare needs.

Proximie\footnote{Proximie. \url{https://www.proximie.com}} is a pioneering platform in the field of remote surgical collaboration and education, designed to transcend geographical boundaries, enabling surgeons to virtually participate in operating rooms across distances and offer their expertise. The platform employs augmented reality (AR), computer vision, and DL algorithms to enhance the visualisation of surgical procedures, facilitating precise guidance, mentoring, and decision making. Proximie's capability to record and analyse surgeries creates a rich database for educational purposes and continuous learning.

Furthermore, Philips Lumify\footnote{Philips Lumify. \url{https://www.philips.com.au/healthcare/sites/lumify-handheld-ultrasound/products/reacts-tele-ultrasound}} offers a remote ultrasound solution that, beyond offering portable handheld ultrasound acquisition, optimises video management, transmission, and communication to deliver superior and efficient remote diagnostics and collaboration. Its intuitive, user-friendly interface simplifies the diagnostic process, enabling a broader range of healthcare professionals to utilise advanced ultrasound technology.

\section{Future Directions}
\label{sec:future}

Notwithstanding many impressive recent advances surveyed above, there is much room for further development of Edge DL in computer vision and medical applications. Complementing our discussion of the state of the art in this nascent field, we finally discuss potential future directions in improving critical aspects of Edge DL, including privacy and security, energy efficiency and performance optimisation, adaptive edge computing architectures, multimodal edge computing fusion techniques, efficient edge inference in resource-constrained environments, explainable AI in edge models, quantum-inspired edge computing, and human-centric edge applications.

\subsection{Privacy and Security}
Edge computing introduces both opportunities and challenges in ensuring data privacy and security. One major area of focus is defending against adversarial attacks that exploit small perturbations in input data to mislead models, leading to dangerous misdiagnoses in medical contexts. To address this challenge, researchers are developing new adversarial training techniques that enhance model robustness by incorporating adversarial samples during training  \citep{isakov2019survey}. Additionally, there is ongoing exploration into novel regularisation techniques and network architectures to increase model resilience against input disturbances \citep{jha2021trinity,vairo2023approach}. 

Another research direction involves utilising privacy-preserving technologies, such as homomorphic encryption and secure multiparty computation (SMPC), to protect training data and model weights from leaking sensitive information \citep{giannopoulos2018privacy}, which is particularly crucial in healthcare where patient data confidentiality is paramount. Ensuring the ethical handling of patient data is critical, as breaches can lead to severe consequences for patient trust and safety. Moreover, federated learning \citep{zhang2021survey} enables model training across multiple edge devices without centralising sensitive data, ensuring that patient information remains local. This approach minimises privacy risks while enabling collaborative learning between institutions. Lastly, advances in trusted execution environments (TEE) \citep{munoz2023survey} at the hardware level offer promising security features by isolating sensitive computations, further reducing vulnerabilities on edge devices.

\subsection{Energy Efficiency and Performance Optimisation}
Energy efficiency and performance optimisation in edge computing require innovations at both hardware and software levels. In medical applications, such optimisations are critical not only for continuous patient monitoring and portable medical devices but also for enabling accurate real-time diagnostics in resource-constrained environments. For instance, devices such as wearable electrocardiogram monitors or portable ultrasound scanners rely on energy-efficient edge processors to provide immediate diagnostic insights without relying on cloud infrastructure. On the hardware front, research is focused on developing processors and storage solutions specifically designed for edge computing, such as low-power microcontrollers, ASICs, and dedicated neural network accelerators \citep{li2020survey,reuther2021ai}.

Software-level optimisations encompass more efficient data compression algorithms and lightweight neural network architectures, designed to minimise computational and storage requirements while sustaining high performance. Additionally, energy-aware scheduling algorithms, which dynamically optimise task allocation and execution based on the device's energy state and computational demands, are a current research hotspot \citep{chang2019ai}. Techniques such as adaptive voltage scaling and power gating are also being integrated at the software level to further reduce energy usage during idle periods. These techniques are particularly useful in scenarios like wearable health monitoring devices, where the system spends a significant portion of time in low-power states but must remain ready for immediate activation in response to critical events, such as detecting arrhythmias \citep{sarma2020vlsi,yousri2023power}.

\subsection{Adaptive Edge Computing Architectures}
The key to adaptive edge computing architectures lies in using advanced algorithms to automatically adjust networks to the ever-changing environment and application requirements. NAS techniques show great potential in this area, automatically optimising the network design to achieve the best performance within given resource constraints \citep{wistuba2019survey}. In particular, hardware-aware NAS \citep{benmeziane2021comprehensive} has emerged as an essential tool, considering device-specific metrics like latency, memory usage, and power consumption during the search process. For instance, ProxylessNAS \citep{cai2018proxylessnas} is a prominent example of hardware-aware NAS applied to edge computing. It introduces a latency constraint directly into the search process, optimising the architecture for specific hardware, such as mobile devices. By conducting the search directly on target hardware instead of relying on proxy tasks, ProxylessNAS can efficiently generate models with a good balance of latency and accuracy.

In addition, dynamic solutions such as Slimmable Neural Networks \citep{yu2019slimmable} allow the width of the network to be dynamically adjusted during inference to suit the available computational resources. This method allows a single model to run at different scales, from full network capacity to a more lightweight version, depending on real-time resource availability \citep{han2021dynamic}.

Another practical approach is the use of resource-aware dynamic pruning strategies \citep{liu2019dynamic}, where unimportant channels are pruned during inference based on input data. This enables the model to adaptively reduce its computational complexity when processing less demanding tasks, optimising both power consumption and performance for edge devices. These specific methods highlight the growing integration of adaptive mechanisms within edge computing architectures, offering tailored, resource-efficient solutions for various tasks and conditions, ensuring flexibility and responsiveness across different device and workload scenarios.

\subsection{Multimodal Edge Computing Fusion Techniques}
Multimodal edge computing aims to process and analyse heterogeneous data from various sensors and sources. In healthcare, this involves integrating data from modalities such as medical imaging, electronic health records, and wearable sensors to provide comprehensive patient assessments \citep{calisto2017medical}. A key challenge lies in developing effective multimodal fusion algorithms that can integrate data from diverse modalities like vision, audio, and text, extracting useful information to support complex decision-making processes. For example, DL and data fusion techniques are being employed to integrate data from multiple devices for more accurate health monitoring and event response. In this domain, researchers are also exploring how to maintain data processing efficiency while ensuring the generalisability and accuracy of algorithms \citep{acosta2022multimodal}, which is vital for reliable and timely medical diagnoses.

Recent advances in transformer models, particularly the Vision-Language Transformer (ViLT) \citep{fields2023vision}, have been instrumental in fusing visual and textual data with minimal preprocessing. ViLT uses a joint embedding space to represent both image and text modalities, allowing it to perform multimodal tasks efficiently, such as medical report generation or cross-modal retrieval \citep{delbrouck2022vilmedic}. The simplification of cross-modal interactions in ViLT demonstrates how transformer-based architectures can potentially be adapted to handle multimodal tasks with reduced computational overhead, which is a key consideration in edge environments. Another emerging approach is the use of contrastive learning for multimodal fusion, where models are trained to maximise agreement between different modalities while maintaining modality-specific representations. For instance, the CLIP model \citep{radford2021learning} has been adapted for medical tasks to align radiology images with text-based reports, making it a powerful tool for image retrieval and diagnostic support \citep{zhao2023clip}. 

\subsection{Efficient Edge Inference in Resource-Constrained Environments}
In resource-constrained environments, efficient edge inference relies not only on model optimisation but also involves intelligent energy management and task scheduling strategies. These strategies maximise energy efficiency by optimising the allocation of computational tasks across different types of processing units, such as CPUs, GPUs, and FPGAs, as well as facilitating collaborative work between edge and cloud computing resources \citep{Krzywda2018power}. Additionally, efficient processing of real-time data streams is a key component of achieving efficient edge inference. These strategies and optimisations can significantly enhance the performance and energy efficiency of edge computing devices in resource-limited settings \citep{zhou2019edge,mao2017survey,yu2017survey}.

One prominent example of optimising edge inference speed is SparseNN \citep{zhu2018sparsenn}, which focuses on sparsity-aware execution. SparseNN leverages the inherent sparsity in neural networks by skipping computations involving zero-value activations and weights, significantly reducing the overall computational load. Similarly, ShiftAddNet \citep{you2020shiftaddnet} replaces computationally expensive multiplication operations in convolutional layers with lightweight shift and add operations. ShiftAddNet maintains competitive performance while reducing latency and resource usage, demonstrating a practical balance between accuracy and efficiency in edge computing environments.

\subsection{Explainable AI in Edge Models}
Explainable AI (XAI) has emerged as a key aspect of deploying AI systems. The demand for transparency, trust, and compliance with regulatory requirements has driven the development of explainability, which is crucial for sensitive applications in healthcare and autonomous driving. Current integrated learning approaches, such as LIME \citep{ribeiro2016should} and SHAP \citep{lundberg2017unified}, provide visual explanations that elucidate how various features influence model decisions. Recent research has focused on developing lightweight and real-time explainability frameworks suitable for edge devices \citep{gilpin2018explaining,li2023lightweight,huang2022real,saini2023e2alertnet}. While progress has been made in integrating XAI into edge models, challenges remain, such as the trade-off between model performance and explainability, and the need for standardised frameworks for model explainability.

In the context of healthcare, explainability becomes even more critical. Medical professionals require clear, understandable insights into how AI models arrive at specific diagnoses or treatment recommendations. This transparency not only aids in clinical decision-making but also fosters trust among patients and healthcare providers. In addition, edge-based diagnostic tools that use XAI can help doctors understand the reasoning behind AI-driven imaging analyses, making it easier to justify and trust AI recommendations in a clinical setting. Therefore, enhancing the explainability of edge models in medical applications is not just a technical challenge but a necessity for broader acceptance and effective use in healthcare environments.

\subsection{Quantum-Inspired Edge Computing}
Quantum-inspired edge computing leverages the principles of quantum computing to enhance traditional computational methods for edge applications. In medical diagnostics, quantum-inspired techniques can potentially revolutionise data processing capabilities and optimise resource allocation. Researchers are exploring the use of quantum ML to enhance efficiency \citep{biamonte2017quantum}. Quantum teleportation methods may enhance secure communications in edge computing \citep{pirandola2015advances,liu2020applications} and quantum optimisation methods such as quantum annealing may help solve complex optimisation problems and improve performance \citep{das2008colloquium}. The integration of quantum-inspired technologies in edge computing is still in its early stages, but it offers broad prospects for overcoming the limitations of classical computing in edge environments. Future research may focus on creating more powerful quantum algorithms, developing quantum-resistant security protocols for edge networks, and exploring the potential of hybrid classical and quantum systems to improve diagnostic accuracy and speed in medical applications.

\subsection{Human-Centric Edge Applications}
Human-centric edge applications emphasise technology design and services centred around human needs and experiences. In health monitoring, human-centric edge applications can utilise portable edge devices for instantaneous analysis and detection of health data to provide customised services, personalised feedback, and health advice. This enables users to better understand their health condition, potentially identifying health issues early and promoting proactive health management \citep{garcia2017midgar,abrantes2023external}. In smart homes, human-centric edge applications leverage various edge devices for user interaction and environmental monitoring, automating home appliances and systems for optimal comfort and energy efficiency \citep{patchava2015smart,schneider2018activity}. An example is the use of thermal imaging to detect human body temperature and adjust room temperature accordingly \citep{vazquez2012thermal}. While human-centric edge applications offer many benefits, they also present challenges related to privacy, data security, and computational limitations. Protecting user data while providing personalised services requires robust security protocols and careful handling of sensitive information, which are key topics of future research \citep{calisto2022modeling,lakshminarayanan2023health}.

\subsection{Integration of Advanced Learning-Based Methods}
Recent years have also witnessed significant advancements in learning-based diagnostic methods for medical images, including the use of transformers \citep{you2022class}, contrastive learning \citep{you2022momentum,you2023bootstrapping,you2023action++}, few-shot learning \citep{you2022mine, you2024rethinking}, transfer learning \citep{you2021incremental,you2022simcvd}, domain adaptation \citep{you2020unsupervised} and generative models \citep{you2018structurally,you2023implicit}. Transformers, with their self-attention mechanisms, have shown a remarkable ability to capture intricate patterns and long-range dependencies in medical images, significantly improving diagnostic accuracy. Contrastive learning leverages unlabeled data to learn robust feature representations, reducing the reliance on large labeled datasets and enhancing model generalisation. Few-shot learning aims to train models with a very small amount of labeled data, which is particularly valuable in medical diagnostics where labeled data is often scarce. Transfer learning allows for fine-tuning or distillation of models pre-trained on large datasets, enabling them to effectively handle specific medical image diagnostic tasks. Domain adaptation techniques address the challenge of varying imaging conditions and equipment by aligning feature distributions between different domains, thereby improving model generalisation. Integrating these advanced learning-based technologies with edge computing can enhance the accuracy and efficiency of medical diagnostics, contributing to more reliable analysis and support in clinical settings.

\section{Conclusion}
\label{sec:conclusion}
The integration of edge computing and deep learning (Edge DL) offers unprecedented opportunities for real-time processing and interpretation of data close to the source in resource-constrained settings. Focusing on applications in computer vision and medical diagnostics, we have explored the fundamental concepts and technical merits of Edge DL, underlining its critical role in the evolution of modern computational paradigms. Specifically, we have delved into model compression and the design of lightweight models suitable for operation on edge devices. Reducing latency, conserving bandwidth, and bolstering data privacy, Edge DL holds the promise of revolutionising various industries, from autonomous vehicles to smart healthcare systems. In surveying the current landscape and potential future developments of Edge DL, we have highlighted its burgeoning possibilities and challenges. No doubt, continuing advances in hardware and software for Edge DL will have an increasingly transformative impact in a growing range of applications. It is our hope that Edge DL will contribute to improve our daily lives and bring us closer to the objective of universal healthcare accessibility.

\section*{Acknowledgments}\label{sec:acknowledgments}
The authors would like to express their gratitude to Matthew Sheedy, George Bou-Rizk, Tim Kannegieter, and Geoff Sizer from Genesys Electronics Design, Sydney, Australia, for engaging and insightful discussions that have been invaluable in refining our understanding and conceptual framing of the discussed themes. This work was supported by the Australian Research Council (ARC) Research Hub for Connected Sensors for Health (grant number IH210100040). Partial financial support was received from Genesys Electronics Design, Sydney, Australia.

% \bibliographystyle{sn-mathphys-ay}
% \bibliography{paper}
% BioMed_Central_Bib_Style_v1.01
% \section*{References}

\end{document}